\documentclass{article}
\usepackage{enumitem}
\usepackage[preprint, nonatbib]{neurips_2025}
\usepackage{comment}
\usepackage[utf8]{inputenc} 
\usepackage[T1]{fontenc}    
\usepackage{hyperref}       
\usepackage{url}            
\usepackage{booktabs}       
\usepackage{amsfonts}       
\usepackage{nicefrac}       
\usepackage{microtype}      
\usepackage{xcolor}         
\usepackage{enumitem}
\usepackage{lipsum}
\usepackage{tikz}
\usepackage{xcolor}
\usepackage[dvipsnames]{xcolor}
\usetikzlibrary{calc}

\usepackage{subcaption}
\usepackage{graphicx}
\usepackage{amsmath}
\usepackage{amssymb}
\usepackage{cancel}
\usepackage{makecell}
\usepackage{array}
\usepackage{multirow}
\usepackage{ragged2e}
\usepackage{adjustbox}
\newcommand\rev[1]{\textcolor{black}{#1}}
\newcommand\revv[1]{\textcolor{black}{#1}}

\newcommand{\norm}[1]{\left\lVert #1 \right\rVert_2}

\title{ANCHOR: Error-Controlled Adaptive Numerical Correction for Neural Operator Time Marching}

\author{%
  Rajyasri Roy \\
  Department of Civil and Systems Engineering\\
  Johns Hopkins University\\
  Baltimore, MD, 21218\\
  \texttt{rroy13@jh.edu}\\
  \And
  Dibyajyoti Nayak \\
  Department of Civil and Systems Engineering\\
  Johns Hopkins University\\
  Baltimore, MD, 21218 \\
  \texttt{dnayak2@jh.edu} \\
  \And
  Somdatta Goswami \\
  Department of Civil and Systems Engineering \\
  Johns Hopkins University\\
  Baltimore, MD, 21218 \\
  \texttt{somdatta@jhu.edu} \\
}

\begin{document}

\maketitle

\begin{abstract}
Numerically solving time-dependent partial differential equations (PDEs) underpins modern science and engineering; but high-fidelity solvers are often far too slow for long horizons, many-query studies, or real-time use. Neural operator (NO) surrogates promise a way out by delivering rapid inference over rich parametric and functional inputs. Yet, despite recent progress, most autoregressive and related NO frameworks remain vulnerable to compounding rollout errors, and standard ensemble-averaged test metrics offer weak guarantees on the reliability of any single trajectory or its error bounds. In many engineering applications, where individual operating conditions must be predicted reliably, errors can silently accumulate to unacceptable levels even when mean errors appear small, and existing methods offer no principled tools for online monitoring, control, or correction of error growth beyond the training horizon. We introduce ANCHOR (\textbf{A}daptive \textbf{N}umerical \textbf{C}orrection for \textbf{H}igh-fidelity \textbf{O}perator \textbf{R}ollouts), an online, instance-aware, error-controlled hybrid time-marching strategy for nonlinear, time-dependent PDEs. ANCHOR treats a pretrained NO as the fast primary engine and dynamically couples it to a classical numerical solver via a physics-informed, residual-based error estimator. Drawing inspiration from adaptive time-stepping in numerical analysis, ANCHOR continuously tracks an exponential moving average (EMA) of the normalized PDE residual, using it as an online alarm to detect error accumulation and trigger targeted solver corrections, without ever needing ground-truth solutions. We show \revv{both theoretically and empirically} that this EMA-based estimator is strongly correlated with the true relative $L_2$ error across a wide range of dynamical regimes, enabling instance-specific, data-free error control during inference. Extensive experiments on \revv{six} canonical PDEs, 1D and 2D Burgers’, 2D Allen-Cahn, \rev{2D Cahn-Hilliard,} \revv{2D Navier-Stokes,} and 3D heat conduction, demonstrate that ANCHOR consistently bounds long-horizon error growth, stabilizes challenging extrapolative rollouts, and markedly boosts robustness over standalone NOs, all while remaining far more efficient than full high-fidelity solvers. By fusing the speed of data-driven surrogates with the reliability of classical numerics, ANCHOR provides a principled route to trustworthy long-horizon neural surrogates and robust digital twins for complex dynamical systems.
\end{abstract}

\section{Introduction}
\label{sec:intro}
High-fidelity numerical solvers remain the gold standard for simulating complex time-dependent physical systems governed by partial differential equations (PDEs). Their robustness, interpretability, and adherence to physical laws make them indispensable across science and engineering. However, their substantial computational cost often renders them impractical for time-critical, many-query, or large-scale applications. \revv{In many practical settings, practitioners have already committed to neural operator~\cite{lu2021learning, lifourier, cao2024laplace, raonic2023convolutional, li2020neural, rahman2022u, tripura2023wavelet} surrogates as the primary inference engine, for precisely these reasons.} Neural operators offer rapid inference by learning mappings directly between function spaces. Yet this efficiency frequently comes at the expense of \revv{reliability in solution accuracy}. When deployed for time-dependent problems in an autoregressive, Markovian fashion, even small local approximation errors can accumulate over successive time steps, leading to numerical instability and loss of physical fidelity~\cite{nayak2026ti, nayak2025data, abueidda2026time, mandl2026physics}. The central challenge, therefore, is not merely short-horizon accuracy, but the stability and trustworthiness of learned surrogates under long-horizon inference for an unseen case. \revv{Critically, this challenge is independent of whether a high-fidelity solver is affordable in principle: even when a numerical solver could be run, purely data-driven rollouts provide no online mechanism to detect error accumulation, no per-instance guarantees, and no way to determine whether a given operating condition lies near or beyond the boundary of the training distribution; all without access to ground-truth solutions.} 
Bridging this divide requires treating learned surrogates as approximate numerical update operators and embedding them within principled error control and correction mechanisms.
\begin{figure}[htb!]
    \centering
    \includegraphics[width=\linewidth]{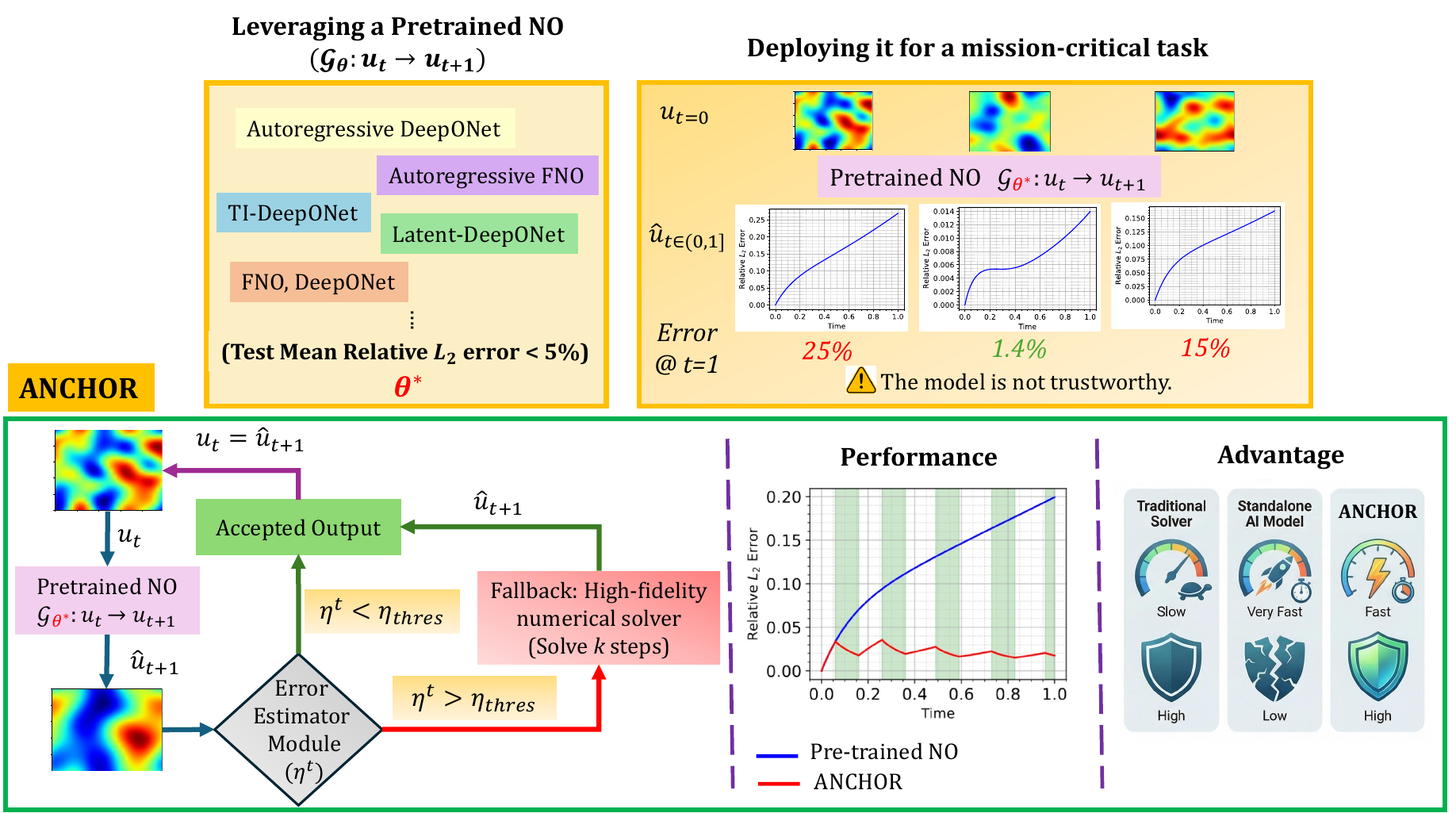}
    \caption{Schematic of the proposed ANCHOR framework, which couples a neural operator with a high-fidelity numerical solver. \revv{Note that the numerical solver is not an alternative to ANCHOR; it is a component of ANCHOR, invoked selectively as a corrective mechanism when the error estimator detects accumulated error beyond an acceptable threshold. ANCHOR leverages the complementary strengths of both approaches: the computational efficiency of the neural operator surrogate and the physical fidelity of the classical numerical solver.}}
    \label{fig:hybrid_framework_schematic}
\end{figure}

To balance efficiency, accuracy, and stability, a growing body of work has explored hybrid modeling strategies that combine classical numerical solvers with machine learning models. Broadly, these approaches can be grouped into several categories. One line of work focuses on domain-decomposition-based coupling, where operator learning surrogates accelerate selected subdomains while classical solvers ensure global consistency. For example, Wang et al.~\cite{WANG2025118319,wang2026non} coupled a physics-informed DeepONet (PI-DeepONet) with a finite element method (FEM) solver via Schwarz alternating methods (SAM) to accelerate simulations requiring local mesh refinement, while Tezaur et al.~\cite{tezaur2025hybrid} introduced overlapping SAM (O-SAM) to couple operator inference reduced-order models (ROMs) with full-order models (FOMs). A second category comprises solver-in-the-loop~\cite{um2020solver} and staggered hybrid schemes, such as I-FENN~\cite{amin2025fenn, pantidis2026integrated}, where DeepONet is embedded directly into FEM to predict coupled fields while the solver advances the primary state. More broadly, the coupling of high-fidelity simulations with ROMs has gained significant traction~\cite{BONNEVILLE2024116535, mcquarrie2023nonintrusive, hesthaven2018non}. Additional paradigms include differentiable physics frameworks that embed solvers into end-to-end learning pipelines~\cite{hu2019difftaichi}, learning-based correction of coarse-grid CFD solvers~\cite{kochkov2021machine, tompson2017accelerating}, neural-augmented multigrid methods~\cite{katrutsa2020black, greenfeld2019learning, he2019mgnet}, adaptive time-stepping strategies learned from data~\cite{owoyele2022chemnode}, and neural interface conditions for ROM-FOM coupling~\cite{fresca2022pod}. Recent hybrid modeling efforts addressing coupled multiphysics systems, including those by Kara et al.~\cite{kara2025physics}, further emphasize solver specialization across designated physical fields.

Despite these advances, most existing hybrid methods target regimes where error accumulation is weak or secondary, including steady-state or weakly time-dependent systems, linear PDEs, or spatial coupling strategies aimed primarily at reducing solver complexity~\cite{zhang2024blending, amin2025fenn, WANG2025118319}. In contrast, robust long-horizon time marching for nonlinear, time-dependent PDEs remains largely unresolved. In such settings, error accumulation is intrinsic: repeated surrogate rollouts can lead to exponential error growth, instability, and violation of physical constraints. Architectures such as the Time-Integrator Deep Operator Network (TI-DeepONet)~\cite{nayak2026ti} partially mitigate this issue by embedding numerical integration schemes directly into the learning process, significantly reducing error accumulation compared to standard autoregressive neural operators beyond the interpolation regime. Other approaches attempt to address long-horizon prediction by augmenting DeepONet with temporal branches~\cite{diab2025temporal}, employing recurrent architectures such as RNNs, LSTMs, or GRUs on top of neural operator outputs~\cite{michalowska2024neural}, introducing memory modules to capture non-Markovian effects~\cite{buitragobenefits, he2024sequential, hu5149007deepomamba}, \rev{or learning adaptive time-stepping strategies within the surrogate itself through a neural Taylor expansion, such as TANTE~\cite{wu2025tante}}. While these strategies can improve short- to medium-horizon performance, none provide mechanisms to detect, control, or bound error growth during time marching, particularly beyond the training horizon. As a result, purely data-driven rollouts remain vulnerable to error drift and loss of physical consistency in underrepresented regions of the state space.

\revv{A parallel, and equally important, challenge lies in how surrogate models are evaluated and deployed. The prevailing practice is to report average error metrics, such as mean squared error, computed over large test sets. While useful for benchmarking, these metrics are poorly aligned with real-world deployment. In many engineering applications, such as structural safety assessment, real-time control, and digital twins for specific physical assets, surrogate models are rarely used in an ensemble-averaged sense; instead, they are queried for one or a few specific operating conditions that directly inform design or safety decisions. The errors corresponding to the best-case and worst-case trajectories can be widely different, and this is not known \emph{a priori}. A low mean error provides no guarantee that a particular instance of interest will not exhibit unacceptably large error. Moreover, classical numerical solvers, while accurate, offer no solution to this problem: they can be rerun for a given condition, but they provide no information about the reliability of a surrogate rollout, and they cannot monitor whether a neural operator is drifting or operating out of distribution during inference. This instance-aware reliability gap, namely the absence of any online, data-free mechanism to monitor and bound error growth for individual trajectories, is independent of solver cost and represents a fundamental barrier to the trustworthy deployment of neural surrogates in engineering practice.}

\revv{To be viable in real-world settings, neural operator surrogates must be not only fast but also instance-aware: capable of monitoring their own reliability and controlling error growth over long time horizons, without access to ground-truth solutions and without requiring the practitioner to know in advance whether a given operating condition lies within the training distribution. We identify and formalize the instance-aware reliability gap in neural operator deployment: the absence of any principled, online, data-free mechanism to monitor and bound error growth for individual surrogate trajectories. ANCHOR (\textbf{A}daptive \textbf{N}umerical \textbf{C}orrection for \textbf{H}igh-fidelity \textbf{O}perator \textbf{R}ollouts) is proposed as a direct solution to this gap, operating in the regime where a neural operator surrogate is already the method of choice and providing the reliability infrastructure that purely data-driven rollouts currently lack.} A schematic of our framework is shown in Figure~\ref{fig:hybrid_framework_schematic}. In ANCHOR, a neural operator serves as the primary time-marching engine, rapidly advancing the solution in time, while a numerical solver is invoked selectively as a corrective mechanism when accumulated error exceeds an acceptable threshold. While we instantiate ANCHOR using TI-DeepONet~\cite{nayak2026ti} \rev{for most PDE examples except the 2D Cahn-Hilliard \revv{and 2D Navier-Stokes cases}, where we use a time-integrator-embedded FNO surrogate inspired by TI-DeepONet}, the framework is model-agnostic and applicable to a broad class of time-dependent surrogate models. From a numerical perspective, ANCHOR can be interpreted as an adaptive time-marching scheme in which a learned operator is corrected on demand to control accumulated rollout error. \rev{In contrast to solver-in-the-loop~\cite{zhang2024blending, amin2025fenn, WANG2025118319} or end-to-end trained hybrid approaches~\cite{buitragobenefits, he2024sequential, hu5149007deepomamba, nayak2026ti}, which operate during training or rely on spatial decomposition strategies, ANCHOR introduces temporal coupling at inference time. This enables adaptive switching between the neural operator and numerical solver based on a physics-informed error estimator, without requiring retraining, backpropagation through numerical solvers, or access to ground-truth solutions during time marching. Crucially, this design makes ANCHOR robust to out-of-distribution test conditions in a way that purely data-driven surrogates cannot be: when a new parametric condition lies far from the training distribution, the error estimator detects the resulting residual growth and invokes the numerical solver more frequently, naturally allocating greater corrective effort precisely where the surrogate is least reliable. Conversely, for well-represented conditions, solver calls are infrequent, and the framework operates close to the speed of the neural operator alone. The user therefore does not need to know in advance whether a given operating condition is in- or out-of-distribution; ANCHOR adapts to each instance online and guarantees that the error remains bounded regardless. To the best of our knowledge, this paradigm of inference-time, instance-aware temporal coupling for hybrid neural-numerical frameworks has not been explored previously.}

A central challenge in such a coupling is determining when corrective intervention is required, particularly in the absence of ground-truth solutions during time marching. To address this, we introduce a physics-informed Exponential Moving Average (EMA)-based error estimator constructed from the normalized PDE residual. Unlike common practices that trigger solver intervention after a fixed number of surrogate iterations~\cite{oommen2022learning, ovadia2025real}, our estimator is evaluated online, requires no reference data, and adapts to the specific dynamics of each instance. We demonstrate, through both rigorous theoretical analysis and empirical validation, that this estimator correlates strongly with the true solution error, enabling data-free, adaptive switching between the neural operator and the numerical solver. This design explicitly leverages the complementary strengths of data-driven and physics-based methods: speed from the neural operator and reliability from the numerical solver. By incorporating online error monitoring and correction into the time marching pipeline, ANCHOR moves beyond average-case accuracy toward bounded-error, instance-aware prediction. 

The remainder of the manuscript is organized as follows. We validate the proposed approach on \revv{six} canonical PDEs, the 1D and 2D Burgers' equations, the 2D Allen-Cahn equation, \rev{the 2D Cahn-Hilliard equation,} \revv{the 2D Navier-Stokes equation,} and the 3D heat conduction equation, demonstrating that the hybrid framework significantly suppresses error growth over long temporal horizons while largely preserving the computational efficiency of neural operators. While ANCHOR is demonstrated here for dissipative systems, the underlying philosophy, online error estimation coupled with adaptive numerical correction, is broadly applicable to other classes of time-dependent PDEs given an appropriate design of physics-aware indicators that can represent the underlying predictive error growth. Extending the ANCHOR framework to other classes of PDEs, such as dispersive or chaotic systems, is conceptually straightforward but remains beyond the scope of this study.

\section{Methodology}
\label{sec:method}
We consider a general time-dependent PDE governing the spatiotemporal evolution of a physical system. Let $u(t, \mathbf{x})$ denote the solution over the temporal domain $t \in [0, T]$ and the spatial domain $\mathbf{x} \in \mathcal{X} \subseteq \mathbb{R}^n$, where $n$ is the number of spatial dimensions of the system, $n \in \lbrace 1,2,3 \rbrace$. The evolution dynamics of $u$ are governed by a PDE that relates the temporal derivative $u_t$ to spatial derivatives $u_{\mathbf{x}}, u_{\mathbf{xx}}, \dots$ via a general nonlinear function $\mathcal{N}$:
\begin{equation}
u_t = \mathcal{N}(t, \mathbf{x}, u, u_{\mathbf{x}}, u_{\mathbf{xx}}, \dots).
\label{eq:time-dep-PDE}
\end{equation}
Solutions to such PDEs are typically obtained through a two-stage process. First, the PDE is discretized in space using classical numerical methods such as finite difference, finite element, or spectral methods to approximate the right-hand side (RHS), i.e., the nonlinear function $\mathcal{N}(\cdot)$ defined in the equation above, which depends on the solution field and its spatial derivatives. This is followed by time integration using numerical schemes such as explicit or implicit Euler, Adams-Bashforth, or Runge-Kutta methods. Alternatively, deep learning-based approaches such as physics-informed neural networks (PINNs)~\cite{raissi2019physics} and neural operators can be used to approximate the solution operator directly, bypassing explicit discretization.

In this work, we focus on improving the reliability of data-driven models used for time-dependent PDEs in a Markovian, autoregressive setting, where the solution is advanced sequentially in time. For the instantiation of the proposed ANCHOR framework, we employ the Time-Integrator Deep Operator Network (TI-DeepONet) as the primary time-marching engine. The details of the TI-DeepONet architecture are presented in \cite{nayak2026ti}, where the authors have demonstrated that the framework significantly reduces error accumulation and improves extrapolation performance compared to standard autoregressive neural operators. While TI-DeepONet serves as a natural choice for demonstrating the ANCHOR framework due to its favorable long-horizon behavior, we emphasize that ANCHOR is model-agnostic. Any time-dependent surrogate capable of autoregressive time marching may be used as the candidate surrogate model. \rev{In a similar spirit, particularly for the 2D Cahn-Hilliard \revv{and 2D Navier-Stokes} equations, we implement a time-integrator embedded FNO (TI-FNO) to serve as the surrogate sequential neural operator, with TI-DeepONet being the method of choice for all other PDE examples.} Nevertheless, even with improved stability, purely data-driven rollouts remain susceptible to error accumulation over sufficiently long horizons, particularly when the system evolves beyond the training regime.

To address this limitation, we design a hybrid framework that selectively combines the speed of neural operator-based time marching with the robustness of classical numerical solvers. Specifically, we couple TI-DeepONet\rev{/TI-FNO} with a high-fidelity numerical solver, which is invoked adaptively as a corrective mechanism when surrogate-induced errors exceed an acceptable threshold. This coupling enables bounded error growth over time while preserving much of the computational efficiency of the learned surrogate. In the subsequent sections, we introduce a physics-informed error estimator that approximates accumulated error in the absence of ground-truth solutions, and we present the algorithm that governs adaptive switching between TI-DeepONet\rev{/TI-FNO} and the numerical solver. \rev{Note that from hereon we collectively refer to the pretrained surrogate models TI-DeepONet and TI-FNO as TI-based neural operator (TI-NO) for brevity.}
Although the high-fidelity solver used in this study is based on finite difference methods (FDM), the proposed framework is solver-agnostic. Both the neural surrogate and the numerical solver can be independently selected and coupled, allowing ANCHOR to be applied broadly across different PDE classes, discretization strategies, and learning architectures.

\subsection{Error Approximation using PDE Residual-based Exponential Moving Average}
\label{subsec:ema}
One of the main challenges in ANCHOR is determining the appropriate time to switch between methods. The relative $L_2$ error as defined in Eq.~\ref{eq:relative_l2_err} is a widely used metric for assessing solution accuracy. However, in most practical scenarios, the ground truth necessary to compute this error is unavailable. Furthermore, error accumulation is highly scenario-dependent. Consequently, a model may achieve a low ensemble-averaged error yet still incur unacceptable errors for individual operating conditions for an engineering application. To address this, we introduce a data-free instance-aware EMA-based (Exponential Moving Average) error estimator. The motivation for introducing an error estimator is to approximate the true error, which is typically unavailable or computationally infeasible to evaluate in practical applications. 
The generalized equation for the EMA is defined as: 
\begin{equation}
  \begin{aligned}
    s^{0} &= a y^{0},\\
    s^t &= a y^t + \left(1-a \right)s^{t-1},
    \label{eq:general_eq_ema}
\end{aligned}  
\end{equation}
where $s^{0}$ is the initial state, $y^0$ is the initial value of the field, and $s^t$ is a state variable that captures the history of how the local variable $y^t$ has evolved over previous time steps. In our problem, we consider the local variable to be the norm of the PDE residual normalized by the norm of the solution \rev{at any time step $t$, i.e., $\hat{r}^t$}, and the corresponding state variable is the ``error estimator,'' \rev{i.e., $\eta^t$,} defined as:
\begin{equation}
    \eta^t = a \hat{r}^t+\left(1-a \right)\eta^{t-1},
    \label{eqn:ema_error_estimator}
\end{equation}
where
\begin{align*}
    \hat{r}^t &= \frac{\norm{r^t}}{\norm{u^t}},\\
    r^t &: \text{PDE residual at time step $t$},\\
    u^t &: \text{Observed solution at time step $t$, and}\\
    a &: \text{Smoothing parameter ($a \in (0,1]$) controlling the weighting between $\hat{r}^t$ and $\eta^{t-1}$.}
\end{align*}
The proposed error estimator is inherently physics-informed, being computed from the PDE residual, and incorporates an EMA-based history of these residuals. It functions as an online, time-marching error predictor that operates without access to ground-truth reference data, leveraging the outputs of a pre-trained neural operator to estimate the error at each time step.

The choice of an EMA-based error estimator is motivated by the inherent error accumulation in autoregressive time-marching frameworks. At each time step, the neural network model's prediction is fed back as input for the subsequent step, causing errors to propagate and compound over time. Consequently, prediction errors tend to grow as the solution advances. By incorporating a history term, EMA retains information from past errors and provides a smoothing mechanism for capturing this cumulative error growth across successive forward passes. The use of a residual-based estimator is motivated by the fact that an exact solution of a PDE yields a zero residual. As a result, larger residuals correspond to larger deviations from the true solution. However, in dissipative systems, where the solution magnitude decays over time, the raw residual can be misleading, as diminishing solution amplitudes may artificially suppress residual values. To mitigate this issue, we employ a normalized PDE residual as the local error indicator, ensuring that the resulting estimator remains meaningful and comparable throughout the temporal evolution. In short, rather than relying on instantaneous residuals to compute the error at a given step, as is standard practice, the estimator tracks the history of normalized residuals using an EMA-based formulation. 
From a dynamical systems perspective, the EMA update can be interpreted as a first-order discrete relaxation equation driven by the instantaneous normalized residual, introducing inherent smoothing and stability that suppress spurious fluctuations while capturing sustained error growth over long time horizons.

\subsection{Adaptive Threshold for Error Estimator}
\label{subsec:adaptive_threshold}
In this study, we focus on dissipative systems, in which the magnitude of the solution field decreases over time. Since the EMA-based error estimator introduced above depends on the magnitude of the solution field, this decay causes the estimator to increase more slowly as the system evolves. As a result, the estimator may require a longer time to reach a prescribed threshold, even though the true prediction error has already grown significantly. Numerical experiments confirm that, under a fixed threshold, the estimator often lags behind the actual error growth in later stages of the temporal evolution. To address this issue, we introduce an adaptive thresholding strategy for the error estimator. Instead of using a fixed threshold throughout the simulation, we allow the error threshold to decay in time in accordance with the diminishing magnitude of the PDE solution. This ensures that, although the estimator’s tolerance is not fixed, the growth of the actual error remains bounded. 
This adaptive policy ensures that the sensitivity of the error estimator remains consistent throughout the simulation, enabling timely detection of error growth even as the solution amplitude decreases. The adaptive error thresholding policy is defined as: 
\begin{equation}
    \eta_{\text{thres}}^t = u_{0,max}e^{-\gamma t}e^{-u_{0,max}},
    \label{eqn:err_thres}
\end{equation}
where $u_{0,\max}$ denotes the maximum absolute value of the solution field at the initial condition, $t$ is the time step, and $\gamma$ is a tunable hyperparameter that controls the rate of decay of the threshold.

It is important to note that the proposed adaptive thresholding strategy is tailored to dissipative PDE systems, for which the solution magnitude decays over time. For PDEs exhibiting dispersive or chaotic dynamics, the thresholding policy can be modified accordingly by incorporating physics-aware priors that reflect the characteristic behavior of the underlying system; such extensions are beyond the scope of the present work. In summary, an effective error-control strategy necessitates a time-dependent, dynamics-aware threshold, rather than a fixed global value as is commonly adopted in standard practice.

\subsection{Adaptive Numerical Correction for High-fidelity Operator Rollouts}
To reiterate, the core idea of this work is to leverage the accuracy of high-fidelity numerical solvers (FDM in our case) while retaining the computational efficiency of \rev{a pretrained neural operator}. To this end, we introduce an EMA-based error estimator, $\eta^t$, which mimics the growth of the solution error as \rev{TI-NO} advances the PDE in time (see Eq.~\ref{eqn:ema_error_estimator}). To quantify the solution error in question, we consider a relative $L_2$ error metric, $\varepsilon_{L_2}$, which is defined as:
\begin{equation}
    \varepsilon_{L_2} = \dfrac{\norm{u_{\text{truth}} - u_{\text{pred}} }}{\norm{u_{\text{truth}}}}, 
    \label{eq:relative_l2_err}
\end{equation}
where $u_{\text{truth}}$ is the solution generated using the numerical solver and $u_{\text{pred}}$ is the solution predicted/obtained during the neural operator surrogate or ANCHOR. Figure~\ref{fig:hybrid_framework_schematic} illustrates the proposed ANCHOR framework, which couples \rev{TI-NO} with a high-fidelity numerical solver. The time marching procedure proceeds as follows. Starting from a prescribed initial condition, the solution is advanced in time using \rev{TI-NO} as the primary inference engine. During the course of the simulation, the algorithm continuously monitors whether $\eta^t$ exceeds a prescribed, time-dependent threshold $\eta_{\text{thres}}^t$. Based on this comparison, a decision is made at each time step regarding whether to continue with \rev{the TI-NO surrogate} or to switch to the numerical solver. If $\eta^t > \eta_{\text{thres}}^t$, the responsibility for time advancement is transferred to the high-fidelity numerical solver, which evolves the PDE for a fixed number of time steps. After this corrective phase, control is handed back to \rev{the TI-NO surrogate}. This adaptive switching procedure is repeated until the final simulation time is reached, thereby combining the computational efficiency of \rev{TI-NO} with the robustness, reliability, and accuracy of classical numerical solvers.

\section{Results}
\label{sec:results}
As discussed, the two key components of our work are: (1) the introduction of an EMA-based error estimator that correlates strongly with the actual error growth during \rev{TI-NO} inference; and (2) establishing the coupling between neural surrogate (\rev{TI-NO}, in our case) and a classical numerical solver (FDM, in our case). To evaluate the efficacy of our framework, we consider \revv{six} PDEs: (1) 1D Burgers', (2) 2D Burgers', (3) 2D Allen-Cahn, \rev{(4) 2D Cahn-Hilliard,} \revv{(5) 2D Navier-Stokes,} and (\rev{6}) 3D heat conduction. Typically, in literature, surrogate model accuracy is reported on an ensemble basis, with error metrics averaged across multiple samples. In practice, however, due to generalization effects, errors can vary significantly across individual realizations. Therefore, we analyze model accuracy of ANCHOR primarily on a per-sample basis, during time marching. The code to reproduce the experiments is publicly available at \url{https://github.com/Centrum-IntelliPhysics/ANCHOR.git}. 

\begin{figure}[htpb]
    \centering
    \includegraphics[width=0.99\linewidth]{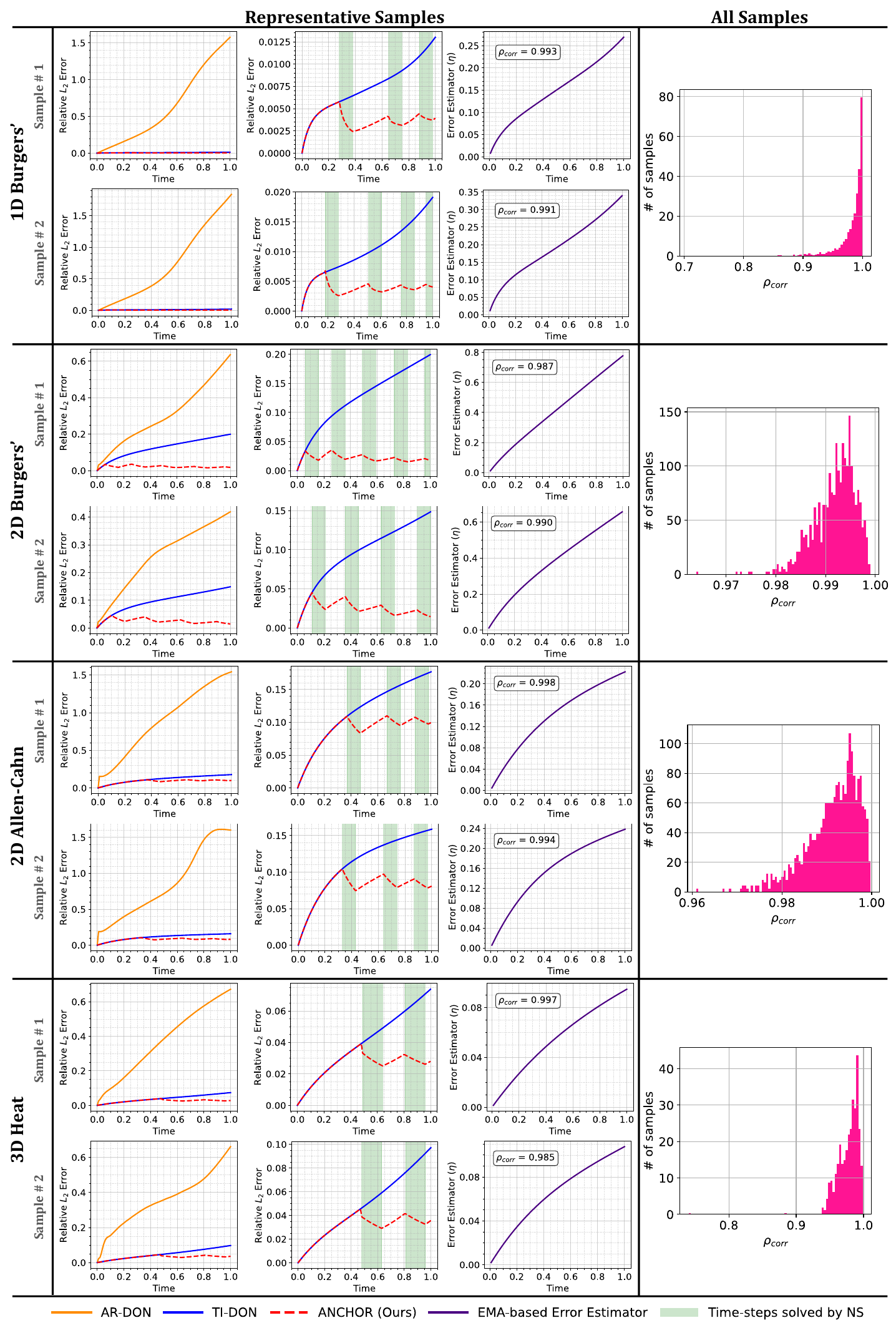}
    \caption{Results for two representative samples from each PDE that uses TI-DeepONet as the NO surrogate. Column 1: Temporal evolution of relative $L_2$ error across different frameworks for a representative sample. Column 2: Stabilization of relative $L_2$ error growth achieved by intermittently invoking the numerical solver for the same samples; green shaded regions indicate time steps handled by the high-fidelity numerical solver. Column 3: EMA-based error estimator behavior over time, with the corresponding Pearson correlation coefficient ($\rho_{corr}$) between the estimator and the underlying relative $L_2$ error mentioned in the boxes. Column 4: Distribution of $\rho_{corr}$ across all test samples.}
    \label{fig:corr_with_err_estm_and_l2_errors}
\end{figure}

\begin{figure}[htpb]
    \centering
    \includegraphics[width=0.99\linewidth]{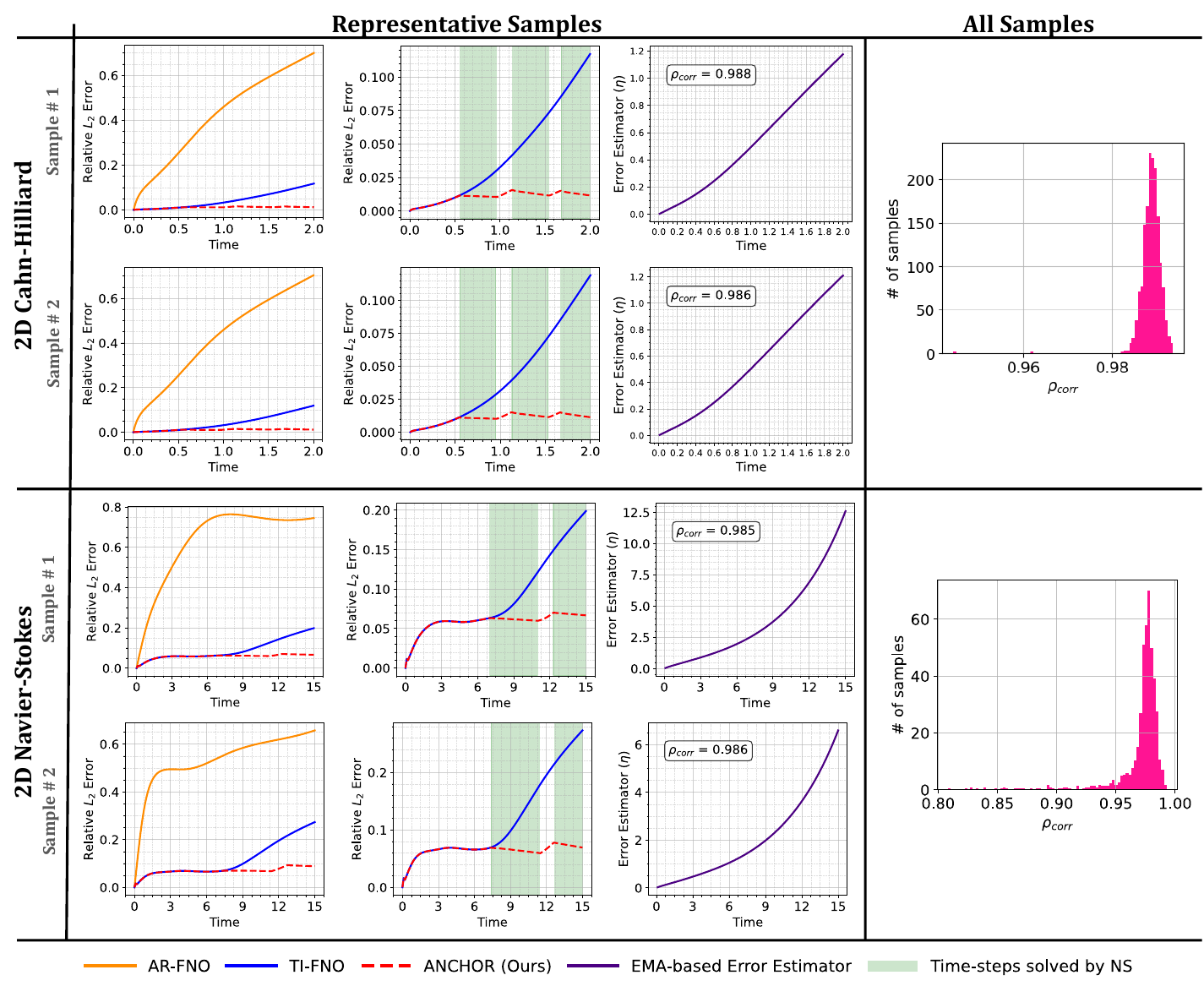}
    \caption{\revv{For two representative samples, each of the 2D Cahn-Hilliard equation and 2D Navier-Stokes equation considered in this work. Column 1: Temporal evolution of relative $L_2$ error across different frameworks. Column 2: Stabilization of relative $L_2$ error growth achieved by intermittently invoking the numerical solver; green shaded regions indicate time steps handled by the high-fidelity numerical solver. Column 3: EMA-based error estimator behavior over time, with the corresponding Pearson correlation coefficient ($\rho_{corr}$) between the estimator and the true relative $L_2$ error shown in the boxes. Column 4: Distribution of $\rho_{corr}$ across all test samples. Unlike other examples in this work where TI-DON is used, TI-FNO serves as the NO surrogate for the 2D Cahn-Hilliard and the 2D Navier-Stokes equations.}}
    \label{fig:corr_with_err_estm_and_l2_errors_fno}
\end{figure}

\begin{table}[htb!]
    \centering
    \renewcommand{\arraystretch}{1.15}
    \caption{Summary of train and test data splits, inference time step $\Delta t$, and relative $L_2$ test errors for the TI-DeepONet\rev{/TI-FNO} framework, for all PDE examples. Errors are reported in an ensemble-averaged sense computed over all test samples across all time steps.}
    \label{tab:tidon_fno_train_test_details}
    \begin{tabular}{|c|c|c|c|c|c|c|c|}
\hline
\multirow{2}{*}{PDE Example} & \multirow{2}{*}{Method}
& \multirow{2}{*}{$N_{\text{train}}$} & \multirow{2}{*}{$N_{\text{test}}$} & \multicolumn{2}{c|}{$t_{\text{test}}$}
& \multirow{2}{*}{$\Delta t$} 
& \multirow{2}{*}{$\varepsilon_{L_2} (\text{test})$} \\
\cline{5-6}
& & & & $t_{\text{interpolation}}$ & $t_{\text{extrapolation}}$ & & \\
\hline
1D Burgers' & TI-DON     
& 2000 & 500 & [0,0.5] & [0.5,1.0]  & 0.01  & 0.0179 \\
\hline
2D Burgers' & TI-DON     
& 1000 & 250 & [0,0.33] & [0.33,1.0] & 0.01 & 0.1299 \\
\hline
2D Allen-Cahn & TI-DON 
& 1000 & 250 & [0,0.33] & [0.33,1.0] & 0.01  & 0.1616 \\
\hline
\rev{2D Cahn-Hilliard} & \rev{TI-FNO} 
& \rev{1000} & \rev{250} & \rev{[0,0.67]} & \rev{[0.67,1.0]} & \rev{0.01}  & \rev{0.0581} \\
\hline
\revv{2D Navier-Stokes} & \revv{TI-FNO} 
& \revv{800} & \revv{200} & \revv{[0,7.5]} & \revv{[7.5,15.0]} & \revv{0.1}  & \revv{0.0643} \\
\hline
3D Heat & TI-DON  
& 800 & 200 & [0,0.33] & [0.33,1.0] & 0.01  & 0.0655 \\
\hline
\end{tabular}
\end{table}
Table~\ref{tab:tidon_fno_train_test_details} presents the details of the training of the \rev{TI-based NO variants (TI-DON/TI-FNO)}, including the number of training and testing samples along with the mean relative $L_2$ errors computed in an ensemble-averaged sense over all the test samples. Figure~\ref{fig:corr_with_err_estm_and_l2_errors} (first column) illustrates the relative $L_2$ error growth obtained using autoregressive DeepONet (AR-DON), TI-DeepONet (TI-DON), and the proposed ANCHOR framework, for a representative sample, for all example PDEs considered in this study, \revv{except for the 2D Cahn-Hilliard and the 2D Navier-Stokes cases}. The second column additionally highlights the time regions where the numerical solver is invoked to correct the system dynamics. The third column presents the EMA-based error estimator used to predict error growth in the absence of ground-truth solutions, along with the Pearson correlation coefficient $(\rho_{corr})$ between the proposed error estimator and the actual error growth for the same sample. Finally, the fourth column shows a histogram that quantifies the correlation between the estimated error and the actual error, using Pearson's correlation coefficient computed across all test samples. \rev{Columns in Fig.~\ref{fig:corr_with_err_estm_and_l2_errors_fno} exhibit similar behavior to those in Fig.~\ref{fig:corr_with_err_estm_and_l2_errors}, the only difference being that the baseline neural operator surrogate considered is TI-FNO.}

\begin{table}[htb!]
    \centering
    \renewcommand{\arraystretch}{1.075}
    \caption{Wall-clock computation times for numerical solver (NS), \revv{TI-NO (TI-DeepONet/TI-FNO),} and the proposed ANCHOR framework across different PDE examples for two representative test samples.}
    \begin{tabular}{|c|c|c|c|c|c|}
        \hline
         \multirow{2}{*}{PDE Example} & \multirow{2}{*}{Sample \#}  & \multicolumn{4}{c|}{Computation Time (in secs)}    \\
         
         \cline{3-6}
         & & NS only & \revv{TI-NO only} & ANCHOR (Ours) & \revv{Error Estimator}\\ 
         \hline
         \multirow{2}{*}{1D Burgers'} & \#1 & 0.15 & 0.049 & 0.22 &  \revv{0.00739}\\
         \cline{2-6}
                                      & \#2 & 0.156 & 0.0714 & 0.163 & \revv{0.0088}\\
         \hline
         \multirow{2}{*}{2D Burgers'} & \#1 & 7.86 & 0.139 & 4.842 &  \revv{0.012}\\
         \cline{2-6}
                                      & \#2 & 7.915 & 0.136 & 4.089 & \revv{0.013}\\
         \hline
         \multirow{2}{*}{2D Allen-Cahn} & \#1 & 3.22 & 0.075 & 1.04 &  \revv{0.021}\\
         \cline{2-6}
                                        & \#2 & 3.13 & 0.076 & 0.99 & \revv{0.022}\\
         \hline
         \multirow{2}{*}{\rev{2D Cahn-Hilliard}} & \rev{\#1} & \rev{12.436} & \rev{0.375} & \rev{8.870} &  \revv{0.0281}\\
         \cline{2-6}
                                  &\rev{\#2} & \rev{12.502} & \rev{0.373} & \rev{8.11} & \revv{0.0244}\\
         \hline
         \multirow{2}{*}{\revv{2D Navier-Stokes}} & \revv{\#1} & \revv{1487.56} & \revv{0.896} & \revv{816.55} &  \revv{0.0983}\\
         \cline{2-6}
                                  &\revv{\#2} & \revv{1466.85} & \revv{0.857} & \revv{826.44} & \revv{0.0973}\\
         \hline
        
         \multirow{2}{*}{3D Heat} & \#1 & 2.735 & 0.178 & 1.084 & \revv{0.025}\\
         \cline{2-6}
                                  &\#2 & 2.677 & 0.176 & 1.017 & \revv{0.0246}\\
        \hline
    \end{tabular}
    \label{tab:placeholder}
\end{table}

\subsection{One-dimensional Burgers' Equation}
\label{subsec:example1}
The first example we consider is the one-dimensional viscous Burgers’ equation, a canonical PDE arising in fluid mechanics, nonlinear acoustics, and traffic flow. For the velocity field $u(x, t)$ with viscosity $\nu = 0.01$, it is defined as:
\begin{equation}
\frac{\partial u}{\partial t} + u \frac{\partial u}{\partial x} = \nu \frac{\partial^2 u}{\partial x^2}, \quad (x,t) \in [0,1] \times [0,1],
\end{equation}
subject to periodic boundary and initial conditions, $u(x, 0) = u_0(x)$. The initial condition $u_0(x)$ is sampled from a Gaussian random field with spectral density $S(k) = \sigma^2(\tau^2 + (2\pi k)^2)^{-\gamma}$, where $\sigma = 25$, $\tau = 5$, and $\gamma = 4$, ensuring periodicity on $x \in [0, 1]$. The corresponding kernel is given by the inverse Fourier transform, $K(\mathbf{x, x'}) = \int_{-\infty}^{\infty} S(k) e^{2 \pi i k (\mathbf{x} - \mathbf{x'})} dk$. We discretize the spatiotemporal domain using 101 grid points along each dimension. The system is simulated with a Fourier pseudospectral solver implemented via the \texttt{spin} framework in the \emph{Chebfun}~\cite{chebfun} package for MATLAB. The viscous Burgers' equation can be written in operator-splitting form as:
\begin{equation}
\frac{\partial u}{\partial t} = \mathcal{L}(u) + \mathcal{N}(u),
\end{equation}
with the linear and nonlinear operators defined as:
\begin{equation}
\mathcal{L}(u) = \nu \frac{\partial^2 u}{\partial x^2}, 
\qquad
\mathcal{N}(u) = -\frac{1}{2}\frac{\partial}{\partial x}\left(u^2\right).
\end{equation}

Spatial derivatives are computed spectrally on a uniform grid of $s=101$ points with periodic boundary conditions, and time integration is performed using the fourth-order exponential time-differencing Runge-Kutta (ETDRK4) scheme~\cite{cox2002exponential} provided by \texttt{spin}. For ease of integration with the neural operator inference code, we adapt the MATLAB solver implementation to Python, as most of our neural operator code is written in JAX~\cite{jax2018github}. In this work, we do not focus on general interoperability, i.e., handling numerical solvers implemented in different frameworks, but rather on demonstrating the effectiveness of the coupling concept pivotal to the ANCHOR framework. To reiterate, the primary goal of our work is to employ the coupled framework to simulate the system accurately and reliably with bounded error growth over long temporal horizons, particularly beyond the training time domain. For this study, TI-DeepONet was trained on $t \in [0, 0.5]$, and during time marching, the system is simulated until $t = 1.0$ using three approaches: (1) autoregressive DeepONet (AR-DON), (2) TI-DeepONet (TI-DON), and (3) the coupled TI-DeepONet-numerical solver (ANCHOR) framework.

The top row of Fig.~\ref{fig:corr_with_err_estm_and_l2_errors} depicts, for a representative sample, in order: (1) the relative $L_2$ error growth over time, (2) the stabilization of TI-DON error growth by intermittently invoking the numerical solver when the EMA-based error estimator $\eta^t$ exceeds the adaptive threshold, (3) the EMA-based error estimator ($\eta^t$) trend and the Pearson correlation coefficient $(\rho_{corr})$ between $\eta^t$ and the true relative $L_2$ solution error, and (4) a histogram of the correlation coefficients across all test samples. The error estimator tuned using $a = 0.1$ exhibits a high Pearson correlation of $\rho_{corr} = 0.991$ with the underlying relative $L_2$ error, closely emulating its trend and confirming its reliability for defining the cutoff threshold for switching between the numerical solver and TI-DON. This is further reinforced by the distribution of $\rho_{corr}$, where it can be observed that the correlation is consistently above 0.95 across all test samples. The decay factor in the adaptive threshold is set to $\gamma = 2$. 

\begin{figure}[!htbp]
    \centering
    \includegraphics[width=0.9\linewidth]{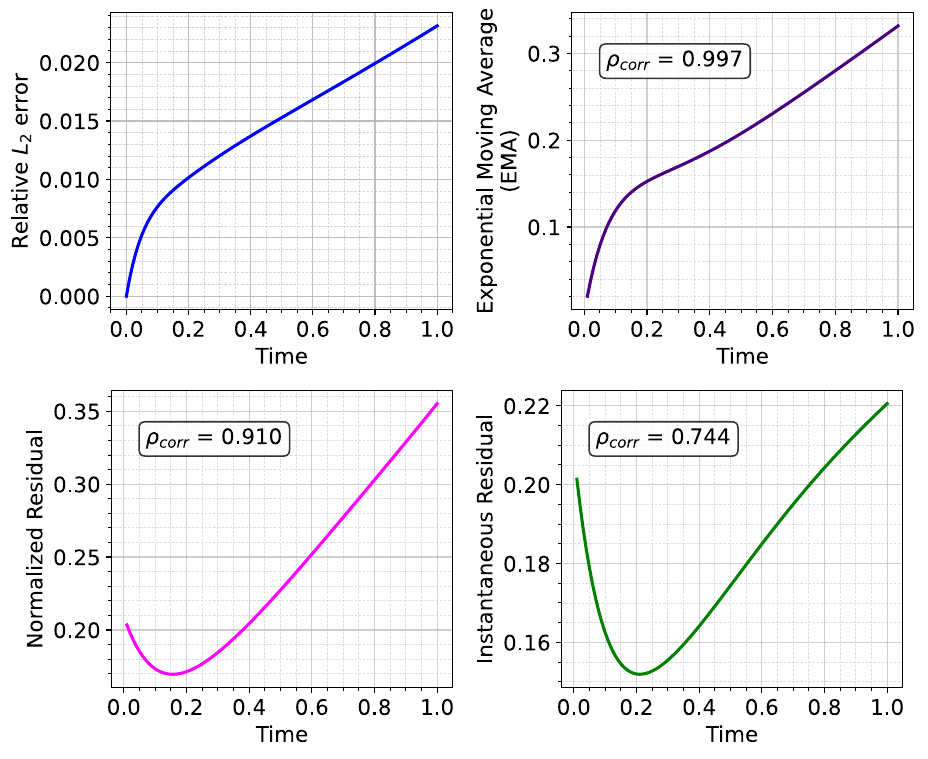}
    \caption{\revv{1D Burgers' equation: Comparison of Pearson correlation coefficients ($\rho_{corr}$) between the true relative $L_2$ solution error and three error estimators. Top left: true error evolution; top right: EMA-based estimator; bottom left: normalized residual; bottom right: instantaneous residual. Among the three formulations, the EMA-based error estimator exhibits the highest correlation and most closely tracks the temporal evolution of the true relative $L_2$ solution error.}}
    \label{fig:comparison_different_err_estimators}
\end{figure}

\revv{\textbf{Rationale for Using an EMA-Based Error Estimator to Capture Autoregressive Error Growth.}}
\revv{In autoregressive prediction, the solution at each time step serves as input to the next, allowing small local errors to propagate and accumulate. This history-dependent error growth calls for an estimator that retains memory of past behavior rather than relying solely on instantaneous quantities. The EMA-based error estimator fulfills this requirement by smoothing short-term residual fluctuations while preserving information about prior residual evolution, yielding a temporally consistent measure of accumulated error.} \revv{As shown in Fig.~\ref{fig:comparison_different_err_estimators}, the true relative $L_2$ solution error exhibits sustained monotonic growth. The EMA-based error estimator captures this trend closely, achieving a correlation of $\rho_{corr}=0.997$. In contrast, the instantaneous and normalized residuals are more local in time and initially decrease before increasing, failing to reflect early-time error accumulation; their correlations are $\rho_{corr}=0.744$ and $\rho_{corr}=0.910$, respectively. This makes the EMA-based estimator a more reliable criterion for detecting sustained error propagation and guiding adaptive solver switching in hybrid neural-numerical frameworks. Beyond this empirical evidence, we proceed to establishing a formal structural equivalence between the EMA-based error estimator and the underlying relative $L_2$ error in the subsequent section.}

\textbf{\revv{Formal Analysis for EMA-Based Error Estimator.}} \revv{The choice of an exponential moving average (EMA) for the error estimator is motivated by its structural equivalence to the mean-free error recurrence derived in Sec.~\ref{sec:theoretical_analysis_error_bound}. We establish this connection more concretely below.}

\revv{
Recall the mean-free error recurrence from Eq.~\ref{eq:error_with_residual}:
}
\begin{equation}
\revv{
    e_{n+1}' \leq \rho \, e_n' + \Delta t \cdot \|R_n'\|,
}
    \label{eq:error_recurrence_recall}
\end{equation}
\revv{
where $\rho = 1 - \lambda \Delta t$ with $0 < \rho < 1$. The EMA-based error estimator defined in Eq.~\ref{eqn:ema_error_estimator} satisfies:
}
\begin{equation}
\revv{
    \eta^{n+1} = (1 - a)\eta^{n} + a \hat{r}^{n+1},
}
    \label{eq:EMA_recall}
\end{equation}
\revv{
where $\hat{r}^n = \|r^n\|/\|u^n\|$ is the normalized residual at time step $n$, and $a \in (0,1]$ is the smoothing parameter.
}

\revv{
\textit{Proposition (Structural Equivalence):}
Both the mean-free error $e_n'$ and the EMA estimator $\eta^n$ satisfy first-order linear recurrence relations of the form:
}
\begin{equation}
\revv{
    X^{n+1} = \gamma X^{n} + \sigma F^{n},
}
    \label{eq:general_recurrence}
\end{equation}
\revv{
where $\gamma \in (0,1)$ is a contraction factor, $\sigma > 0$ is a scaling coefficient, and $F^n \geq 0$ is a forcing term derived from the residual. Specifically:
}
\begin{itemize}[leftmargin=*,nosep]
    \item \revv{For the mean-free error: $\gamma = \rho = 1 - \lambda \Delta t$, $\sigma = \Delta t$, and $F^n = \|R_n'\|$.}
    \item \revv{For the EMA estimator: $\gamma = 1 - a$, $\sigma = a$, and $F^n = \hat{r}^{n+1}$.}
\end{itemize}
\revv{
Note that the forcing terms $F^n$ are evaluated at slightly different points in the time-stepping cycle for the two recurrences, but this does not affect the structural equivalence or the steady-state analysis.
}

\revv{
\textit{Steady-State Analysis:}
For a first-order linear recurrence of the form in Eq.~\ref{eq:general_recurrence} with constant forcing $F^n = F$, the steady-state solution satisfies:
}
\begin{equation}
\revv{
    X_\infty = \gamma X_\infty + \sigma F \quad \Rightarrow \quad X_\infty = \frac{\sigma F}{1 - \gamma}.
}
    \label{eq:steady_state_general}
\end{equation}

\revv{
Applying this to both recurrences:
}
\begin{itemize}[leftmargin=*,nosep]
    \item \revv{For the mean-free error with constant forcing $\|R'\| = \epsilon_S'$:
    \begin{equation}
        e_\infty' = \frac{\Delta t \cdot \epsilon_S'}{1 - \rho} = \frac{\Delta t \cdot \epsilon_S'}{\lambda \Delta t} = \frac{\epsilon_S'}{\lambda}.
        \label{eq:error_steady_state_recall}
    \end{equation}}
    \item \revv{For the EMA with constant time-invariant forcing $\hat{r}$:
    \begin{equation}
        \eta_\infty = \frac{a \hat{r}}{1 - (1-a)} = \hat{r}.
        \label{eq:EMA_steady_state}
    \end{equation}}
\end{itemize}

\revv{
\textit{Corollary (Error-EMA Relationship):}
At steady state, the mean-free error and EMA estimator are related by:
}
\begin{equation}
\revv{
    e_\infty' = \frac{\epsilon_S'}{\lambda} \quad \text{and} \quad \eta_\infty = \hat{r} \approx \epsilon_S',
}
\end{equation}
\revv{
where the approximation $\hat{r} \approx \epsilon_S'$ holds when the solution norm $\|u^n\|$ is $O(1)$. Therefore:
}
\begin{equation}
\revv{
    \boxed{e_\infty' \approx \frac{\eta_\infty}{\lambda}}
}
    \label{eq:error_EMA_steady_state}
\end{equation}

\revv{
This relationship motivates a similar interpretation in the transient regime. When the EMA reaches the threshold $\tau_n$, the corresponding mean-free error is bounded by:
}
\begin{equation}
\revv{
    \boxed{e_n' \lesssim \frac{\tau_n}{\lambda}}
}
    \label{eq:error_threshold_relation}
\end{equation}

\revv{
\textit{Interpretation:}
The structural equivalence establishes that the EMA estimator $\eta^n$ serves as a computationally tractable proxy for the mean-free error $e_n'$. Both quantities satisfy the same class of first-order contractive recurrences driven by the residual, and their steady-state values are related by the dissipation rate $\lambda$. Consequently, controlling the EMA through the threshold $\tau_n$ provides a practical mechanism for controlling the accumulated mean-free error.
}

\revv{
\textbf{Remark:}
This equivalence provides a principled justification for using EMA-based switching rather than instantaneous residual thresholding. While simpler rules based on instantaneous residuals (i.e., switching when $\hat{r}^n \geq \tau_n$) could be employed, such approaches are susceptible to high-frequency fluctuations and may trigger frequent, unnecessary switching between surrogate and solver phases. The EMA acts as a low-pass filter that tracks the accumulated error behavior, which, as shown above, is the quantity governing solution accuracy.
}

\rev{\textbf{Importance of Adaptive Threshold over Fixed PDE Residual Threshold.}} 
\rev{As discussed in Sec.~\ref{subsec:adaptive_threshold}, in dissipative systems the characteristic scale of the solution and its dynamics evolve, which can make a fixed trigger threshold poorly calibrated at later times. In particular, we observed that using a constant threshold can delay or suppress numerical-solver fallback even when the rollout error continues to grow in the extrapolation regime.}

\begin{figure}[!htbp]
    \centering
    \includegraphics[width=0.99\linewidth]{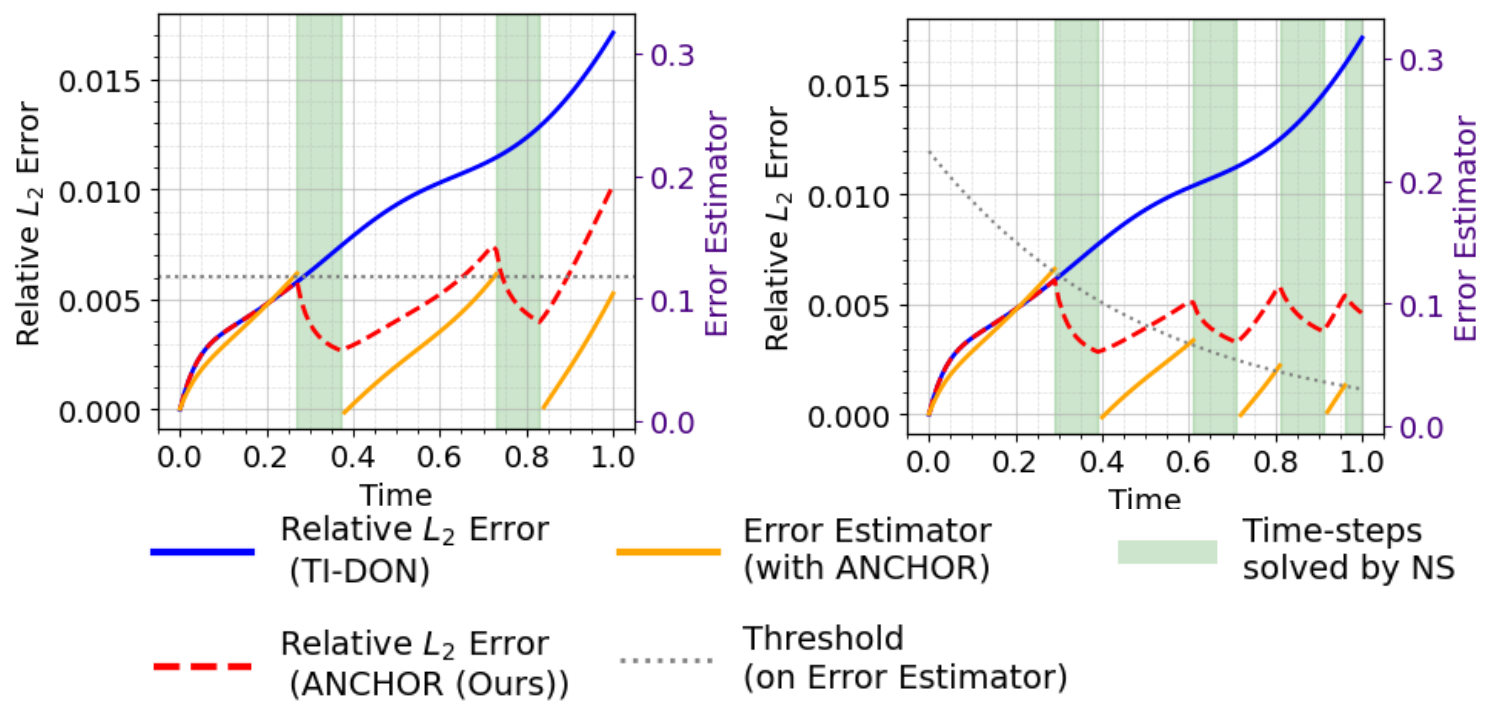}
    \caption{\rev{1D viscous Burgers' equation: Effect of thresholding policy for the EMA-based estimator. \textbf{Left}: fixed threshold. \textbf{Right}: adaptive threshold with $\gamma=2$. Curves show the true relative $L_2$ error, the estimator $\eta_t$, and the threshold $\eta_{\text{thres}}$; shaded regions indicate solver fallback intervals.}}
\label{fig:burger_adapt_compare}
\end{figure}

\rev{To demonstrate this effect, Fig.~\ref{fig:burger_adapt_compare} compares ANCHOR on the 1D viscous Burgers' test case using (left) a fixed threshold and (right) the proposed time-varying adaptive threshold. In each panel we plot the true relative $L_2$ error, the EMA-based estimator $\eta_t$, and the corresponding threshold $\eta_{\text{thres}}$; shaded regions indicate time intervals where the numerical solver is invoked. For a controlled comparison, the fixed threshold is set to the initial value of the adaptive policy, i.e., $\eta_{\text{thres}}=\eta_{\text{thres}}(t{=}0) = 0.12$. The fixed-threshold policy leads to delayed/insufficient triggering in the late-time regime, allowing larger peak errors, whereas the adaptive policy maintains sensitivity and triggers timely corrections, resulting in bounded error growth.}

\begin{figure}
    \centering
    \includegraphics[width=\linewidth]{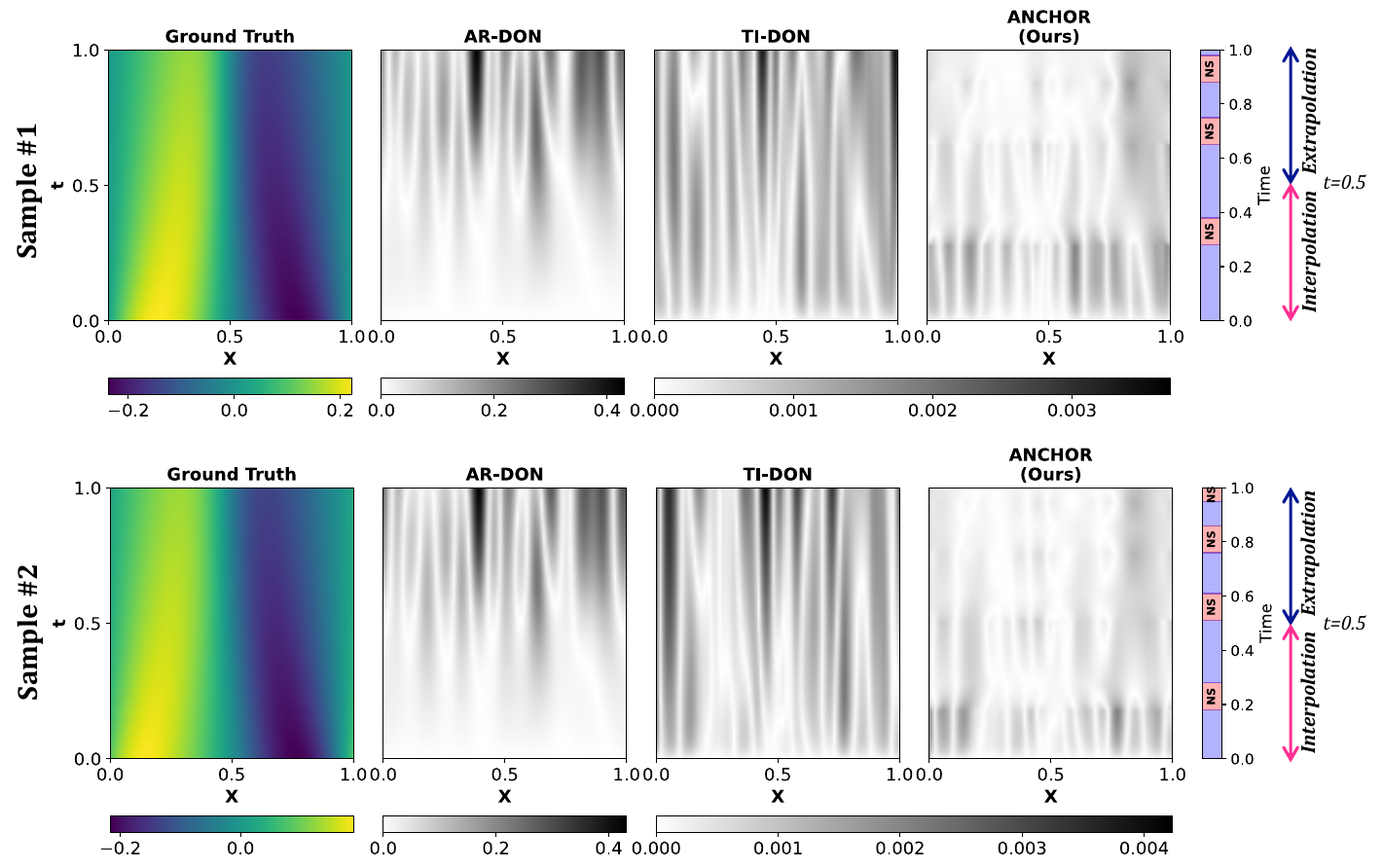}
    \caption{1D Burgers’ equation: Spatiotemporal error distributions over $t \in [0,1]$ for all frameworks, illustrated using two representative samples. The error color bar for AR-DON is shown separately from those of TI-DON and ANCHOR, since the errors of TI-DON and ANCHOR are approximately two orders of magnitude lower than those of AR-DON. The color bar on the far right indicates the time steps solved by TI-DON (blue) and by the high-fidelity numerical solver (pink). Here, $t \in [0,0.5]$ corresponds to the interpolation regime, while $t \in [0.5,1.0]$ denotes the extrapolation regime.}
    \label{fig:1d_burgers_error_contours}
\end{figure}

Upon analyzing the error accumulation plots, it is evident that AR-DON exhibits exponentially increasing error, as expected, due to its high susceptibility to compounding of model approximation errors. Consistent with the findings in~\cite{nayak2026ti}, TI-DON effectively mitigates this growth but does not fully bound it. In contrast, the ANCHOR framework leverages intermittent solver corrections guided by the error estimator to successfully bound the error. Notably, the first solver invocation results in an immediate drop in the relative $L_2$ error, clearly reflecting the corrective effect of the numerical solver. We hypothesize that this behavior arises from the solver’s ability to accurately and explicitly represent the right-hand side (RHS) of the governing PDE, in contrast to the approximate RHS representation learned by TI-DON. Since the RHS encodes the underlying spatial dynamics at a given time step, its exact evaluation enables a direct correction of accumulated model approximation errors, which are the primary contributors to error growth during recursive inference. This corrective behavior is consistently observed across other PDE systems and will be discussed further in the subsequent sections. After each solver intervention (e.g., 10 time steps), control returns to TI-DON until the error estimator again exceeds the running threshold. Intuitively, because the threshold decays over time (see Eq.~\ref{eqn:err_thres}), subsequent solver calls occur at lower error levels, effectively capping the maximum error - typically determined by the first solver call during time marching. This maximum error defines an envelope that bounds the subsequent temporal error incurred by ANCHOR.
Figure~\ref{fig:1d_burgers_error_contours} compares spatiotemporal errors across all frameworks. Qualitatively, for $t > 0.5$ (the extrapolation domain), the errors incurred by TI-DON are corrected and effectively curtailed by the numerical solver, which is in excellent agreement with the trends observed in the aforementioned analyses. Since the 1D Burgers' equation is a low-dimensional case and relatively inexpensive to solve classically, this example is primarily used to validate the accuracy of the proposed method; consequently, no noticeable computational speedup is observed compared to the standalone numerical solver.

\begin{figure}[!htbp]
    \centering
    \includegraphics[width=0.99\linewidth]{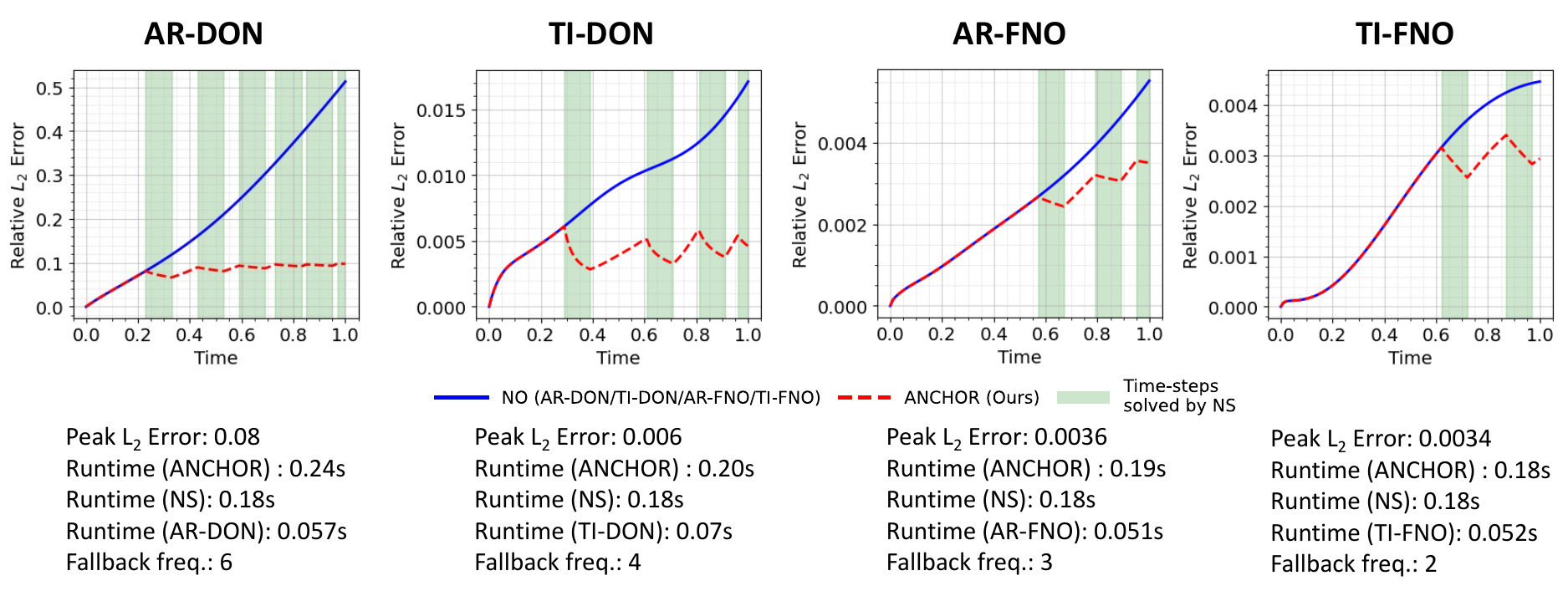}
    \caption{\rev{1D Burgers' equation: Comparison of relative $L_2$ error growth across different neural operator architectures. Each subplot shows the error evolution for a given surrogate (AR-DON, TI-DON, AR-FNO, TI-FNO) operating either as a pure surrogate or within the ANCHOR framework. TI-based neural operators (TI-DON, TI-FNO) maintain well-bounded error over time, whereas purely autoregressive neural operators (AR-DON, AR-FNO) exhibit gradual error accumulation in long-horizon predictions that are harder to bound.}}
    \label{fig:diff_no_1d_burgers}
\end{figure}

\rev{Although the primary implementation of this study uses TI-DON as the neural operator, the approach readily extends to other classes of autoregressive neural operators. To support this claim, we present a consolidated set of $L_2$ error growth plots (Fig.~\ref{fig:diff_no_1d_burgers}) for AR-DON, AR-FNO, and TI-FNO alongside TI-DON, either operating as a pure surrogate or within the ANCHOR framework with numerical solver calls. The results indicate that time-integrated NOs, such as TI-DON and TI-FNO, maintain well-bounded error over time. In contrast, purely autoregressive neural operators (e.g., AR-DON and AR-FNO), while achieving low short-term error, tend to exhibit gradual error accumulation in long-horizon predictions, and bounding long-term errors becomes difficult. This is not the case for TI-based NOs (see Sec.~\ref{sec:theoretical_analysis_error_bound} for why incorporating time-stepping within the neural operator supports the numerical solver in reliable error boundedness). Since the 1D Burgers' equation admits a fast numerical solver, we do not observe any gain in computational efficiency for this example. Additionally, across all methods explored, the runtime for ANCHOR depends on the number of solver calls.}

\subsection{Two-dimensional Burgers' Equation}
\label{subsec:example2}
A natural extension of the viscous one-dimensional Burgers' equation is its two-dimensional counterpart. To assess performance on high-dimensional spatiotemporal dynamics, we consider the 2D Burgers' equation modeling a scalar field $u(x,y,t)$, defined as:
\begin{equation}
    \frac{\partial u}{\partial t} 
    + u \frac{\partial u}{\partial x} 
    + u \frac{\partial u}{\partial y} 
    = \nu \left( \frac{\partial^2 u}{\partial x^2} + \frac{\partial^2 u}{\partial y^2} \right),
    \quad \forall \ (x,y,t) \in [0,1]^2 \times [0,1],
    \label{eq:2d_burgers}
\end{equation}
where $\nu = 0.01$ denotes the kinematic viscosity. The initial condition $u(x,y,0) = s(x,y)$ is sampled from two-dimensional periodic Mat\'ern-type Gaussian random fields with length scale $l = 0.125$ and standard deviation $\sigma = 0.15$. Periodic boundary conditions are imposed in both spatial directions, enforcing periodicity of the solution and its first-order spatial derivatives. A numerical solution procedure similar to that used for the 1D Burgers' equation is employed, following a pseudospectral approach. Spatial derivatives are computed in spectral space, and time integration is performed using the ETDRK4 scheme with a time step of $\Delta t = 10^{-4}$ up to a final time of $t = 1.0$. Solution snapshots are saved at intervals of $\Delta t_{\text{save}} = 0.01$. This numerical solver serves as the baseline high-fidelity model coupled with TI-DeepONet within the proposed ANCHOR framework. For further details on the data generation procedure, the reader is referred to \cite{rosofsky2023applications}, from which our implementation is largely adapted with appropriate modifications.
\begin{figure}
    \centering
    \includegraphics[width=\linewidth]{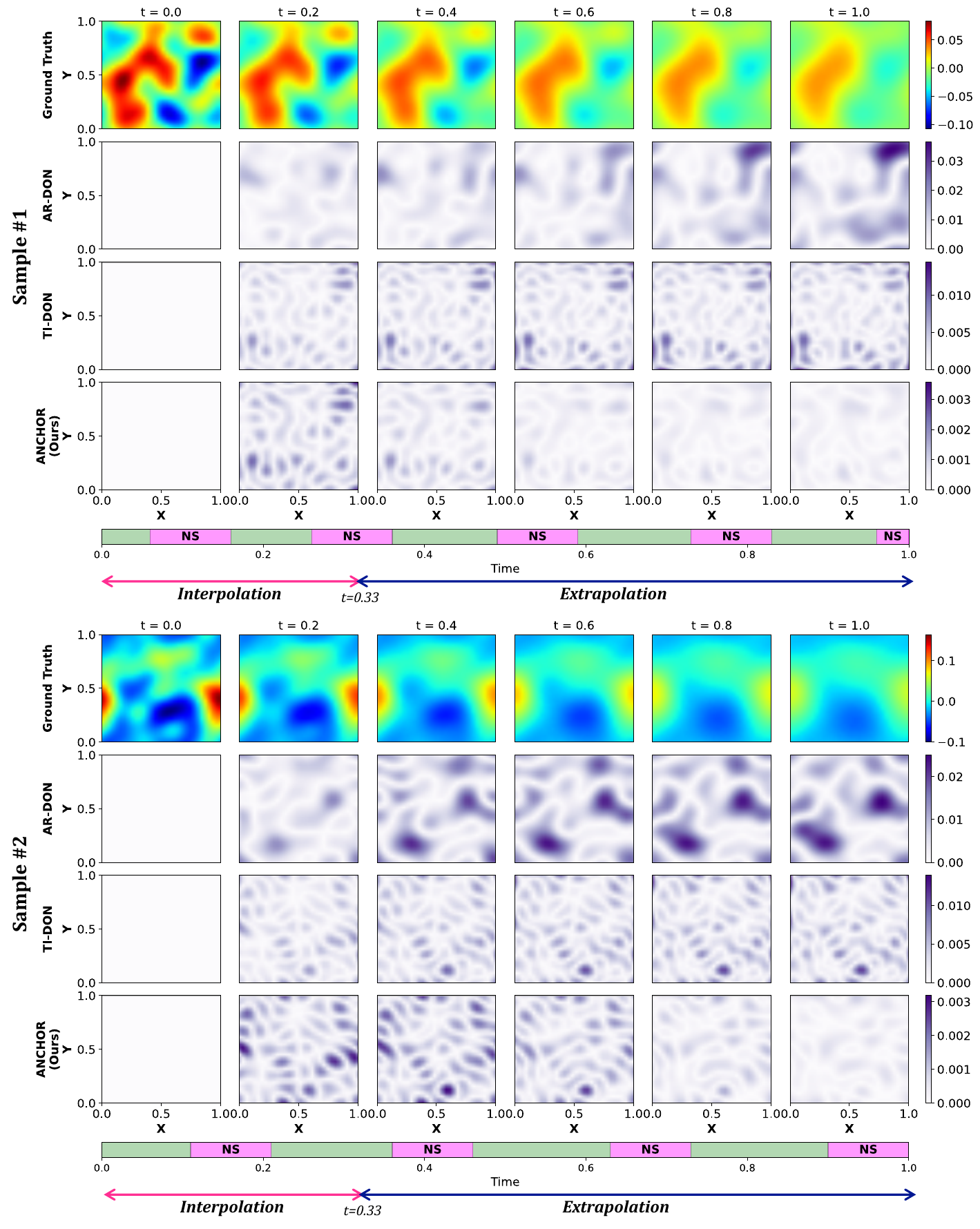}
    \caption{2D Burgers' equation: Spatial error distributions over $t\in[0,1]$, where $t\in[0,0.33]$ corresponds to the interpolation regime and $t\in[0.33,1.0]$ corresponds to extrapolation for all frameworks, illustrated using two representative samples. The error color bars for each framework are shown separately due to significant differences in the ranges of error magnitudes. The color bar below each set of contours indicates the time steps solved by TI-DON (green) and by the high-fidelity numerical solver (magenta).}
    \label{fig:2d_burgers_error_contours}
\end{figure}

Figure~\ref{fig:corr_with_err_estm_and_l2_errors} highlights several key trends: AR-DON exhibits rapid, near-exponential error growth, reflecting the severe compounding of model approximation errors in recursive inference. TI-DON partially alleviates this growth by embedding a numerical integration step during time marching; however, the resulting error remains unbounded over long time horizons. In contrast, the proposed ANCHOR framework stabilizes the rollout by intermittently invoking the high-fidelity numerical solver, thereby effectively bounding the error evolution. Consistent with the one-dimensional case, the EMA-based error estimator with a tuning parameter $a = 0.02$ remains strongly correlated with the true relative $L_2$ error ($\rho_{corr} = 0.990$) and reliably triggers solver interventions when the error estimator exceeds the adaptive threshold. Importantly, this confirms that the estimator continues to function as a physics-aware proxy for error control in settings where ground-truth errors are unavailable at inference time. For defining the adaptive threshold, we fix $\gamma = 2$.

To further assess solution quality beyond aggregate error metrics, Fig.~\ref{fig:2d_burgers_error_contours} presents spatial error contours at selected discrete time instances for two representative test samples. The AR-DON predictions exhibit errors that are several orders of magnitude larger than those of the other frameworks, underscoring their pronounced sensitivity to error accumulation in two-dimensional spatiotemporal dynamics. While TI-DON substantially reduces these errors, noticeable deviations in the solution profiles persist and continue to grow with time. By contrast, ANCHOR consistently suppresses error amplification by leveraging targeted solver corrections, resulting in spatiotemporally stable and accurate solutions even over extended rollouts. These results demonstrate that the proposed framework remains robust in high-dimensional regimes, where error propagation is significantly more challenging than in the one-dimensional case. Consistent with our hypothesis, the proposed approach reduces computational time by approximately 30\% relative to the classical numerical solver, while reducing the peak \rev{relative} $L_2$ error by roughly 80\% compared to the neural-operator surrogate. This peak-error reduction comes at the expense of the TI-DeepONet’s speed, highlighting the trade-off between accuracy and computational cost.
This signifies that ANCHOR effectively balances the computational efficiency of TI-DON with the accuracy of the high-fidelity solver.

\begin{figure}
    \centering
    \includegraphics[width=0.99\linewidth]{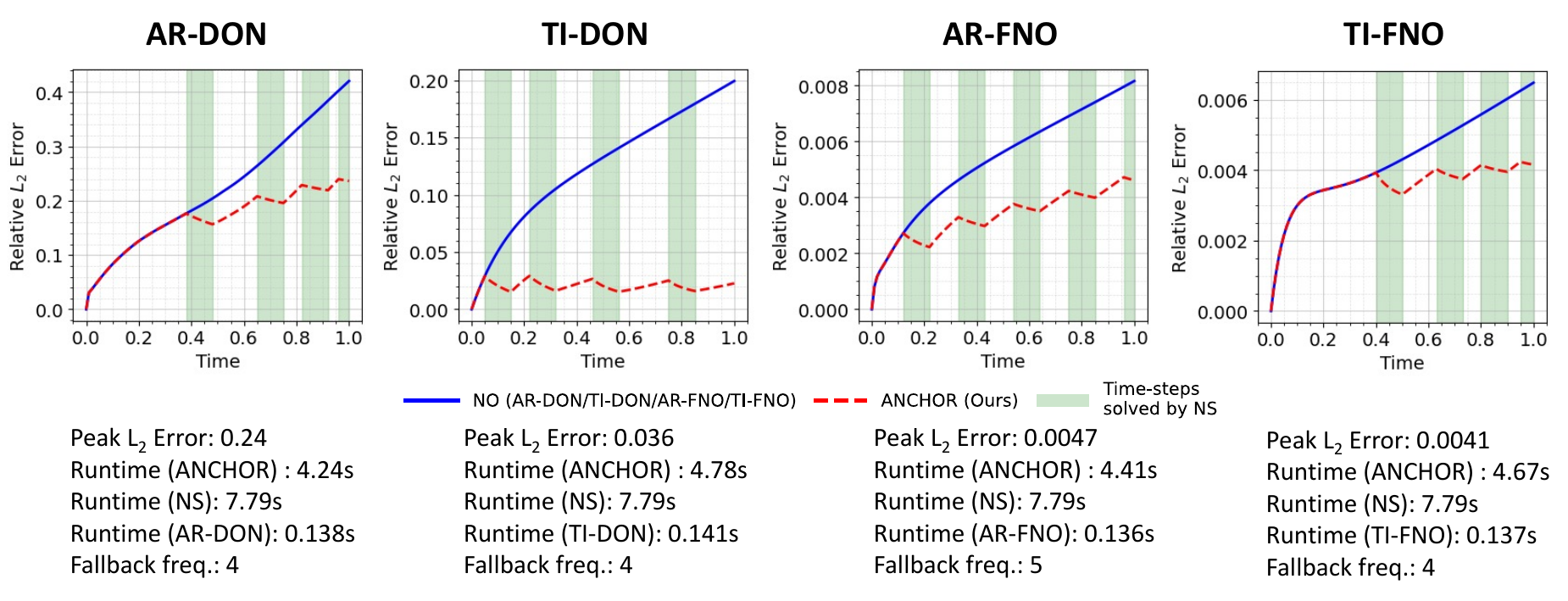}
    \caption{\rev{2D Burgers' equation: Comparison of relative $L_2$ error growth across different neural operator architectures. Each subplot shows the error evolution for a given surrogate (AR-DON, TI-DON, AR-FNO, TI-FNO) operating either as a pure surrogate or within the ANCHOR framework. TI-based neural operators (TI-DON, TI-FNO) maintain well-bounded error over time, whereas purely autoregressive neural operators (AR-DON, AR-FNO) exhibit gradual error accumulation in long-horizon predictions that are harder to bound.}}
    \label{fig:diff_no_2d_burgers}
\end{figure}

\rev{In a similar spirit to the 1D Burgers' case, we also implement AR-DON, AR-FNO, and TI-FNO as the inexpensive neural operator surrogate within the ANCHOR framework for the 2D Burgers' equation. The relative $L_2$ error behavior for a representative sample is shown in Fig.~\ref{fig:diff_no_2d_burgers}. As expected, purely autoregressive approaches such as AR-DON and AR-FNO do not yield bounded error even within the ANCHOR framework. In contrast, TI-based neural operators (TI-DON and TI-FNO), when coupled with the ANCHOR framework, exhibit bounded error behavior over time.}

\subsection{Two-dimensional Allen-Cahn Equation}
\label{subsec:example3}
The third example that we consider is the two-dimensional Allen-Cahn Equation, which models the evolution of a phase field in a material undergoing phase separation, particularly in binary alloys. The governing PDE is defined as:
\begin{equation}
    \dfrac{\partial u}{\partial t} = \epsilon^2 \left( \dfrac{\partial^2 u}{\partial x^2} + \dfrac{\partial^2 u}{\partial y^2}\right) - (u^3 - u) \quad \forall \ (x,y,t) \in [0,1]^2 \times [0,1],
\end{equation}
where $\epsilon = 0.05$ is the interfacial width or the diffusion length. Consistent with the Burgers' equation examples, a pseudo-spectral method is employed to solve the two-dimensional Allen-Cahn equation and generate the ground-truth dataset. The equation is solved on a periodic spatial domain $[0,1] \times [0,1]$, discretized using a uniform $32 \times 32$ grid. Time integration is performed up to $T_{\mathrm{final}} = 1.0$ using the ETDRK4 scheme with a coarse time step of $\Delta t = 0.01$, internally refined by a factor of 200 for numerical stability, yielding $n_{\mathrm{coarse}} = 101$ stored snapshots. Initial conditions are sampled from Gaussian-filtered random fields and scaled to the range of $[-1,1]$. During the development of the ANCHOR framework, the same solver is invoked as needed, with the final time adjusted according to the fixed number of solver steps (e.g., 10) before switching to the TI-DON surrogate at inference.

Mathematically, the Allen-Cahn equation is a nonlinear parabolic reaction-diffusion equation, and for surrogate modeling poses two key challenges: (i) the strong nonlinearity of the cubic term amplifies prediction errors in long-horizon rollouts, and (ii) the interplay between sharp interfacial layers (set by $\epsilon$) and large-scale domain coarsening requires models to resolve both fine-scale structures and large-scale phase separation dynamics. We begin with a similar analysis of the trends observed in Fig.~\ref{fig:corr_with_err_estm_and_l2_errors}. On first glance, one can observe that the errors are noticeably of a higher magnitude owing to the complexity of the PDE dynamics. This is also reflected in the maximum error threshold, which is observed at the first cutoff from the TI-DON surrogate at around 0.15, which is higher than the previous cases. However, once again, ANCHOR can reliably bound the error growth by intermittently invoking the numerical solver when the adaptive threshold is overshot, reflecting the adaptive, corrective mechanism intrinsic to ANCHOR. As with the previous PDEs, our designed physics-aware EMA-based error estimator exhibits a high Pearson correlation coefficient, $\rho_{corr} = 0.994$, with the majority of values distributed in the interval $[0.98, 1.00]$ across the test samples. The error estimator is tuned with $a = 0.01$, and the adaptive threshold decays with $\gamma = 3$. Finally, the relative $L_2$ error accumulation essentially gives a first comparison between AR-DON, with the errors blowing up to $1.5$ at the final time step, TI-DON stabilizing the error growth for AR-DON by introducing the numerical integration module embedded in the time marching process, and finally ANCHOR providing a stable, reliable, and bounded error growth for TI-DON over extended horizons.

\begin{figure}
    \centering
    \includegraphics[width=\linewidth]{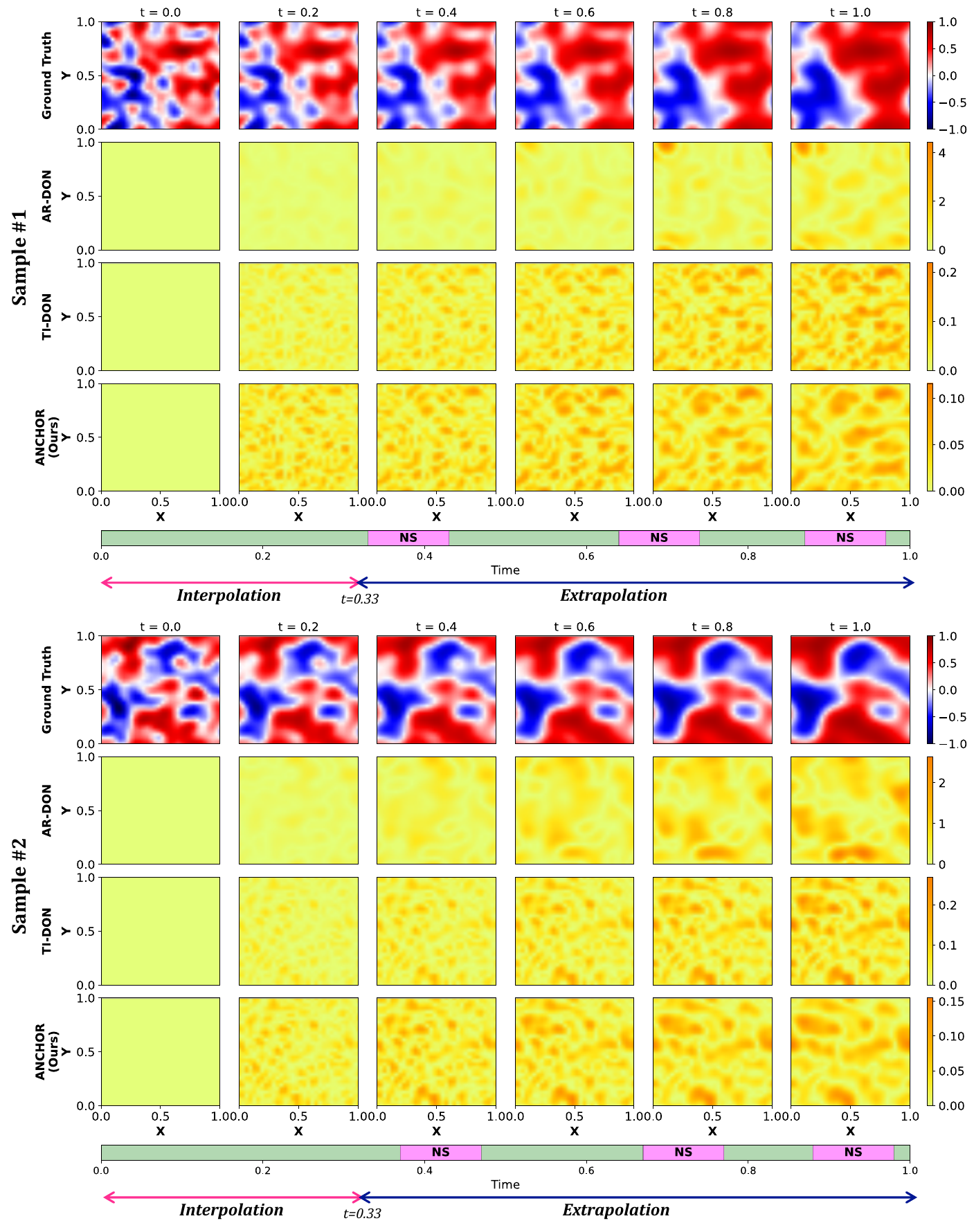}
    \caption{2D Allen-Cahn equation: Spatial error distributions over $t\in[0,1]$, where $t\in[0,0.33]$ corresponds to the interpolation regime and $t\in[0.33,1.0]$ corresponds to extrapolation for all frameworks, illustrated using two representative samples. The error color bars for each framework are shown separately due to significant differences in the ranges of error magnitudes. The color bar below each set of contours indicates the time steps solved by TI-DON (green) and by the high-fidelity numerical solver (magenta).}
    \label{fig:2d_allen_cahn_error_contours}
\end{figure}

The next step is to qualitatively assess the predictive fidelity of the solution contours across the different frameworks, as shown in Fig.~\ref{fig:2d_allen_cahn_error_contours} for two representative test cases. As expected, AR-DON exhibits errors that are orders of magnitude larger than those of the other approaches, further highlighting its pronounced sensitivity to error accumulation during recursive inference. TI-DON alleviates this behavior to some extent; however, the errors remain appreciable and are not fully bounded over time. In contrast, the ANCHOR framework, owing to its built-in adaptive corrective mechanism, effectively suppresses error growth, as clearly evidenced by the error contours, particularly in extrapolation regimes well beyond the training time horizon. These qualitative observations are in excellent agreement with the corresponding quantitative $L_2$ error trends (see Fig.~\ref{fig:corr_with_err_estm_and_l2_errors}) and are consistent with the behavior observed for the previously studied PDE systems. Owing to the high complexity of the governing PDE, classical numerical solvers incur substantial computational costs. By leveraging the computational efficiency of TI-DON, the ANCHOR framework achieves an overall reduction in computation time of approximately 75\% relative to the numerical solver, along with a 33\% reduction in peak $L_2$ error compared to TI-DON.

\begin{figure}[!htbp]
    \centering
    \includegraphics[width=0.99\linewidth]{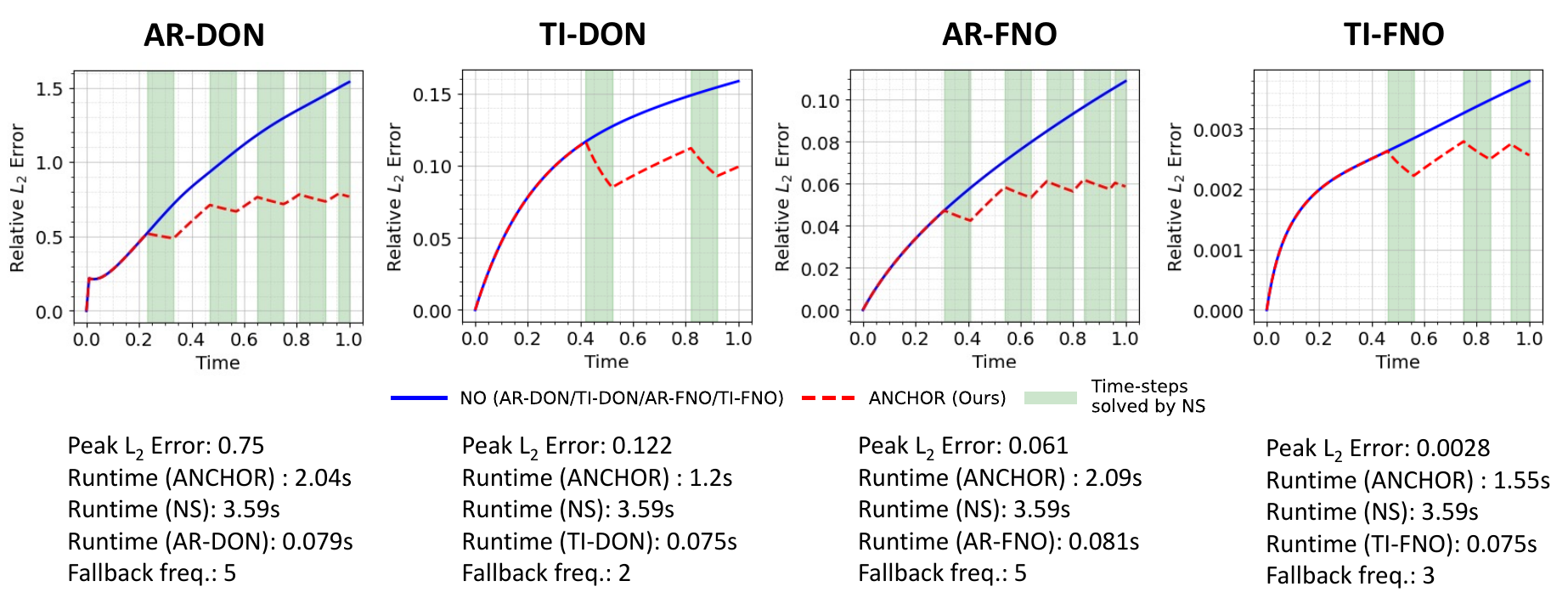}
    \caption{\rev{2D Allen-Cahn equation: Comparison of relative $L_2$ error growth across different neural operator architectures. Each subplot shows the error evolution for a given surrogate (AR-DON, TI-DON, AR-FNO, TI-FNO) operating either as a pure surrogate or within the ANCHOR framework. TI-based neural operators (TI-DON, TI-FNO) maintain well-bounded error over time, whereas purely autoregressive neural operators (AR-DON, AR-FNO) exhibit gradual error accumulation in long-horizon predictions that are harder to bound.}}
    \label{fig:diff_no_2d_ac}
\end{figure}

\rev{Consistent with earlier observations, we evaluate AR-DON, AR-FNO, and TI-FNO alongside TI-DON for the 2D Allen-Cahn equation. Figure~\ref{fig:diff_no_2d_ac} confirms the same trend: purely autoregressive surrogates (AR-DON, AR-FNO), even when fortified with numerical solver calls in ANCHOR, reduce error owing to solver correction but fail to bound it. In contrast, TI-based neural operators (TI-DON, TI-FNO) work symbiotically with the numerical solver corrections, achieving bounded error within the ANCHOR framework. As with other examples, the runtime of ANCHOR scales with the number of solver calls.}

\subsection{\rev{Two-dimensional Cahn-Hilliard Equation}}
\label{subsec:example4}
\rev{The two-dimensional Cahn-Hilliard equation is a nonlinear, fourth-order partial differential equation that models phase separation in a conserved concentration field $c(x,y,t)$. In Eq.~\ref{eq:cahn_hilliard_2d}, $M$ denotes the mobility, $W$ controls the double-well bulk free energy driving separation into two phases ($c \in [0,1]$), and $\kappa$ governs interfacial energy and thickness. The equation describes diffusion driven by chemical potential gradients while conserving total mass, and is widely used in materials science, fluid dynamics, and phase-field modeling. The governing PDE reads as:}
\begin{equation}
\rev{    \frac{\partial c}{\partial t} 
    = M \nabla^2 \left[ 2Wc(1-c)(1-2c) - \kappa \nabla^2 c \right], \quad \forall \ (x,y) \in [0,1]^2,\; t \in [0,2]}
    \label{eq:cahn_hilliard_2d}
\end{equation}

\rev{The spatial domain is a two-dimensional unit square in the $x$--$y$ plane. The Cahn-Hilliard equation is solved on a $64 \times 64$ Cartesian grid over the domain $(x, y) \in [0,1]^2$, using a second-order central difference scheme for spatial discretization and a semi-implicit (implicit-explicit) Euler method for time integration. A time step of $\Delta t = 0.01$ is employed, and the solution is advanced up to a prescribed final time. The initial concentration field is defined with a mean value of $c_0 = 0.5$, with Gaussian white noise of $10\%$ amplitude superimposed to generate multiple realizations of the initial condition.}

\begin{figure}[htb!]
    \centering
    \includegraphics[width=\linewidth]{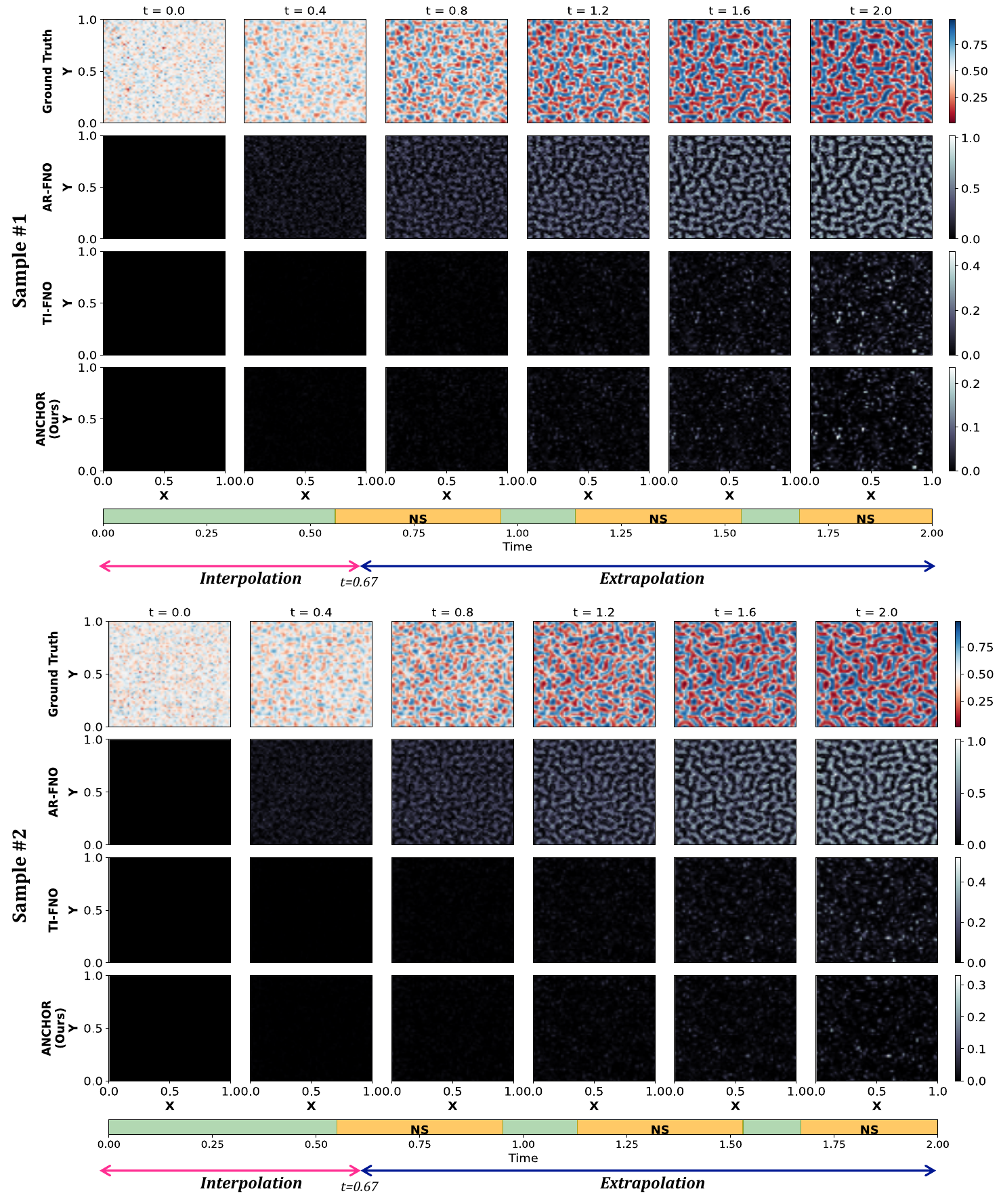}
    \caption{\rev{2D Cahn-Hilliard equation: Spatial error distributions over $t\in[0,2]$, where $t\in[0,0.67]$ corresponds to the interpolation regime and $t\in[0.67,2.0]$ corresponds to extrapolation for all frameworks, illustrated using two representative samples. The error color bars for each framework are shown separately due to significant differences in the ranges of error magnitudes. The color bar below each set of contours indicates the time steps solved by TI-FNO (green) and by the high-fidelity numerical solver (orange).}}
    \label{fig:2d_cahn_hilliard_error_contours}
\end{figure}

\rev{Unlike the previous examples, the Cahn-Hilliard equation is solved here using the TI-FNO surrogate. The Cahn-Hilliard dynamics, characterized by sharp interfacial gradients and multiscale features arising from phase separation, are well-suited to spectral methods. Hence, we employ FNO, which learns in the Fourier space, as the base NO architecture, making it an ideal candidate for coupling with a high-fidelity numerical solver within ANCHOR. Figure~\ref{fig:corr_with_err_estm_and_l2_errors_fno} presents the temporal evolution of the relative $L_2$ error for three frameworks: AR-FNO, TI-FNO, and ANCHOR. As expected, TI-FNO mitigates the error growth observed in AR-FNO. The ANCHOR framework further improves performance by not only reducing the error but also effectively bounding it over time. The proposed error estimator, applied to the 2D Cahn-Hilliard system with a smoothing factor $a=0.001$, achieves a strong correlation of $\rho_{corr}=0.986$ for the sample shown. This high correlation demonstrates its effectiveness in guiding solver switching. An adaptive threshold with decay factor $\gamma=2$ is used in this process. A qualitative comparison of error fields is provided in Fig.~\ref{fig:2d_cahn_hilliard_error_contours}. The first row shows the ground truth solution. The second row illustrates the AR-FNO predictions, which diverge significantly during long-term forecasting. The third row shows TI-FNO results, where the error is substantially reduced but continues to accumulate over time. In contrast, ANCHOR successfully controls and bounds the error, enabling stable long-horizon predictions. Consistent with earlier experiments, ANCHOR reduces computational cost by approximately 30\% relative to the numerical solver, while also decreasing the error by nearly 75\% compared to TI-FNO.}

\begin{figure}[htb!]
    \centering
    \includegraphics[width=0.7\linewidth]{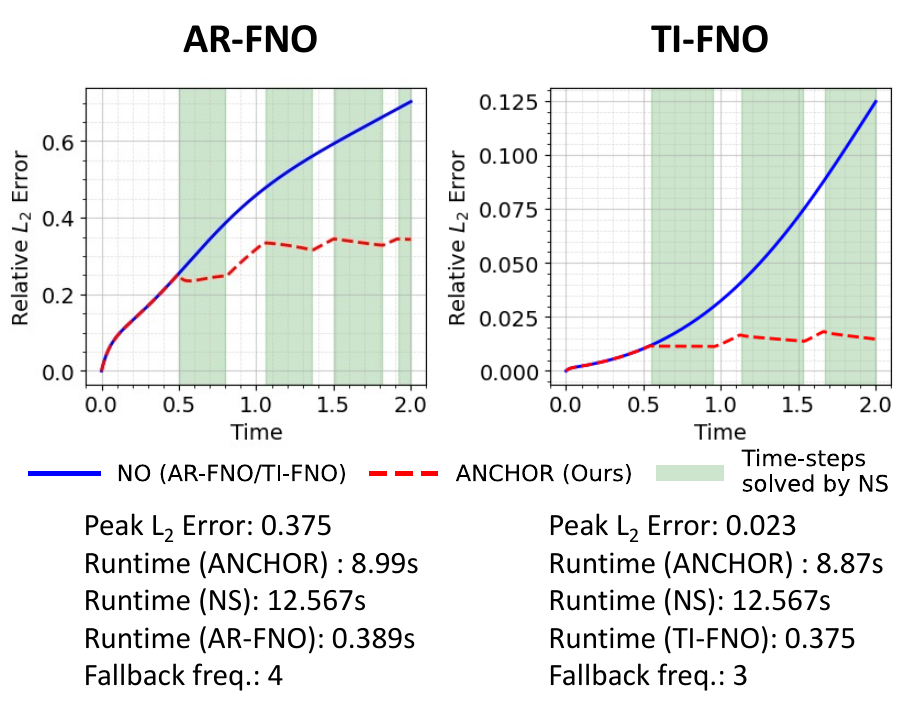}
    \caption{\rev{2D Cahn-Hilliard equation: Comparison of relative $L_2$ error growth across different neural operator architectures. Each subplot shows the error evolution for a given surrogate (AR-FNO, TI-FNO) operating either as a pure surrogate or within the ANCHOR framework. TI-based neural operators (TI-FNO) maintain well-bounded error over time, whereas purely autoregressive neural operators (AR-FNO) exhibit gradual error accumulation in long-horizon predictions that are harder to bound.}}
    \label{fig:diff_no_2d_ch}
\end{figure}

\rev{The 2D Cahn-Hilliard dynamics consist of two distinct regimes: (1) phase separation and (2) diffusion. Phase separation dominates at early times, followed by diffusion at later stages. Notably, the reduction in error due to numerical solver intervention is more pronounced during the diffusion phase, whereas the error remains relatively stable during the early phase separation stage. Consequently, the slope of the dip in error is steeper at longer horizons, exhibiting more pronounced solver correction effects on the PDE RHS approximation.}
\rev{We also couple AR-FNO with the high-fidelity solver in addition to coupling TI-FNO with the solver. Although AR-FNO coupled with the solver reduces the error (Fig.~\ref{fig:diff_no_2d_ch}), it does not effectively bound it, whereas TI-FNO coupled with the solver successfully controls and bounds the error over time. This is consistent with the theoretical analysis presented in Sec.~\ref{sec:theoretical_analysis_error_bound}, underscoring the importance of embedding numerical time integration within the neural operator surrogate to ensure sufficient corrective capacity of the numerical solver. Without this structure, as in AR-FNO, even numerical solver interventions fail to bound the error. It is worth noting that, due to the increased complexity of this case, the computation of the PDE residual introduces significant overhead to the ANCHOR framework. While ANCHOR remains faster than the numerical solver, the overall computational cost is non-trivial, and further efficiency improvements remain an opportunity for future work.}

\subsection{\revv{Two-Dimensional Navier-Stokes Equation}}
\label{subsec:example5}
{\color{black}
The fifth, and most computationally expensive, example is the two-dimensional incompressible Navier-Stokes equation, which governs viscous fluid motion and arises in applications such as weather prediction, ocean circulation, aerodynamic design, and turbulence modeling. We consider the vorticity formulation:
\begin{equation}
    \frac{\partial \omega}{\partial t} + u\frac{\partial \omega}{\partial x} + v\frac{\partial \omega}{\partial y} = \nu \Delta \omega + f(x,y,t), \quad \forall \ (x,y) \in [0,1)^2,\; t \in [0,15],
\end{equation}
where $\omega(x,y,t)$ is the scalar vorticity field, $(u,v)$ is the velocity vector field, $\nu = 10^{-3}$ is the kinematic viscosity, and $f(x,y,t)$ is an external forcing term. The equation is solved on the periodic unit torus $[0,1)^2$ using a Fourier pseudo-spectral method on a uniform $N \times N$ grid with $N = 256$. The velocity field is recovered from the vorticity via the stream function $\psi$, which satisfies $\Delta \psi = -\omega$, with:
\begin{equation*}
    u = \frac{\partial \psi}{\partial y}, \qquad v = -\frac{\partial \psi}{\partial x}.
\end{equation*}
The forcing function is defined as:
\begin{align*}
    f(x,y,t) = A e^{-\alpha t}\left[\sin(2\pi(x+y)) + \cos(2\pi(x+y))\right]\cos(\sigma_f t),
\end{align*}
where $A = 0.1$ is the forcing amplitude, $\alpha = 0.5$ is the exponential decay rate, and $\sigma_f = 0.5$ is the temporal oscillation frequency. Time integration is performed using a semi-implicit scheme: the viscous diffusion term is treated implicitly via a Crank-Nicolson update, while the nonlinear advection and forcing terms are treated explicitly. The solver advances the solution with a time step of $\Delta t = 10^{-4}$ and stores snapshots at uniform intervals of $\Delta t_{\text{save}} = 0.1$.

Similar to the 2D Cahn-Hilliard equation, the 2D incompressible Navier-Stokes equations in vorticity form are solved here using the TI-FNO surrogate. The vorticity dynamics are characterized by nonlinear vortex transport, advection-dominated behavior at low viscosity (e.g. $\nu = 10^{-3}$), and the transfer of vortical structures across spatial scales, with viscosity dissipating small-scale vorticity and enstrophy. These features make long-horizon prediction particularly challenging, since local errors in the vorticity field can be advected by the flow, introduce phase errors in coherent vortical structures, and interact nonlinearly through the convective term. At the same time, the periodic setting and spectral structure of the vorticity formulation make FNO a suitable base neural operator architecture, since it can efficiently represent global spatial interactions in Fourier space. Therefore, we employ TI-FNO as the base neural operator model and couple it with a high-fidelity pseudo-spectral numerical solver within the ANCHOR framework. Column 1 in Fig.~\ref{fig:corr_with_err_estm_and_l2_errors_fno} presents the temporal evolution of the relative $L_2$ error for three frameworks: AR-FNO, TI-FNO, and ANCHOR. As expected, AR-FNO exhibits progressive error accumulation during long-term rollout due to repeated autoregressive prediction. TI-FNO mitigates this growth by directly learning the time-integrated solution map, leading to improved stability over the autoregressive baseline. The ANCHOR framework further improves performance by adaptively invoking the numerical solver whenever the error estimator indicates increasing surrogate drift. As a result, ANCHOR not only reduces the relative $L_2$ error but also effectively bounds its growth over time. The proposed error estimator (Column 3 of Fig~\ref{fig:corr_with_err_estm_and_l2_errors_fno}) tuned with $a = 0.001$ achieves a strong correlation $(\rho_{corr} = 0.985)$ with the true relative $L_2$ error for the representative sample shown, demonstrating its effectiveness in identifying time regions where the neural operator prediction begins to drift. This enables reliable solver switching and prevents the accumulation of long-horizon errors. An adaptive threshold with decay factor $\gamma=1$ is used to control the switching criterion. 

A qualitative comparison of the error fields is provided in Fig.~\ref{fig:2d_navier_stokes_error_contours}. The first row shows the ground truth vorticity field. The second row illustrates the AR-FNO predictions, where errors gradually increase as the rollout proceeds, with the predicted vortical structures deviating from the reference solution. The third row shows TI-FNO predictions, where the error is substantially reduced compared to AR-FNO, but noticeable discrepancies still develop at later times due to nonlinear vortex transport and accumulated phase errors. In contrast, ANCHOR successfully controls the error by correcting the trajectory through the numerical solver, resulting in more stable long-horizon predictions and better preservation of vortical structures. Consistent with the trends observed in the previous experiments, ANCHOR reduces the computational cost relative to the full numerical solver while also decreasing the prediction error compared to TI-FNO. These results demonstrate that the proposed adaptive coupling strategy remains effective for advection-dominated fluid dynamics, where accurate long-time prediction requires controlling the nonlinear transport, phase drift, and accumulation of surrogate errors.
\begin{figure}[htb!]
    \centering
    \includegraphics[width=\linewidth]{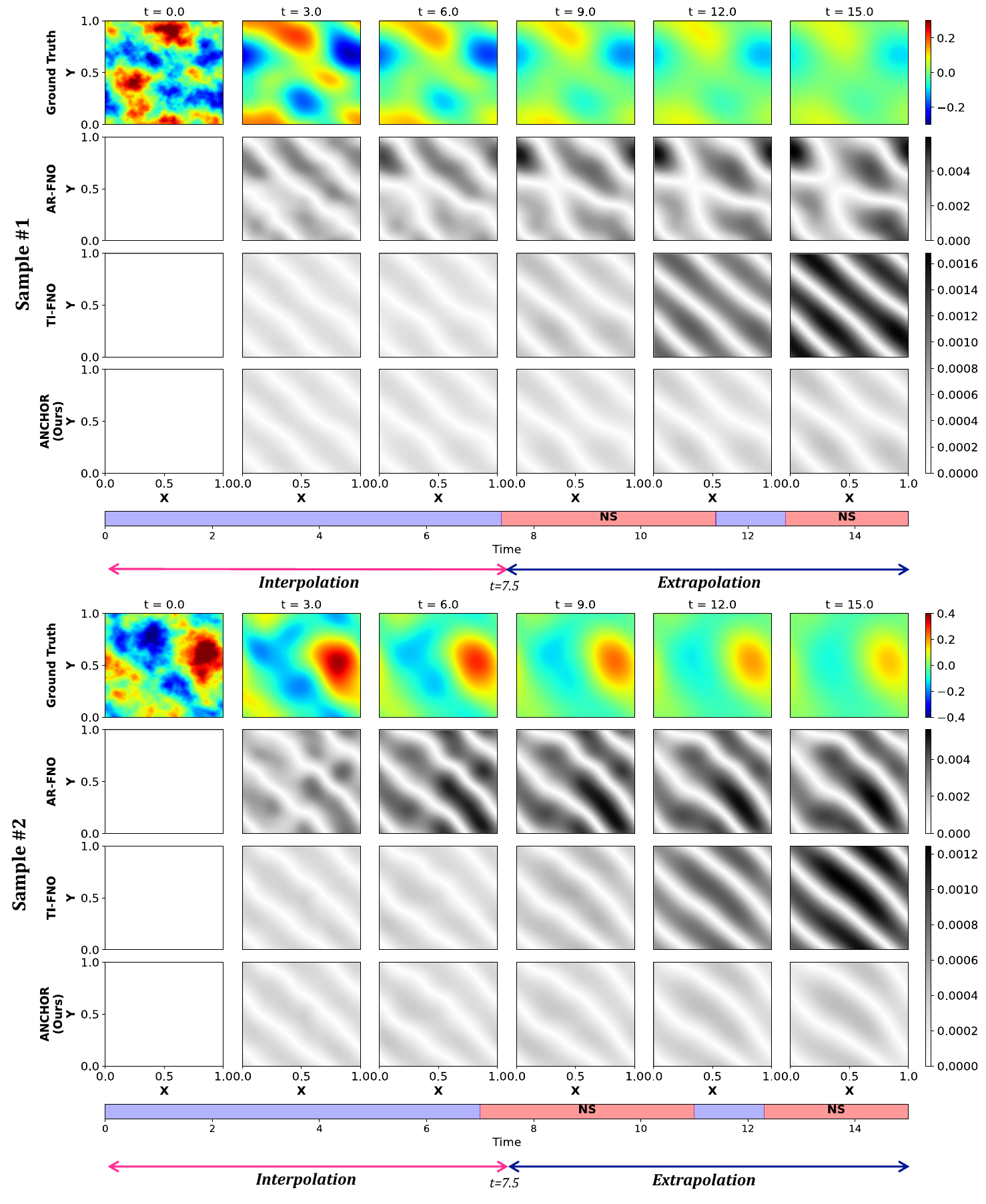}
    \caption{\revv{2D Navier-Stokes equation: Spatial error distributions over $t\in[0,15.0]$, where $t\in[0,7.5]$ corresponds to the interpolation regime and $t\in[7.5,15.0]$ corresponds to extrapolation for all frameworks, illustrated using two representative samples. The error color bars for each framework are shown separately due to significant differences in the ranges of error magnitudes. The color bar below each set of contours indicates the time steps solved by TI-FNO (blue) and by the high-fidelity numerical solver (red).}}
    \label{fig:2d_navier_stokes_error_contours}
\end{figure}

\textbf{Scaling of ANCHOR with increase in resolution.} Figure~\ref{fig:navier_stokes_scaling} compares the computational cost of the proposed ANCHOR framework with that of the full Navier-Stokes numerical solver for increasing spatial discretization levels $N=\lbrace64,128,256,512\rbrace$. As expected, the computation time increases for both approaches as the grid is refined, due to the larger number of spatial degrees of freedom. Across all resolutions, however, ANCHOR remains consistently less expensive than the full numerical solver. This reduction in cost depends on the number of times the numerical solver is invoked, which is governed by the error threshold used in the adaptive switching criterion. In this study, the number of time steps advanced using the neural operator is kept fixed, allowing the comparison to isolate the effect of spatial resolution on the overall computational cost. Although both methods exhibit increasing runtime with spatial discretization ($N\times N$), ANCHOR maintains a lower overall cost by relying primarily on the neural operator and invoking the Navier-Stokes solver only when correction is required.
\begin{figure}[!htbp]
    \centering
    \includegraphics[width=0.55\linewidth]{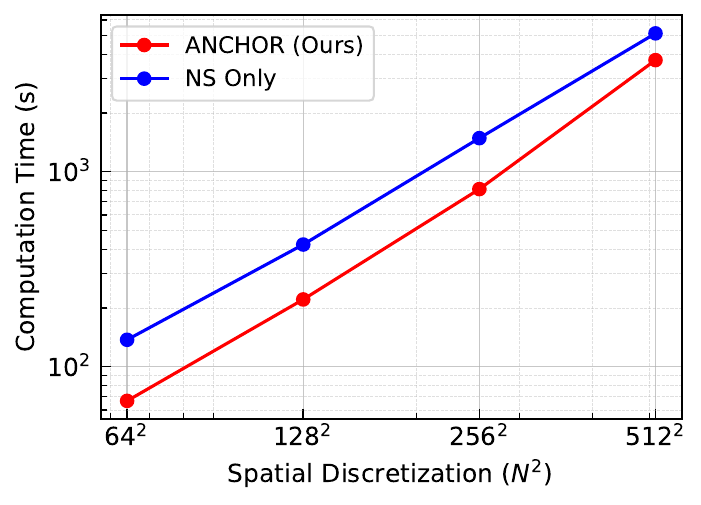}
    \caption{\revv{2D Navier–Stokes equation: Comparison of average computational time (in seconds) per sample between ANCHOR and the classical numerical solver across grid resolutions ($N \times N$). ANCHOR's cost depends on both the number of solver calls and the PDE residual computation, with the latter scaling with grid resolution.}}
    \label{fig:navier_stokes_scaling}
\end{figure}

In addition to TI-FNO, we also couple AR-FNO (Fig~\ref{fig:diff_no_2d_ns}) with the high-fidelity Navier-Stokes solver. Although AR-FNO coupled with the solver reduces the error compared to purely autoregressive rollout, it does not effectively bound the error over long-time prediction. In contrast, TI-FNO coupled with the solver successfully controls error growth and maintains it within a bounded range over time. This behavior is consistent with the theoretical analysis presented in Sec.~\ref{sec:theoretical_analysis_error_bound}, highlighting the importance of embedding numerical time integration within the NO surrogate. For the Navier-Stokes system, where nonlinear advection and low-viscosity dynamics lead to strong error accumulation, the corrective effect of the high-fidelity solver is effective only when the surrogate evolution is structured to be compatible with time-integrated state updates. Without this structure, as in AR-FNO, solver interventions can reduce instantaneous error but are insufficient to prevent subsequent error growth. It is also worth noting that, due to the increased complexity of the Navier-Stokes dynamics, the computation of the PDE residual and the velocity recovery from vorticity introduce additional overhead in the ANCHOR framework. While ANCHOR remains computationally more efficient than relying entirely on the numerical solver, the overall cost is non-negligible, and improving the efficiency of residual evaluation remains an important direction for future work.
\begin{figure}[htb!]
    \centering
    \includegraphics[width=0.7\linewidth]{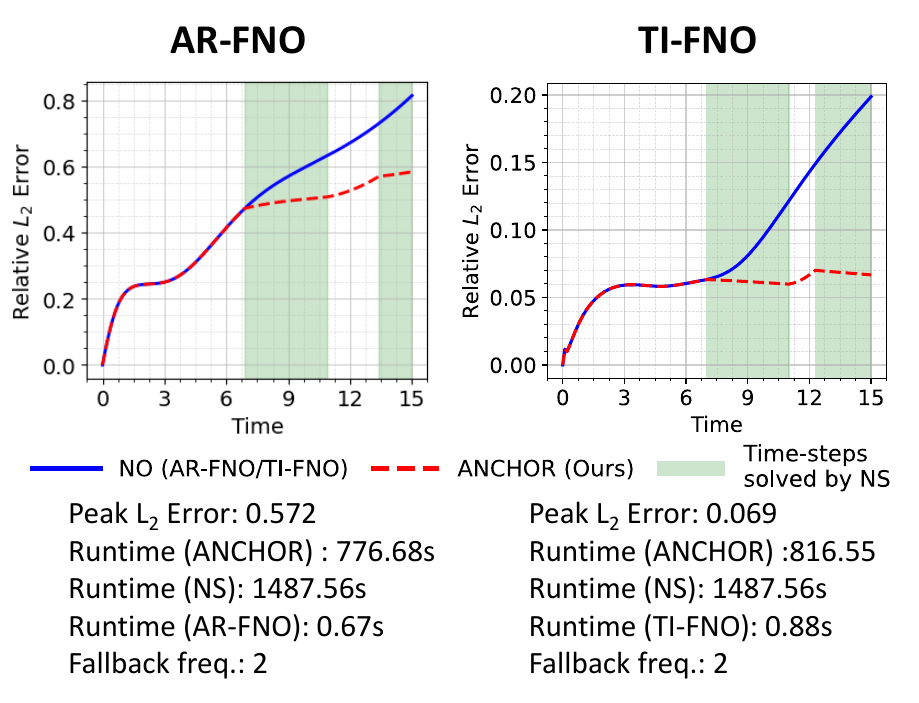}
    \caption{\revv{2D Navier-Stokes equation: Comparison of relative $L_2$ error growth across different neural operator architectures. Each subplot shows the error evolution for a given surrogate (AR-FNO, TI-FNO) operating either as a pure surrogate or within the ANCHOR framework. TI-based neural operators (TI-FNO) maintain well-bounded error over time, whereas purely autoregressive neural operators (AR-FNO) exhibit gradual error accumulation in long-horizon predictions that are harder to bound.}}
    \label{fig:diff_no_2d_ns}
\end{figure}
}

\subsection{Three-dimensional Heat Conduction}
\label{subsec:example6}
We consider three-dimensional heat conduction as the final example in this study. The heat equation is a linear parabolic PDE with a wide range of applications: it naturally arises in modeling heat transfer through a medium, appears in probability theory as the governing equation for Brownian motion, plays a fundamental role in mathematical finance through option pricing models, and is widely used in image processing for smoothing and denoising. In a three-dimensional domain, the governing equation is given by:
\begin{equation}
    \frac{\partial T}{\partial t} = \alpha\left(
    \frac{\partial^2 T}{\partial x^2} +
    \frac{\partial^2 T}{\partial y^2} +
    \frac{\partial^2 T}{\partial z^2}
    \right), \quad \forall \ (x,y,z,t) \in [0,1]^3 \times [0,1],
\end{equation}
where $\alpha = 1.0$ denotes the thermal diffusivity. Here, $T$ represents the scalar temperature field with initial condition $T_0$. The spatial domain is a three-dimensional L-shaped region, with the L-shaped cross-section situated in the XY-plane.
The three-dimensional heat conduction equation is solved using an explicit FDM solver on a 32$\times$32$\times$16 cartesian grid over the spatial domain $(x, y, z) \in [0, 1]^3$. Based on von-Neumann stability analysis, a time step of $\Delta t = 0.01$ is chosen, and the solution is advanced in time up to $t = 1.0$. The initial condition consists of a Gaussian blob, normalized to an amplitude within the range $[0, 1]$, with a standard deviation, $\sigma = 4.0$. The center of the Gaussian is randomly placed within the corner block of the L-shaped domain. Homogeneous Dirichlet boundary conditions of $T = 0$ are imposed on all eight faces of the L-shaped block.

\begin{figure}
    \centering
    \includegraphics[width=\linewidth]{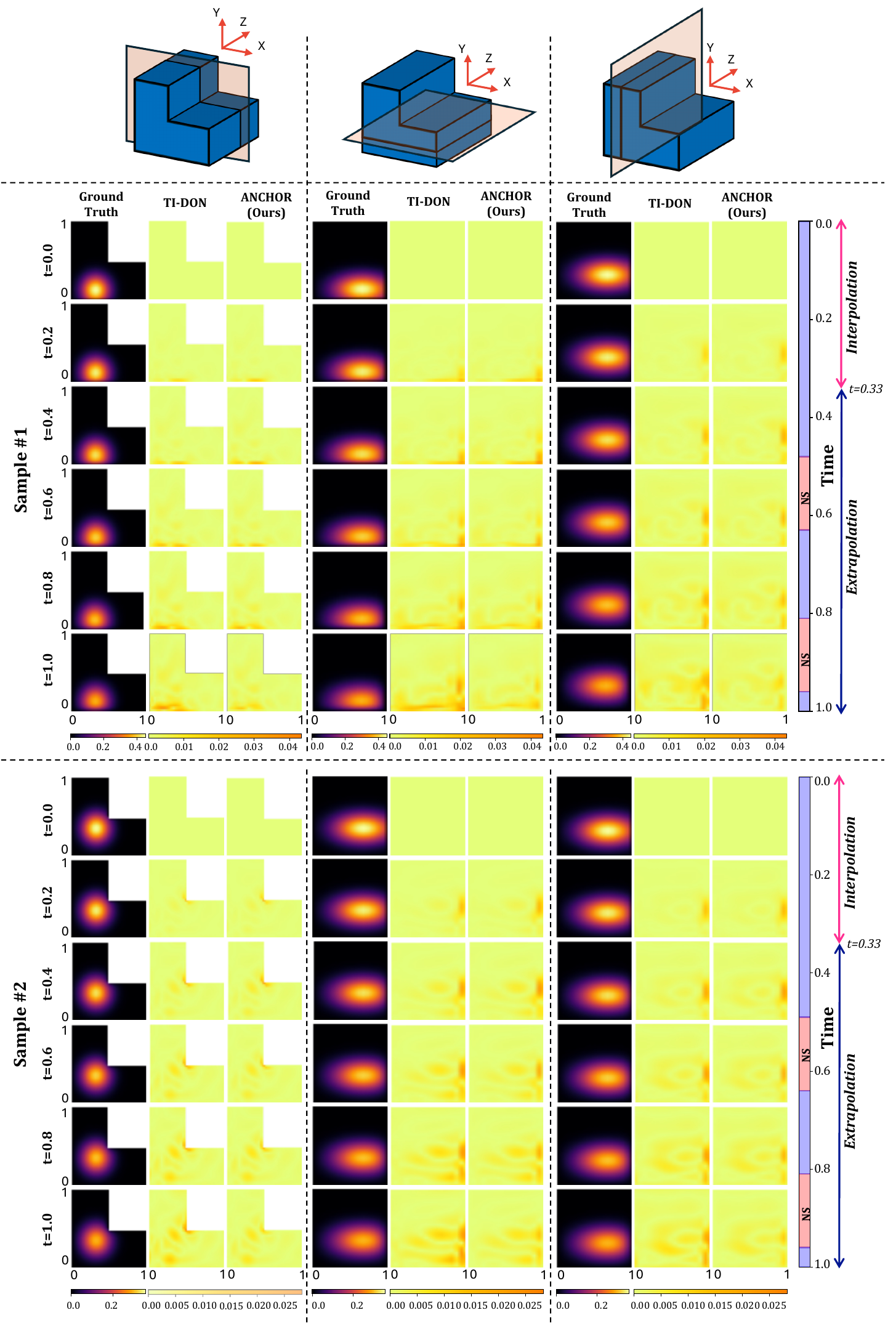}
    \caption{3D heat conduction equation: Ground truth solution field and comparison of spatial errors for TI-DON and the ANCHOR framework over $t \in [0,1]$ at selected time steps. For visualization, two-dimensional error contours on three slicing planes - XY, ZX, and YZ (shown in this order) are presented for two representative test samples. The color bar at the extreme right indicates the time steps solved by TI-DON (blue) and by the high-fidelity numerical solver (pink). Here, $t \in [0,0.33]$ corresponds to the interpolation regime, while $t \in [0.33,1.0]$ denotes the extrapolation regime.}
    \label{fig:3d_heat_error_contours}
\end{figure}

Figure~\ref{fig:corr_with_err_estm_and_l2_errors} (last row) illustrates the temporal evolution of the relative $L_2$ error across all frameworks. As expected, TI-DON mitigates the compounding error growth observed in AR-DON, while the ANCHOR framework further stabilizes the rollout through targeted corrective interventions, effectively bounding the error growth of TI-DON. The physics-aware EMA-based error estimator with $a = 0.01$ exhibits near-perfect correlation with the true relative $L_2$ solution error ($\rho_{{corr}} = 0.987$) for the sample presented, confirming its reliability for guiding solver interventions during time marching. In this case, the adaptive threshold has a decaying factor, $\gamma = 3$. A detailed qualitative comparison of prediction errors for the TI-DON and ANCHOR frameworks is presented in Fig.~\ref{fig:3d_heat_error_contours} for two unseen test conditions. For brevity, results for AR-DON are omitted, as its behavior is consistent with that observed in the previous cases. Error contours are visualized using two-dimensional profiles extracted from three orthogonal cross-sections: XY, ZX, and YZ, shown from three different isometric views. In the XY-plane, errors are observed to localize near the corner of the L-shaped domain in the first test sample and near the bottom-left region in the second. In both cases, ANCHOR reliably corrects these localized errors, demonstrating effective stabilization and bounding of error growth. Similar trends are evident in the ZX and YZ cross-sections, where ANCHOR's adaptive corrective mechanism consistently rectifies accumulated errors upon invocation of the high-fidelity numerical solver. Overall, these results demonstrate that the proposed ANCHOR framework successfully bounds prediction error growth, enabling stable and accurate long-horizon simulations in three-dimensional settings. Along similar lines, and consistent with the 2D Burgers' and Allen-Cahn cases, the computation time with ANCHOR in this case is reduced by nearly 50\% compared to the traditional solver, with around 55\% reduction in peak $L_2$ error with respect to TI-DON. 

\begin{figure}[!htbp]
    \centering
    \includegraphics[width=0.7\linewidth]{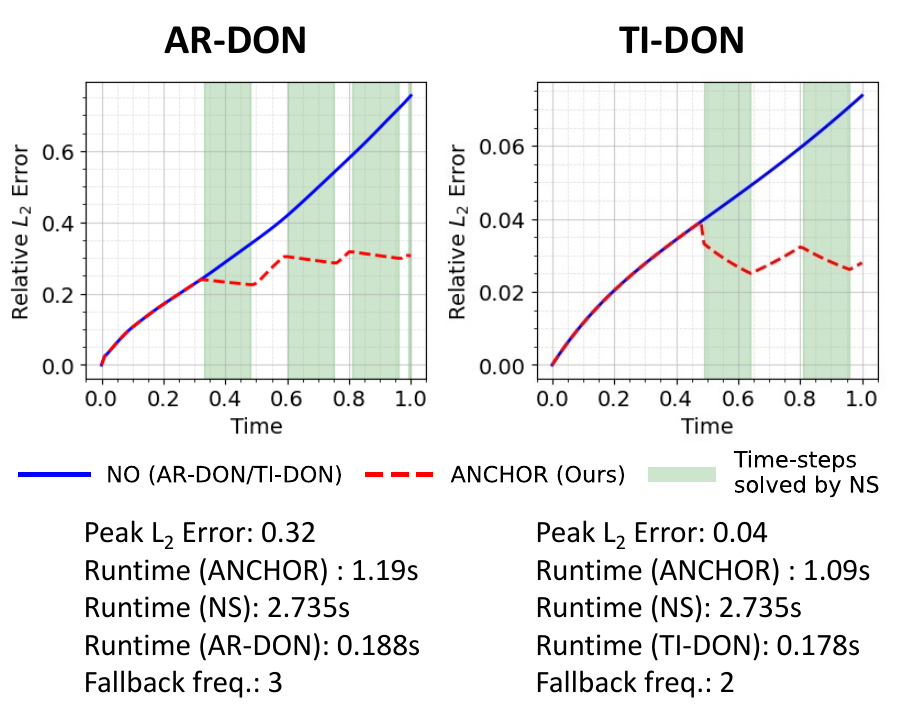}
    \caption{\rev{3D heat conduction equation: Comparison of relative $L_2$ error growth across different neural operator architectures. Each subplot shows the error evolution for a given surrogate (AR-DON, TI-DON) operating either as a pure surrogate or within the ANCHOR framework. TI-based neural operators (TI-DON) maintain well-bounded error over time, whereas purely autoregressive neural operators (AR-DON) exhibit gradual error accumulation in long-horizon predictions that are harder to bound.}}
    \label{fig:diff_no_3d_heat}
\end{figure}

\rev{Similar to the previous examples, Fig.~\ref{fig:diff_no_3d_heat} shows a consistent trend for the 3D heat equation. In a purely autoregressive setting, the error exhibits slight but noticeable growth over long time horizons. In contrast, time-integrated neural operators, i.e., TI-DON in this case, effectively bound the error over time. We do not include FNO-based experiments for this problem, as FNO requires a uniform grid; however, the computational domain for the 3D heat equation considered here is L-shaped, which is incompatible with the standard FNO formulation. As with other examples, the computational cost of ANCHOR is governed by the number of solver calls.}

\section{\rev{Ablation Study of Smoothing Parameter and Decay Factor}}
\rev{In this work, the primary hyperparameters are those associated with the pretrained NOs. Since these models are pretrained, their hyperparameters are kept fixed and are not subject to further tuning in this study. More specific to our study are the hyperparameters associated with the different components of ANCHOR that are relevant at inference time. Notably, we focus on two tunable hyperparameters: the smoothing parameter $a$ used in the EMA-based error estimator, and the decay factor $\gamma$ governing the adaptive threshold. We conduct a detailed ablation study to evaluate the impact of $a$ and $\gamma$ on performance.}

\rev{In Table~\ref{tab:ablation_smoothing_factor}, we present a comprehensive ablation study examining how different values of $a \in \{0.1, 0.05, 0.01, 0.005, 0.001\}$ influence the Pearson correlation coefficient, $\rho_{corr}$, between the proposed EMA-based error estimator and the true relative $L_2$ error computed with respect to the ground-truth solution. Across most cases, we observe that assigning a smaller weight to the residual at the current time step (i.e., using a lower value of $a$) results in an error estimator that more closely aligns with the actual error. In other words, decreasing $a$ generally improves the correlation with the true error. This behavior may be related to the diffusive nature of the PDEs considered. During early transient phases, the dynamics are more pronounced, whereas at longer times, diffusion dominates and the dynamics slow down. In such regimes, greater reliance on past values (i.e., smaller $a$) is sensible, as the system evolves more gradually and smoothing helps filter out spurious fluctuations in the residual. However, this trend is not consistent across all scenarios. To date, the choice of $a$ has been largely heuristic; a more systematic investigation, including an adaptive strategy for selecting $a$ based on the dynamical regime, may further improve performance and is deferred to future work.}

\begin{table}[htb!]
\centering
\color{black}
\renewcommand{\arraystretch}{1.2}
\caption{\rev{Pearson correlation coefficient, $\rho_{corr}$, between the EMA-based error estimator and the true relative $L_2$ error for varying smoothing parameter $a$ across different PDE examples and surrogate models.}}

\begin{tabular}{|c|c|c|c|c|c|c|}
\hline
\multirow{2}{*}{\rev{PDE Example}} & \multirow{2}{*}{Method} & \multicolumn{5}{c|}{Smoothing Parameter $a$} \\
\cline{3-7}
 &  & 0.1 & 0.05 & 0.01 & 0.005 & 0.001 \\
\hline
\multirow{4}{*}{1D Burgers'}
& AR-DON & 0.995 & 0.993 & 0.986 & 0.984 & 0.973 \\
& TI-DON & 0.990 & 0.992 & 0.990 & 0.987 & 0.980 \\
& AR-FNO & 0.778 &  0.904     &   0.997    &   0.998    &    0.996   \\
& TI-FNO &  0.888     & 0.949      &  0.996     &    0.994    & 0.993      \\
\hline
\multirow{4}{*}{2D Burgers'}
& AR-DON &  0.653     &    0.846   &   0.973    &     0.987  &  0.994     \\
& TI-DON &  0.971     &      0.989 &    0.990   &  0.986     &    0.961   \\
& AR-FNO &   0.897    &     0.968  &   0.990    &      0.991 &   0.990    \\
& TI-FNO &  0.927     &  0.952     &  0.972     &    0.978   &     0.981  \\
\hline
\multirow{4}{*}{2D Allen-Cahn}
& AR-DON &  0.770     &   0.998    &   0.996    &  0.992     &   0.990    \\
& TI-DON &    0.873   &   0.970    &   0.976    &   0.999    &   0.991    \\
& AR-FNO &   0.833    &  0.921     &   0.936    &   0.985    &  0.996     \\
& TI-FNO &   0.768    &  0.910     &  0.939     &  0.967     &  0.991     \\
\hline
\multirow{2}{*}{2D Cahn-Hilliard}
& AR-FNO &   0.798    &   0.877    &  0.919     &   0.989    &   0.995    \\
& TI-FNO &   0.786    &   0.820    &    0.946   &  0.968     &   0.984    \\
\hline
\revv{\multirow{2}{*}{2D Navier-Stokes}}
& \revv{AR-FNO} &   \revv{0.801}    &   \revv{0.899}    &  \revv{0.955}    & \revv{0.967}    &   \revv{0.978}    \\
& \revv{TI-FNO} &   \revv{0.954}    &   \revv{0.965}    &  \revv{0.968}   &  \revv{0.977}     &  \revv{0.986}    \\
\hline
\multirow{2}{*}{3D Heat}
& AR-DON &  0.877     &   0.890    &  0.977     &   0.989    &   0.995    \\
& TI-DON &  0.863     &   0.897    &  0.997     &   0.996    &   0.995    \\
\hline
\end{tabular}
\label{tab:ablation_smoothing_factor}
\end{table}

\rev{In Table~\ref{tab:ablation_decay_factor}, we analyze the effect of $\gamma$ on error decay and the number of numerical solver calls across all PDE examples and methods considered. As expected, increasing $\gamma$ leads to a reduction in peak error. However, this improvement comes at the cost of a higher number of solver calls, thereby increasing the overall computational expense. This is intuitive because $\gamma$ controls the decay rate of the adaptive threshold, and inducing a steeper decay in the threshold causes the cutoff on the error estimator to decrease accordingly, thereby triggering earlier and more frequent high-fidelity solver interventions. In our experiments, we heuristically identify a trade-off region where $\gamma$ balances error control and computational cost effectively. While the current formulation treats $\gamma$ as a user-specified hyperparameter, it can be informed by the physics of the problem. For many dissipative PDEs, the decay rate can be estimated \emph{a priori} from the governing equations; for example, in linear diffusion-dominated problems, the decay rate is related to the smallest eigenvalue of the diffusion operator, which scales as $\mathcal{O}(1/L^2)$, where $L$ is the characteristic domain length. Alternatively, $\gamma$ can be estimated online by monitoring the decay of the solution norm $\|u^n\|$ over a few initial time steps and fitting an exponential decay model. Nevertheless, a principled approach for selecting an optimal $\gamma$ remains an open question; potential avenues include data-driven methods that adapt the threshold based on evolving solution dynamics or \emph{a posteriori} error estimators, and this constitutes an important direction for future work.}
\begin{table}[!htbp]
\centering
\color{black}
\renewcommand{\arraystretch}{1.2}
\caption{\rev{Effect of the adaptive threshold decay factor $\gamma$ on error reduction and number of numerical solver (NS) calls for different surrogate models across PDE examples.}}
\resizebox{\linewidth}{!}{%
\begin{tabular}{|c|c|cc|cc|cc|cc|}
\hline
\multirow{2}{*}{PDE Example} & \multirow{2}{*}{Method} & \multicolumn{2}{c|}{$\gamma=1$}
    & \multicolumn{2}{c|}{$\gamma=3$}
    & \multicolumn{2}{c|}{$\gamma=10$}
    & \multicolumn{2}{c|}{$\gamma=20$} \\
\cline{3-10}
 &  & Error $\downarrow$ (\%) & \# NS
& Error $\downarrow$ (\%) & \# NS
& Error $\downarrow$ (\%) & \# NS
& Error $\downarrow$ (\%) & \# NS \\
\hline

\multirow{4}{*}{1D Burgers'}
& AR-DON & 66.00 & 3 & 80.40 & 6 & 90.60 & 8 & 94.00 & 9 \\
& TI-DON & 52.94 & 3 & 35.29 & 1 & 70.59 & 6 & 76.47 & 8 \\
& AR-FNO & 0.00 & 0 & 60.00 & 4 & 70.00 & 6 & 84.00 & 8 \\
& TI-FNO & 20.00 & 0 & 40.00 & 2 & 86.00 & 6 & 92.00 & 8 \\
\hline

\multirow{4}{*}{2D Burgers'}
& AR-DON & 16.28 & 1 & 56.28 & 5 & 72.09 & 7 & 76.74 & 7 \\
& TI-DON & 83.50 & 5 & 80.00 & 4 & 85.00 & 5 & 90.00 & 6 \\
& AR-FNO & 58.33 & 5 & 66.67 & 6 & 68.33 & 6 & 70.00 & 7 \\
& TI-FNO & 20.00 & 3 & 45.00 & 5 & 50.00 & 6 & 60.00 & 7 \\
\hline

\multirow{4}{*}{2D Allen-Cahn}
& AR-DON & 10.67 & 1 & 47.93 & 5 & 62.93 & 6 & 69.67 & 6 \\
& TI-DON & 13.33 & 1 & 22.00 & 2 & 56.00 & 8 & 57.33 & 8 \\
& AR-FNO & 26.67 & 2 & 49.17 & 5 & 62.50 & 6 & 68.33 & 6 \\
& TI-FNO & 5.00 & 0 & 30.00 & 3 & 42.50 & 6 & 45.00 & 6 \\
\hline

\multirow{2}{*}{2D Cahn-Hilliard}
& AR-FNO & 20.43 & 2 & 50.86 & 4 & 73.29 & 5 & 76.29 & 6 \\
& TI-FNO & 90.40 & 4 & 85.60 & 3 & 93.60 & 5 & 98.40 & 5 \\
\hline

\multirow{2}{*}{\revv{2D Navier-Stokes}}
& \revv{AR-FNO} & \revv{34.34} & \revv{2} & \revv{61.62} & \revv{3} & \revv{65.45} & \revv{4} & \revv{76.65} & \revv{4} \\
& \revv{TI-FNO} & \revv{67.17} & \revv{2} & \revv{69.20} & \revv{3} & \revv{80.23} & \revv{3} & \revv{91.41} & \revv{4} \\
\hline

\multirow{2}{*}{3D Heat}
& AR-DON & 26.75 & 1 & 59.61 & 3 & 61.17 & 4 & 62.60 & 4 \\
& TI-DON & 27.27 & 1 & 46.75 & 2 & 57.14 & 3 & 59.48 & 3 \\
\hline
\end{tabular}%
}
\label{tab:ablation_decay_factor}
\end{table}

\section{Conclusions}
\label{sec:conclusions}
In this work, we presented ANCHOR, an adaptive numerical-neural coupling framework that enhances existing sequential neural surrogates with targeted high-fidelity numerical corrections to enable stable, accurate, and reliable long-horizon predictions. The first key contribution is the introduction of a physics-aware EMA-based error estimator that incorporates the underlying PDE dynamics as a prior, yielding a robust surrogate for the true relative $L_2$ solution error. Second, we developed a seamless solver-in-the-loop mechanism that efficiently couples TI-DeepONet/TI-FNO with a high-fidelity numerical solver, resulting in a solver-augmented neural surrogate that inherits both the computational efficiency of neural operators and the accuracy of classical solvers. This approach transforms sequential neural operators from black-box predictors into transparent, self-correcting computational tools. 

Our observations can be summarized as follows:
\begin{itemize}[leftmargin=*]
    \item For the 1D Burgers' equation, which served as a primary validation example, ANCHOR reliably bounds long-horizon error growth to within $5\%$. However, computational gains are limited due to the low dimensionality of the problem.
    \item For higher-dimensional problems such as the 2D Burgers', 2D Allen-Cahn, \rev{2D Cahn-Hilliard,} \revv{2D Navier-Stokes,} and 3D heat conduction equations, ANCHOR achieves computational speedups of up to $2\times$--$3\times$ over traditional numerical solvers, while simultaneously ensuring bounded error growth over temporal domains extending up to twice the training interval. \rev{From our numerical experiments, we observed that the error bounds reduce the peak relative $L_2$ error by approximately 30--70\%.} These results demonstrate promising scalability to complex, high-dimensional, and multiphysics settings. However, there is an inherent trade-off between peak \rev{relative} $L_2$ error reduction with respect to the neural operator surrogate and computational cost. Increasing the frequency of solver calls to suppress error growth improves accuracy at the expense of increased computational cost, whereas fewer solver calls reduce cost but allow larger errors. \rev{For more complex PDEs, such as the 2D Cahn-Hilliard and \revv{the 2D Navier-Stokes equations}, residual computation for error estimation also adds significant computational overhead to the ANCHOR framework.}
    
    \item Collectively, these results highlight a key insight: the value of hybrid neural-numerical approaches scales with problem dimensionality and complexity. While pure neural methods excel at capturing high-dimensional dynamics efficiently and numerical methods provide guaranteed accuracy, ANCHOR leverages the complementary strengths of both paradigms. Incorporating a physics-aware EMA-based error estimator along with an adaptive, decaying, time-dependent threshold facilitates online, instance-aware prediction that accounts for physical priors embedded in the system. This paves the way for building solvers that account for variability in system initial and parametric conditions, thereby increasing trustworthiness in the forward surrogate model and making it practical for real-world deployment.
\end{itemize}
Looking ahead, while the current error estimator relies on PDE residual evaluations, introducing additional computational overhead and potential limitations in certain settings, future work may explore recovery-based or alternative physics-informed error estimators to further improve efficiency and robustness. Although this study focused on dissipative systems, the ANCHOR paradigm is broadly extensible; analogous physics-aware error estimators and adaptive thresholds can be developed for chaotic, dispersive, or multiscale PDEs. Perhaps most importantly, ANCHOR demonstrates that the future of scientific computing lies not in replacing numerical methods with neural networks, but in developing intelligent frameworks that orchestrate their complementary strengths. As computational demands in science and engineering continue to grow, such hybrid approaches offer a practical path toward simulations that are both efficient and reliable enough for decision-making in critical applications.

\section*{Acknowledgments}
The authors acknowledge partial support from the National Science Foundation (NSF) under Grant Nos. 2438193 and 2436738, and from the U.S. Department of Energy (DOE), Office of Science, Office of Advanced Scientific Computing Research, under Award No. DE-SC0024162. Furthermore, for the computing support, the authors would like to acknowledge the Advanced Research Computing at Hopkins (ARCH) core facility at Johns Hopkins University and the Rockfish cluster. ARCH core facility (\url{rockfish.jhu.edu}) is supported by the NSF grant number OAC1920103. Any opinions, findings, conclusions, or recommendations expressed in this material are those of the author(s) and do not necessarily reflect the views of the funding organizations.

\bibliographystyle{unsrt}
\bibliography{references}

\newpage
\renewcommand{\thetable}{A\arabic{table}}  
\renewcommand{\thefigure}{A\arabic{figure}} 
\makeatother
\setcounter{figure}{0}
\setcounter{table}{0}
\setcounter{section}{1}
\setcounter{page}{1}
\appendix
\section{\revv{Theoretical Analysis}}
\label{sec:theoretical_analysis_error_bound}
\revv{
As mentioned earlier, switching to a high-fidelity numerical solver post the TI-NO surrogate phase induces numerical correction and helps reduce the ``drift'' of the predicted solution from the true trajectory. We hypothesize that this happens due to the correction of the RHS of the PDE, which is a function of the state and its corresponding spatial derivatives (assuming autonomous dynamical systems). Below, we present a theoretical analysis of the error growth during the pure TI-NO surrogate phase and what happens after invoking the numerical solver once a prescribed error estimator threshold is reached.
}

\revv{
The analysis employs the one-dimensional heat equation as an analytically tractable model problem representative of dissipative PDEs. Our aim is to elucidate the fundamental error correction mechanism of ANCHOR, namely the interplay between surrogate-induced error accumulation and dissipation-driven contraction, rather than to establish fully general error bounds. The applicability of this mechanism to the nonlinear dissipative PDEs considered in our numerical experiments is supported by their shared mathematical structure (dissipativity) and validated empirically.
}

\revv{
Consider a simple one-dimensional heat conduction equation given by:
}
\begin{equation}
\revv{
    u_t = \alpha u_{xx},
    }
    \label{eq:1D_heat}
\end{equation}
\revv{
where $\alpha > 0$ is the thermal diffusivity. Assume periodic boundary conditions on the domain $[0, L]$. It is important to note that Eq.~\ref{eq:1D_heat} can be expressed in the general form described in Eq.~\ref{eq:time-dep-PDE}, where $\mathcal{N}(u) \equiv \alpha u_{xx}$.
}

\revv{
Let $u_n$ and $\hat{u}_n$ represent the solution states obtained at time step $n$ from the high-fidelity numerical solver and the TI-NO surrogate, respectively. Furthermore, let $\mathcal{N}$ and $\hat{\mathcal{N}}$ be the PDE right-hand side (RHS) given by the numerical solver and the TI-NO surrogate, respectively. As alluded to previously, the TI-NO model learns a neural operator-based approximation to the PDE RHS, denoted by $\hat{\mathcal{N}}$.
}

\revv{Let us assume that:}
\begin{equation}
\revv{
    \hat{\mathcal{N}}(u) = \mathcal{N}(u) + \varepsilon(u),
    }
    \label{eq:surrogate_error}
\end{equation}
\revv{
where $\varepsilon(u)$ represents the point-wise error in learning the PDE RHS.
}

\revv{Assume this error is bounded:}
\begin{equation}
\revv{
    \|\varepsilon(u)\| \leq \epsilon_S,
    }
    \label{eq:error_bound}
\end{equation}
\revv{
for all $u$ in the relevant solution manifold. Here, $\epsilon_S$ represents the surrogate's accuracy in learning $\mathcal{N}$.
We do not strictly require the surrogate RHS error to have exactly zero spatial mean. Instead, for any periodic function $f$, define its spatial mean and mean-free component as:
}
\begin{equation}
\revv{
    \bar{f} := \frac{1}{L}\int_0^L f(x)\,dx,
    \qquad
    f' := f - \bar{f}.
}
\end{equation}
\revv{
Applying this decomposition to the state error $w_n := u_n-\hat{u}_n$ and the surrogate RHS error $\varepsilon_n := \varepsilon(\hat{u}_n)$ gives
}
\begin{equation}
\revv{
    w_n = \bar{w}_n + w_n',
    \qquad
    \varepsilon_n = \bar{\varepsilon}_n + \varepsilon_n',
}
\end{equation}
\revv{
where $w_n'$ and $\varepsilon_n'$ have zero spatial mean. The Poincaré inequality will be applied only to the mean-free component $w_n'$, while the spatial mean $\bar{w}_n$ is tracked separately. This distinction is necessary because the heat equation dissipates spatially varying modes but does not dissipate the spatially constant mode.
}

\revv{
For simplicity, let us assume the numerical time-stepping scheme involved in both the TI-NO surrogate and the high-fidelity numerical solver is the first-order explicit Forward Euler method.}

\revv{One step of the numerical solver can be expressed as:}
\begin{equation}
\revv{
    u_{n+1} = u_n + \Delta t \cdot \mathcal{N}(u_n).
    }
    \label{eq:euler_step_NS}
\end{equation}

\revv{One step of the TI-NO surrogate can be expressed as:}
\begin{align}
   \revv{\hat{u}_{n+1}} &= \revv{\hat{u}_n + \Delta t \cdot \hat{\mathcal{N}}(\hat{u}_n)} \nonumber \\
   &= \revv{\hat{u}_n + \Delta t \cdot \mathcal{N}(\hat{u}_n) + \Delta t \cdot \varepsilon(\hat{u}_n).}
    \label{eq:euler_step_TI}
\end{align}
\revv{\textbf{Error evolution:} Define $w_n = u_n - \hat{u}_n$ and $e_n = \|u_n - \hat{u}_n\|=\|w_n\|$. Subtracting Eq.~\ref{eq:euler_step_TI} from Eq.~\ref{eq:euler_step_NS}, we get:}
\begin{equation}
\revv{
    u_{n+1} - \hat{u}_{n+1} = (u_n - \hat{u}_n) + \Delta t \left[\mathcal{N}(u_n) - \mathcal{N}(\hat{u}_n)\right] - \Delta t \cdot \varepsilon(\hat{u}_n).}
    \label{eq:error_evolution}
\end{equation}

\revv{Taking norms and applying the triangle inequality:}
\begin{align}
    \revv{e_{n+1}} &= \revv{\|u_{n+1} - \hat{u}_{n+1}\|} \nonumber \\
    &= \revv{\|(u_n - \hat{u}_n) + \Delta t[\mathcal{N}(u_n) - \mathcal{N}(\hat{u}_n)] - \Delta t \cdot \varepsilon(\hat{u}_n)\|} \nonumber \\
    &\leq \revv{\|(u_n - \hat{u}_n) + \Delta t[\mathcal{N}(u_n) - \mathcal{N}(\hat{u}_n)]\| + \Delta t \cdot \|\varepsilon(\hat{u}_n)\|} \nonumber \\
    &\leq \revv{\|(u_n - \hat{u}_n) + \Delta t[\mathcal{N}(u_n) - \mathcal{N}(\hat{u}_n)]\| + \Delta t \cdot \epsilon_S.}
    \label{eq:error_triangle}
\end{align}

\textbf{\revv{Dissipativity analysis:}} \revv{We now leverage the dissipativity property of $\mathcal{N}$ in the heat equation. Recall that $\mathcal{N}(u) = \alpha u_{xx}$.}

\revv{Let $u_n$ and $\hat{u}_n$ be two different solution states for the heat equation. Consider one step of the explicit first-order Forward Euler method:}
\begin{align}
    \revv{u_{n+1}} &= \revv{u_n + \Delta t \cdot \alpha \frac{\partial^2 u_n}{\partial x^2},} \\
    \revv{\hat{u}_{n+1}} &= \revv{\hat{u}_n + \Delta t \cdot \alpha \frac{\partial^2 \hat{u}_n}{\partial x^2}.}
\end{align}

\revv{Define the discrepancy between the two solution states as $w_n = u_n - \hat{u}_n$. Since the operator $\mathcal{N}(u) = \alpha u_{xx}$ is linear, the discrepancy satisfies:}
\begin{equation}
   \revv{ \mathcal{N}(u_n) - \mathcal{N}(\hat{u}_n) = \mathcal{N}(u_n - \hat{u}_n) = \alpha \frac{\partial^2 w_n}{\partial x^2}.}
\end{equation}

\revv{Therefore, $w_n$ satisfies the governing homogeneous error equation, and an equivalent Euler step can be written as:}
\begin{equation}
    \revv{w_{n+1} = w_n + \Delta t \cdot \alpha \frac{\partial^2 w_n}{\partial x^2}.}
    \label{eq:w_evolution}
\end{equation}

\revv{Let us measure the ``size'' of $w$ using the $L^2$ norm and analyze whether this discrepancy grows or decays.}

\revv{Define $\|w\|^2 = \int_0^L w(x)^2 \, dx$. We have:}
\begin{align}
   \revv{\|w_{n+1}\|^2} &= \revv{\int_0^L \left(w_n + \Delta t \cdot \alpha \frac{\partial^2 w_n}{\partial x^2}\right)^2 dx} \nonumber \\
    &= \revv{\int_0^L w_n^2 \, dx + 2\alpha\Delta t \int_0^L w_n \cdot \frac{\partial^2 w_n}{\partial x^2} \, dx + (\Delta t \cdot \alpha)^2 \int_0^L \left(\frac{\partial^2 w_n}{\partial x^2}\right)^2 dx.}
    \label{eq:w_norm_expansion}
\end{align}

\revv{Applying integration by parts to the middle term:}
\begin{equation}
  \revv{\int_0^L w_n \frac{\partial^2 w_n}{\partial x^2} \, dx = \left[w_n \frac{\partial w_n}{\partial x}\right]_0^L - \int_0^L \left(\frac{\partial w_n}{\partial x}\right)^2 dx.}
\end{equation}
\revv{The boundary term vanishes due to periodic boundary conditions. Substituting back and retaining terms up to $O(\Delta t)$:}
\begin{align}
   \revv{\|w_{n+1}\|^2} &= \revv{\|w_n\|^2 - 2\alpha\Delta t \left\|\frac{\partial w_n}{\partial x}\right\|^2 + O(\Delta t^2).}
    \label{eq:w_norm_decay}
\end{align}

\revv{Since $-\left\|\frac{\partial w_n}{\partial x}\right\|^2$ is always non-positive, we have $\|w_{n+1}\|^2 \leq \|w_n\|^2$ up to higher-order terms, confirming that the homogeneous heat equation is non-expansive in the $L^2$ norm. However, to reiterate, the inequality is strict only for spatially varying components of the error; the constant spatial mode is not dissipated.}

\revv{To obtain a quantitative bound, we employ the Poincaré inequality on the mean-free component $w_n'$. For periodic functions with zero mean:}
\begin{equation}
  \revv{ \left\|\frac{\partial w_n'}{\partial x}\right\|^2 \geq C \cdot \|w_n'\|^2,}
    \label{eq:poincare}
\end{equation}
\revv{where $C > 0$ depends on the domain size. For a domain $[0, L]$ with periodic boundary conditions: $C = \frac{4\pi^2}{L^2}$.}

\revv{Substituting into Eq.~\ref{eq:w_norm_decay} for the mean-free component:}
\begin{align}
  \revv{\|w_{n+1}'\|^2} &\leq \revv{\|w_n'\|^2 - 2\alpha\Delta t \cdot C \|w_n'\|^2} \nonumber \\
    &= \revv{(1 - 2\alpha\Delta t \cdot C) \|w_n'\|^2.}
    \label{eq:w_contraction_squared}
\end{align}

\revv{Taking square roots:}
\begin{equation}
 \revv{   \|w_{n+1}'\| \leq \sqrt{1 - 2\alpha\Delta t \cdot C} \, \|w_n'\|.}
\end{equation}

\revv{For small $x$, we use the approximation $\sqrt{1-x} \approx 1 - \frac{x}{2}$. Thus:}
\begin{equation}
  \revv{\|w_{n+1}'\| \leq (1 - \alpha\Delta t \cdot C) \|w_n'\|.}
    \label{eq:w_contraction}
\end{equation}

\revv{Let $\lambda = \alpha C$. We define the contraction factor:}
\begin{equation}
   \revv{\rho = 1 - \lambda\Delta t.}
    \label{eq:contraction_factor}
\end{equation}

\textbf{\revv{Stability condition:}} \revv{The above analysis requires that the contraction factor $\rho$ satisfies $0 < \rho < 1$, which imposes a constraint on the time step:}
\begin{equation}
   \revv{0 < \lambda\Delta t < 1 \quad \Rightarrow \quad \Delta t < \frac{1}{\lambda} = \frac{1}{\alpha C} = \frac{L^2}{4\pi^2\alpha}.}
    \label{eq:CFL_condition}
\end{equation}
\revv{This ensures $\rho = 1 - \lambda\Delta t > 0$. Note that the classical CFL stability condition for Forward Euler on the heat equation is $\Delta t < 2/\lambda$, which ensures $|1 - \lambda\Delta t| < 1$. Our analysis uses the stronger assumption $\lambda\Delta t < 1$ to guarantee monotonic contraction (rather than oscillatory decay). Throughout this analysis, we assume that $\Delta t$ is chosen to satisfy this condition.}

\revv{Furthermore, the approximation $\sqrt{1-x} \approx 1 - x/2$ used above is valid when $2\alpha\Delta t C \ll 1$, i.e., when the time step is sufficiently small compared to the stability limit. For time steps closer to the stability limit, the exact expression $\rho = \sqrt{1 - 2\alpha\Delta t C}$ should be used, which remains less than unity under condition~\ref{eq:CFL_condition}.}

\revv{Since $\lambda > 0$ and $\Delta t$ satisfies the stability condition~\ref{eq:CFL_condition}, we have $0 < \rho < 1$. Therefore:}
\begin{equation}
   \revv{\|w_{n+1}'\| \leq \rho \|w_n'\|,}
    \label{eq:contraction_result}
\end{equation}
\revv{confirming that the discrete heat equation is contractive in the $L^2$ norm for the mean-free component of the error.}

\textbf{\revv{Mean-mode evolution:}}
\revv{
The spatial mean of the error evolves separately. Taking the spatial average of Eq.~\ref{eq:error_evolution} and using periodic boundary conditions gives
}
\begin{equation}
\revv{
    \bar{w}_{n+1}
    =
    \bar{w}_n
    -
    \Delta t\,\bar{\varepsilon}_n,
}
    \label{eq:mean_mode_evolution}
\end{equation}
\revv{
because $\int_0^L \alpha w_{xx}\,dx=0$. If $u_0=\hat{u}_0$, then $\bar{w}_0=0$, and therefore
}
\begin{equation}
\revv{
    \bar{w}_n
    =
    -\Delta t\sum_{j=0}^{n-1}\bar{\varepsilon}_j.
}
    \label{eq:mean_mode_accumulation}
\end{equation}
\revv{
Thus, the spatial-mean component is not dissipated by the heat equation; it is governed by the accumulated spatial mean of the surrogate RHS error.
}

\textbf{\revv{Connection to the error equation:}} \revv{Examining the first term on the RHS of Eq.~\ref{eq:error_triangle}:}
\begin{equation}
    \revv{(u_n - \hat{u}_n) + \Delta t \cdot [\mathcal{N}(u_n) - \mathcal{N}(\hat{u}_n)].}
\end{equation}

\revv{Using the linearity of $\mathcal{N}$:}
\begin{equation}
  \revv{  \mathcal{N}(u_n) - \mathcal{N}(\hat{u}_n) = \mathcal{N}(u_n - \hat{u}_n) = \alpha \frac{\partial^2(u_n - \hat{u}_n)}{\partial x^2} = \alpha \frac{\partial^2 w_n}{\partial x^2}.}
\end{equation}

\revv{Therefore, the bracketed term is precisely:}
\begin{equation}
  \revv{ w_n + \Delta t \cdot \alpha \frac{\partial^2 w_n}{\partial x^2} = w_{n+1}^{\text{(homogeneous)}},}
\end{equation}
\revv{which is the evolution of the error under the homogeneous heat equation (without forcing from the surrogate error). This identification allows us to leverage the dissipativity analysis conducted above. For the mean-free component, this gives the contraction estimate in Eq.~\ref{eq:contraction_result}.}

\textbf{\revv{Remark:}} \revv{For nonlinear PDEs where $\mathcal{N}$ is not linear, the analogous analysis would require a Lipschitz condition of the form $\|\mathcal{N}(u) - \mathcal{N}(v)\| \leq L_{\mathcal{N}}\|u - v\|$, and the contraction property would depend on the interplay between the Lipschitz constant and any inherent dissipativity of the system.}

\revv{
Let $e_n' := \|w_n'\|$ denote the norm of the mean-free component of the error. Let $\epsilon_S'$ denote a bound on the mean-free component of the surrogate RHS error, i.e., $\|\varepsilon_n'\|\leq \epsilon_S'$. Thus, we obtain the following mean-free error recurrence relation:
}
\begin{align}
   \revv{e_{n+1}'} &\leq \revv{\|w_{n+1}^{\prime,\text{(homogeneous)}}\| + \Delta t \cdot \epsilon_S'} \nonumber \\
    &\leq \revv{\rho e_n' + \Delta t \cdot \epsilon_S'.}
    \label{eq:error_recurrence}
\end{align}

\textbf{\revv{Long-time behavior of pure TI-NO:}}

\revv{Starting from $e_0' = 0$ (same initial condition), let us track the mean-free error over $n$ steps:}
\begin{align}
   \revv{ e_1'} &\leq \revv{\rho \cdot 0 + \epsilon_S'\Delta t = \epsilon_S'\Delta t,} \\
    \revv{e_2'} &\leq \revv{\rho \cdot e_1' + \epsilon_S'\Delta t \leq \rho\epsilon_S'\Delta t + \epsilon_S'\Delta t = \epsilon_S'\Delta t(1 + \rho),} \\
    \revv{e_3'} &\leq \revv{\rho \cdot e_2' + \epsilon_S'\Delta t \leq \epsilon_S'\Delta t(\rho + \rho^2) + \epsilon_S'\Delta t = \epsilon_S'\Delta t(1 + \rho + \rho^2).}
\end{align}

\revv{After $n$ steps:}
\begin{equation}
   \revv{e_n' \leq \epsilon_S'\Delta t \sum_{j=0}^{n-1} \rho^j = \epsilon_S'\Delta t \cdot \frac{1 - \rho^n}{1 - \rho}.}
    \label{eq:error_geometric_series}
\end{equation}

\revv{As $n \to \infty$:}
\begin{equation}
    \revv{e_\infty' \leq \frac{\epsilon_S'\Delta t}{1 - \rho} = \frac{\epsilon_S'\Delta t}{\lambda\Delta t} = \frac{\epsilon_S'}{\lambda}.}
    \label{eq:error_steady_state}
\end{equation}

\textbf{\revv{Interpretation for pure TI-NO:}}
\begin{equation}
  \revv{\boxed{e_\infty^{\prime,\text{surrogate}} \leq \frac{\epsilon_S'}{\lambda}}}
    \label{eq:pure_surrogate_bound}
\end{equation}

\revv{Here, $\epsilon_S'$ represents how well the network learned the mean-free component of $\mathcal{N}$, and $\lambda$ is the dissipation rate of the PDE. Thus, the mean-free component of the error saturates at a level determined by the ratio of mean-free network error to dissipation strength.}

\revv{
The full error additionally contains the spatial-mean contribution. For a periodic function decomposed as $w_n = \bar{w}_n + w_n'$, the $L^2$ norm satisfies
}
\begin{equation}
\revv{
    \|w_n\|^2 = \int_0^L |w_n|^2\,dx = \int_0^L |\bar{w}_n|^2\,dx + \int_0^L |w_n'|^2\,dx = L|\bar{w}_n|^2 + \|w_n'\|^2,
}
\end{equation}
\revv{
where the cross-term vanishes because $w_n'$ has zero spatial mean. Taking the square root and applying the inequality $\sqrt{a^2 + b^2} \leq |a| + |b|$ yields
}
\begin{equation}
\revv{
    \|w_n\| = \sqrt{\|w_n'\|^2 + L|\bar{w}_n|^2} \leq \|w_n'\| + \sqrt{L}\,|\bar{w}_n|.
}
\end{equation}
\revv{
From Eq.~\ref{eq:error_steady_state}, the mean-free component satisfies $\|w_n'\| \leq \epsilon_S'/\lambda$. From Eq.~\ref{eq:mean_mode_accumulation}, the spatial mean satisfies
}
\begin{equation}
\revv{
    |\bar{w}_n| = \left|\Delta t\sum_{j=0}^{n-1}\bar{\varepsilon}_j\right| \leq \Delta t\sum_{j=0}^{n-1}|\bar{\varepsilon}_j|.
}
\end{equation}
\revv{
Combining these bounds, we obtain the relaxed full-error estimate
}
\begin{equation}
\revv{
    \boxed{
    \|w_n\|
    \leq
    \frac{\epsilon_S'}{\lambda}
    +
    \sqrt{L}\,\Delta t\sum_{j=0}^{n-1}|\bar{\varepsilon}_j|
    }.
}
    \label{eq:relaxed_full_error_bound}
\end{equation}
\revv{
If the spatial mean of the surrogate RHS error is small, the second term is small and the mean-free bound provides a good approximation to the observed full $L^2$ behavior. In the special case where $\bar{\varepsilon}_j=0$ for all $j$, the original full-error bound is recovered exactly.
}

\revv{The error recurrence in Eq.~\ref{eq:error_recurrence} reveals that the pure TI-NO surrogate phase follows a contractive (due to the dissipativity characteristics encoded in $\rho$) and forced system (due to the error induced by the TI-NO surrogate, $\epsilon_S'$) for the mean-free component of the error.}

\revv{Following a similar analysis, the error recurrence relation during the numerical solver phase (where $\varepsilon = 0$) can be derived as:}
\begin{equation}
\revv{    \boxed{e_{n+1}' \leq (1 - \lambda\Delta t) e_n'} } 
    \label{eq:solver_error_recurrence}
\end{equation}
\revv{
while the spatial mean remains unchanged:
}
\begin{equation}
\revv{
    \bar{w}_{n+1}=\bar{w}_n.
}
    \label{eq:solver_mean_recurrence}
\end{equation}

\textbf{\revv{Residual-based error estimator:}}

\revv{Let us now define the discrete PDE residual evaluated on the surrogate solution $\hat{u}_n$. In the context of Forward Euler time-stepping, the surrogate computes:}
\begin{equation}
  \revv{  \hat{u}_{n+1} = \hat{u}_n + \Delta t \cdot \hat{\mathcal{N}}(\hat{u}_n).}
\end{equation}

\revv{The residual measures how well this update satisfies the true PDE operator:}
\begin{equation}
   \revv{ R_n := \hat{\mathcal{N}}(\hat{u}_n) - \mathcal{N}(\hat{u}_n).}
    \label{eq:residual_def}
\end{equation}

\revv{This represents the instantaneous error in the right-hand side approximation at time step $n$. From our earlier assumption on the surrogate error:}
\begin{equation}
    \revv{R_n = \varepsilon(\hat{u}_n), \quad \text{and thus} \quad \|R_n\| \leq \epsilon_S.}
\end{equation}

\revv{
We decompose the residual into its spatial mean and mean-free components:
}
\begin{equation}
\revv{
    R_n = \bar{R}_n + R_n',
    \qquad
    R_n' = R_n-\bar{R}_n.
}
\end{equation}
\revv{
The mean-free residual $R_n'$ drives the dissipative mean-free error recurrence, while $\bar{R}_n$ governs the accumulated spatial-mean error.
}

\revv{The residual $R_n$ is computable during the simulation: given the surrogate's output $\hat{\mathcal{N}}(\hat{u}_n)$, one can evaluate the true operator $\mathcal{N}(\hat{u}_n) = \alpha \frac{\partial^2 \hat{u}_n}{\partial x^2}$ using finite differences and compute their difference.}

\revv{In effect, during the surrogate phase, we can write for the mean-free error:}
\begin{equation}
  \revv{e_{n+1}' \leq (1 - \lambda\Delta t) e_n' + \|R_n'\| \Delta t.}
    \label{eq:error_with_residual}
\end{equation}

\revv{Furthermore, an exponential moving average (EMA)-based error estimator is defined as:}
\begin{equation}
   \revv{\eta_n = (1 - \beta)\eta_{n-1} + \beta\|R_n\|,}
    \label{eq:EMA_def}
\end{equation}
\revv{where $\beta \in (0, 1)$ is the smoothing parameter.}

\revv{Expanding the recurrence:}
\begin{equation}
   \revv{ \eta_n = \beta\|R_n\| + \beta(1-\beta)\|R_{n-1}\| + \beta(1-\beta)^2\|R_{n-2}\| + \cdots + (1-\beta)^n\eta_0.}
\end{equation}

\revv{Since all terms are non-negative:}
\begin{equation}
   \revv{ \eta_n \geq \beta\|R_n\|,}
\end{equation}
\revv{which implies:}
\begin{equation}
    \revv{\|R_n'\| \leq \|R_n\| \leq \frac{\eta_n}{\beta}.}
    \label{eq:residual_bound}
\end{equation}

\revv{Now, during the surrogate phase:}
\begin{equation}
  \revv{e_{n+1}' \leq (1 - \lambda\Delta t) e_n' + \frac{\Delta t}{\beta}\eta_n.}
    \label{eq:error_with_EMA}
\end{equation}

\textbf{\revv{Derivation of steady-state error bound during surrogate phase:}}

\revv{The switching criterion triggers when $\eta_n \geq \tau_n$, where $\tau_n$ is the time-dependent threshold of the error estimator at timestep $n$. To establish the error at the switching point, we consider the quasi-steady-state behavior when the EMA hovers near the threshold.}

\revv{If the system were at steady state with $\eta_n = \tau$ (constant), the mean-free error would satisfy:}
\begin{equation}
    \revv{e_\infty' = (1 - \lambda\Delta t) e_\infty' + \frac{\Delta t}{\beta}\tau.}
\end{equation}

\revv{Solving for $e_\infty'$:}
\begin{align}
   \revv{ e_\infty'[1 - (1 - \lambda\Delta t)]} &= \revv{\frac{\Delta t}{\beta}\tau,} \nonumber \\
    \revv{e_\infty' \cdot \lambda\Delta t} &= \revv{\frac{\Delta t \cdot \tau}{\beta},} \nonumber \\
    \revv{e_\infty'} &= \revv{\frac{\tau}{\beta\lambda}.}
    \label{eq:steady_state_derivation}
\end{align}

\revv{Let us say the switching happens at timestep $n_0$. At the switching point, where $\eta_{n_0} \geq \tau_{n_0}$, the mean-free error is bounded by this steady-state value:}
\begin{equation}
   \revv{ e_{n_0}' \leq \frac{\tau_{n_0}}{\beta\lambda}.}
    \label{eq:error_at_switch}
\end{equation}

\revv{This bound may be conservative if the surrogate phase was short or if the error had not yet reached steady state; in such cases, the actual error would be smaller.}

\textbf{\revv{Error decay during numerical solver phase:}}

\revv{From our earlier analysis, during the numerical solver phase (where the exact operator $\mathcal{N}$ is used):}
\begin{equation}
  \revv{e_{n+1}' \leq (1 - \lambda\Delta t) e_n'.}
\end{equation}

\revv{If we run $M$ steps of the numerical solver beginning from timestep $n_0$:}
\begin{equation}
   \revv{ e_{n_0 + M}' \leq (1 - \lambda\Delta t)^M e_{n_0}'.}
    \label{eq:error_decay_M_steps}
\end{equation}

\revv{Combining with the bound in Eq.~\ref{eq:error_at_switch}:}
\begin{equation}
    \revv{e_{n_0 + M}' \leq (1 - \lambda\Delta t)^M \frac{\tau_{n_0}}{\beta\lambda}.}
    \label{eq:error_after_solver}
\end{equation}

\revv{
For the full error during the numerical solver phase, the spatial mean remains unchanged, while the mean-free component decays. Therefore,
}
\begin{equation}
\revv{
    \boxed{
    \|w_{n_0+M}\|^2
    \leq
    (1-\lambda\Delta t)^{2M}\|w_{n_0}'\|^2
    +
    L|\bar{w}_{n_0}|^2.
    }
}
    \label{eq:full_error_after_solver}
\end{equation}
\revv{
This explains why invoking the numerical solver can produce a visible dip in the relative $L^2$ error: the solver dissipates the spatially varying component of the surrogate-induced error. However, any spatially constant error component is not dissipated by the heat equation and is controlled separately by Eq.~\ref{eq:mean_mode_accumulation}.
}

\textbf{\revv{Error bounds for the ANCHOR framework:}}

\revv{Thus, we establish the following error bound for the ANCHOR framework on the mean-free component of the error:}
\begin{equation}
   \revv{ \boxed{e_n' \leq \frac{\tau_n}{\beta\lambda}}}
    \label{eq:ANCHOR_upper_bound}
\end{equation}

\revv{This upper bound is achieved when the surrogate phase reaches quasi-steady-state before switching. After $M$ steps of the numerical solver phase, the mean-free error decays as:}
\begin{equation}
   \revv{ e_{n_0 + M}' \leq (1 - \lambda\Delta t)^M e_{n_0}' \leq (1 - \lambda\Delta t)^M \frac{\tau_{n_0}}{\beta\lambda}.}
    \label{eq:error_decay_solver}
\end{equation}

\revv{
For the full $L^2$ error, the corresponding relaxed ANCHOR estimate is
}
\begin{equation}
\revv{
   \boxed{
   e_n
   \leq
   \frac{\tau_n}{\beta\lambda}
   +
   \sqrt{L}\Delta t
   \sum_{j=0}^{n-1}|\bar{R}_j|
   }.
}
    \label{eq:ANCHOR_full_error_relaxed}
\end{equation}
\revv{
Thus, the original ANCHOR bound is recovered exactly when the spatial mean of the residual vanishes, i.e., $\bar{R}_j=0$ for all $j$. Without this condition, the first term captures the dissipative mean-free error controlled by ANCHOR, while the second term captures the accumulated spatial-mean bias of the surrogate RHS.
}

\textbf{\revv{Interpretation:}} 
\revv{
The ANCHOR framework maintains the mean-free component of the error within a band characterized by:
}
\begin{itemize}[leftmargin=*,nosep]
    \item \revv{\textbf{Upper envelope:} 
    $e_{\max}' = \dfrac{\tau_n}{\beta\lambda}$, reached when the surrogate phase is active and the EMA error estimator approaches the threshold.}
    
    \item \revv{\textbf{Lower envelope:} 
    $e_{\min}' \sim (1 - \lambda\Delta t)^M \cdot e_{\max}'$, approached after $M$ steps of the numerical solver phase, where the dissipative nature of the PDE reduces the accumulated mean-free error.}
\end{itemize}

\revv{
The ratio between the mean-free upper and lower envelopes is controlled by $(1 - \lambda\Delta t)^M$, which depends on:
}
\begin{itemize}[leftmargin=*,nosep]
    \item \revv{The dissipation rate $\lambda = \alpha C$ of the underlying PDE,}
    \item \revv{The time step $\Delta t$,}
    \item \revv{The number of solver steps $M$ before switching back to the surrogate.}
\end{itemize}

\revv{
For the full $L^2$ error, the same banded behavior holds up to the spatial-mean contribution. Specifically, since
}
\begin{equation}
\revv{
    e_n^2 = \|w_n\|^2 = \|w_n'\|^2 + L|\bar{w}_n|^2,
}
\end{equation}
\revv{
the numerical solver phase reduces the mean-free component $\|w_n'\|$, while the spatial-mean component $|\bar{w}_n|$ is not dissipated by the heat equation. Thus, the observed decrease in relative $L^2$ error after solver invocation is attributed to the decay of the spatially varying component of the error. Any persistent spatial-mean bias contributes an additional non-dissipative offset governed by
}
\begin{equation}
\revv{
    \bar{w}_n
    =
    -\Delta t\sum_{j=0}^{n-1}\bar{R}_j.
}
\end{equation}

\revv{
Consequently, the relaxed full-error upper envelope may be written as
}
\begin{equation}
\revv{
    e_{\max}
    \lesssim
    \frac{\tau_n}{\beta\lambda}
    +
    \sqrt{L}\Delta t
    \sum_{j=0}^{n-1}|\bar{R}_j|,
}
\end{equation}
\revv{
while after $M$ numerical solver steps the full error satisfies
}
\begin{equation}
\revv{
    e_{n_0+M}^2
    \leq
    (1-\lambda\Delta t)^{2M}(e_{n_0}')^2
    +
    L|\bar{w}_{n_0}|^2.
}
\end{equation}

\revv{\textbf{Remark:} 
The lower envelope represents a typical minimum error rather than a strict lower bound. The actual error could be smaller due to favorable phase alignment between numerical and surrogate errors. However, in the absence of such fortuitous cancellations, the error band characterization provides a practical description of the framework's accuracy. To reiterate, the contraction factor $(1-\lambda\Delta t)^M$ should be interpreted as governing the decay of the mean-free, dissipative component of the error. The full $L^2$ error may additionally contain a non-dissipative spatial-mean contribution, which vanishes when the residual has zero spatial mean or remains small when the residual mean bias is small.}

\revv{\textbf{Connection to numerical experiments:} 
The theoretical predictions derived above are corroborated by the numerical experiments in Section~\ref{sec:results}. Specifically: (i) the error reduction upon solver invocation, manifesting as characteristic ``dips'' in the relative $L^2$ error plots, and (ii) the banded error structure oscillating between upper and lower envelopes are consistently observed across all six test cases. This agreement, spanning linear and nonlinear PDEs in 1D, 2D, and 3D, provides empirical validation that the error correction mechanism elucidated by the analysis extends beyond the 1D heat equation.}

\section{Computational Cost}
\subsection{\revv{Data Generation}}
\revv{Numerical simulations were performed on a shared HPC cluster partition with 4 GB memory per core. The number of parallel workers used for each PDE case is listed in Table~\ref{tab:pde_runtime}.}
{\color{black}
\begin{table}[htbp]
\centering
\caption{\revv{Computational time for dataset generation across all PDEs. Here, $T$ denotes the final simulation time.}}
\renewcommand{\arraystretch}{1.1}
\label{tab:pde_runtime}
\resizebox{\textwidth}{!}{%
\begin{tabular}{|c|c|c|c|c|c|c|c|}
\hline
\revv{PDE Example} & \revv{Execution Time (secs)} & \revv{\# Workers} 
& \revv{$N_x$} & \revv{$N_y$} & \revv{$N_z$} 
& \revv{$\Delta t_{solve}$} & \revv{$T$} \\
\hline
\revv{Burgers 1D}              & \revv{25.916}    & \revv{10} & \revv{101} & \revv{--}  & \revv{--} & \revv{$10^{-4}$} & \revv{1.0}  \\
\hline
\revv{Burgers 2D}             & \revv{700.26}   & \revv{20} & \revv{101} & \revv{101} & \revv{--} & \revv{$10^{-4}$} & \revv{1.0}  \\
\hline
\revv{Allen-Cahn 2D}        & \revv{171.87}    & \revv{20} & \revv{32}  & \revv{32}  & \revv{--} & \revv{$5.0\times 10^{-5}$} & \revv{1.0}  \\
\hline
\revv{Cahn-Hilliard 2D}      & \revv{2413.44}   & \revv{40} & \revv{64}  & \revv{64}  & \revv{--} & \revv{$10^{-4}$} & \revv{2.0}  \\
\hline
\revv{Navier-Stokes 2D}    & \revv{34824.36} & \revv{50} & \revv{256} & \revv{256} & \revv{--} & \revv{$10^{-4}$} & \revv{15.0} \\
\hline
\revv{Heat 3D}               & \revv{511.85}    & \revv{20} & \revv{32}  & \revv{32}  & \revv{16} & \revv{$10^{-2}$} & \revv{1.0}  \\
\hline
\end{tabular}%
}
\end{table}
}

\subsection{Training}
\rev{Training for the 1D Burgers’, 2D Burgers', and 2D Allen–Cahn cases was carried out on a single NVIDIA A100 GPU with 40 GB memory, paired with an Intel Xeon Gold Cascade Lake CPU. For the 2D Cahn-Hilliard, \revv{2D Navier-Stokes,} and 3D heat conduction cases, experiments were conducted on an NVIDIA A100 GPU with 80 GB memory, paired with an Intel Xeon Gold Ice Lake CPU.}

\begin{table}[htb!]
\centering
\renewcommand{\arraystretch}{1.2}
\caption{\rev{Training throughput and model size for the neural operator surrogates considered in this study. Training speed is reported as iterations per second (higher is faster) under a fixed hardware/software configuration for each PDE, and model size as the number of trainable parameters.}}
\begin{tabular}{|c|c|c|c|c|}
\hline
\multirow{2}{*}{PDE Example} & \multirow{2}{*}{Method} & \multirow{2}{*}{\# Parameters} & \multirow{2}{*}{Batch Size} & Training Speed \\
 & & & & (iter/s) \\
\hline
\multirow{4}{*}{1D Burgers'}
 & AR-DON & 114,025  & \multirow{4}{*}{256} & 296.09 \\
 & TI-DON & 144,025  & & 163.62 \\
 & AR-FNO & 1,204,865 & & 263.72 \\
 & TI-FNO & 1,204,865 & & 81.98 \\
\hline
\multirow{4}{*}{2D Burgers'}
 & AR-DON & 1,207,560 & \multirow{4}{*}{64} & 98.59 \\
 & TI-DON & 1,174,537 & & 37.01 \\
 & AR-FNO & 4,198,689 & & 169.47 \\
 & TI-FNO & 4,198,689 & & 115.22 \\
\hline
\multirow{4}{*}{2D Allen-Cahn}
 & AR-DON & 1,224,072 & \multirow{4}{*}{64} & 108.44 \\
 & TI-DON & 1,306,632 & & 34.55 \\
 & AR-FNO & 4,198,689 & & 314.72 \\
 & TI-FNO & 4,198,689 & & 305.62 \\
\hline
\multirow{2}{*}{2D Cahn-Hilliard}
 & AR-FNO & 16,781,601 & \multirow{2}{*}{256} & 65.01 \\
 & TI-FNO & 16,781,601 & & 17.08 \\
\hline
\multirow{2}{*}{\revv{2D Navier-Stokes}}
 & \revv{AR-FNO} & \revv{14,162,273} & \multirow{2}{*}{\revv{256}} & \revv{20.46} \\
 & \revv{TI-FNO} & \revv{14,162,273} &  & \revv{7.23} \\
\hline
\multirow{2}{*}{3D Heat}
 & AR-DON & 246,193 & \multirow{2}{*}{32} & 76.61 \\
 & TI-DON & 246,193 & & 20.32 \\
\hline
\end{tabular}
\label{tab:training_cost}
\end{table}

\clearpage
\section{Additional Results}
\subsection{One-dimensional Burgers' Equation}
\begin{figure}[!htb]
    \centering
    \includegraphics[width=\linewidth]{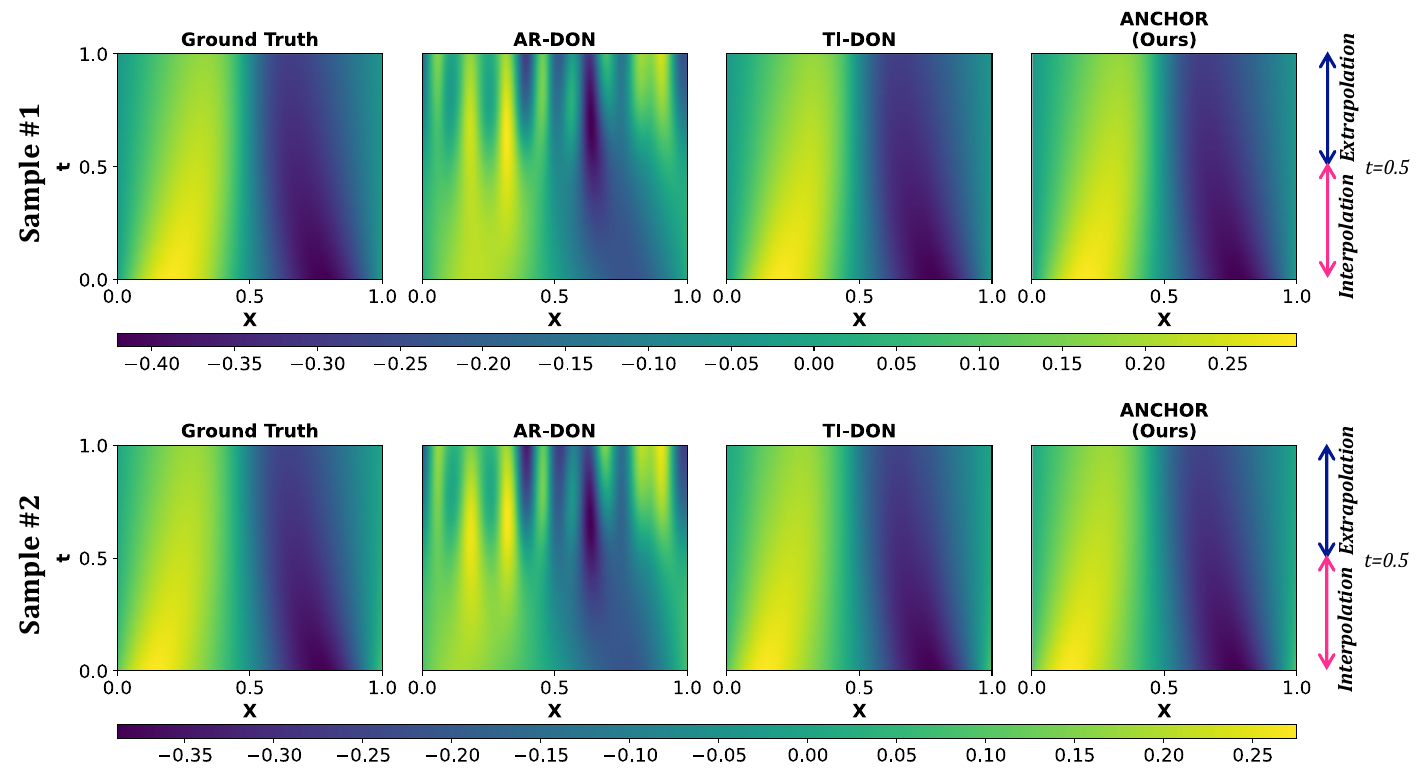}
    \caption{1D Burgers’ equation: Solution contours over $t \in [0,1]$ for all frameworks, illustrated using two representative samples. The color bar on the far right indicates the time steps solved by TI-DON (blue) and by the high-fidelity numerical solver (pink). Here, $t \in [0,0.5]$ corresponds to the interpolation regime, while $t \in [0.5,1.0]$ denotes the extrapolation regime.}
    \label{fig:1d_burgers_sol_contours}
\end{figure}

\clearpage
\subsection{Two-dimensional Burgers' Equation}
\begin{figure}[htb!]
    \centering
    \includegraphics[width=\linewidth]{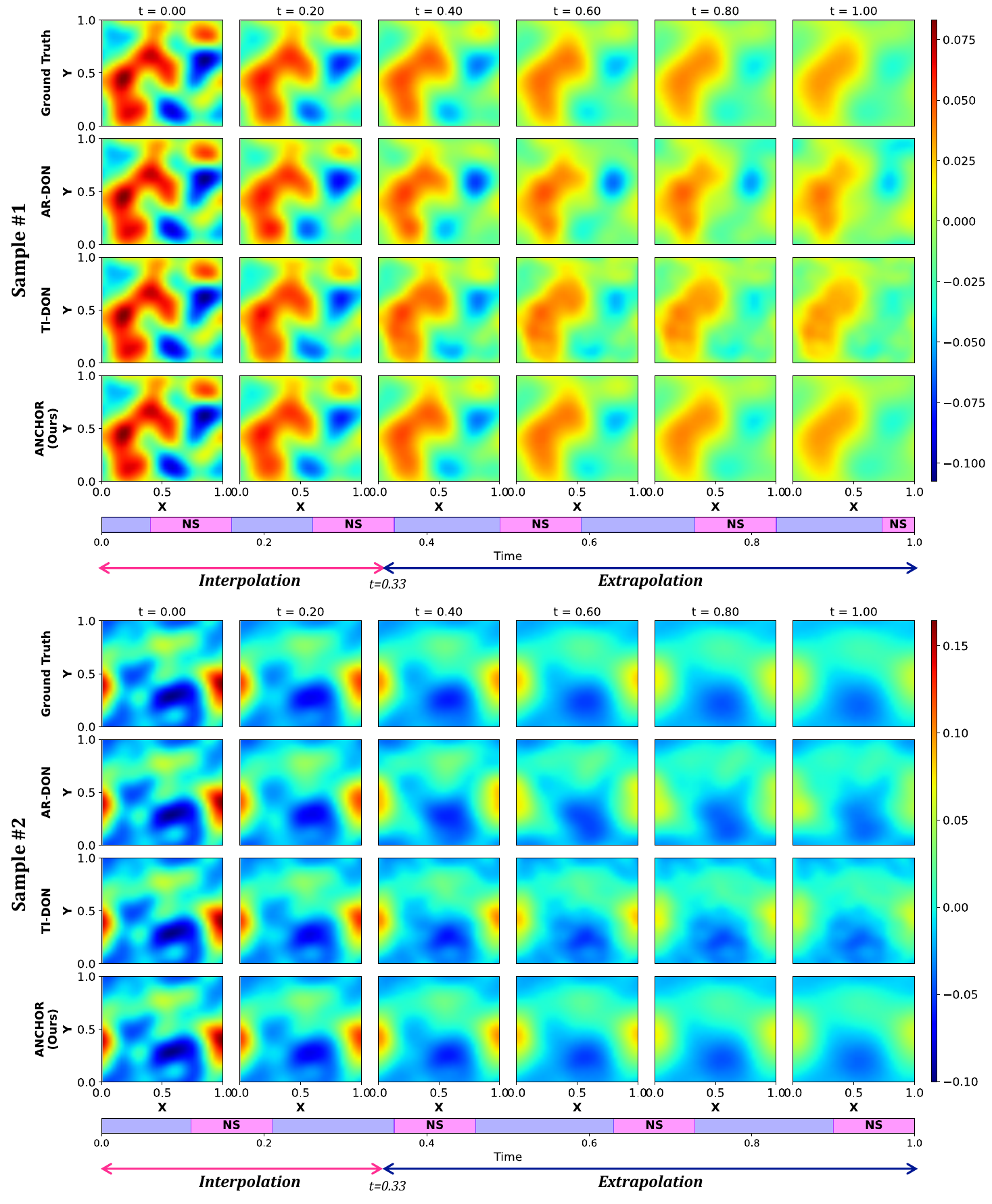}
    \caption{2D Burgers' equation: Solution contours over $t\in[0,1]$, where $t\in[0,0.33]$ corresponds to the interpolation regime and $t\in[0.33,1.0]$ corresponds to extrapolation for all frameworks, illustrated using two representative samples. The color bar below each set of contours indicates the time steps solved by TI-DON (blue) and by the high-fidelity numerical solver (magenta).}
    \label{fig:2d_burgers_sol_contours}
\end{figure}

\clearpage
\subsection{Two-dimensional Allen-Cahn Equation}
\begin{figure}[htb!]
    \centering
    \includegraphics[width=\linewidth]{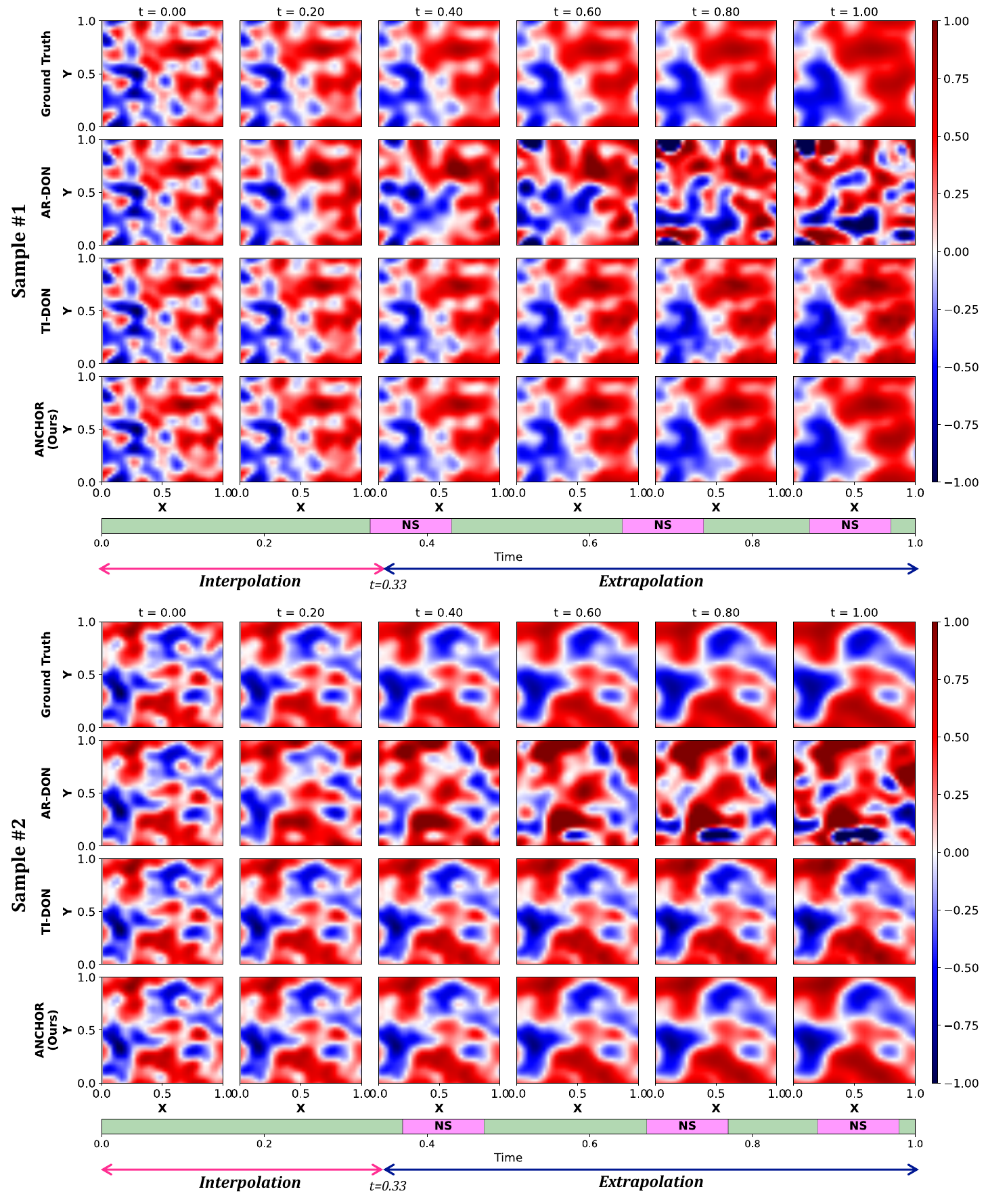}
    \caption{2D Allen-Cahn equation: Solution contours over $t\in[0,1]$, where $t\in[0,0.33]$ corresponds to the interpolation regime and $t\in[0.33,1.0]$ corresponds to extrapolation for all frameworks, illustrated using two representative samples. The color bar below each set of contours indicates the time steps solved by TI-DON (green) and by the high-fidelity numerical solver (magenta).}
    \label{fig:2d_allen_cahn_sol_contours}
\end{figure}

\clearpage
\subsection{Two-dimensional Cahn-Hilliard Equation}
\begin{figure}[htb!]
    \centering
    \includegraphics[width=\linewidth]{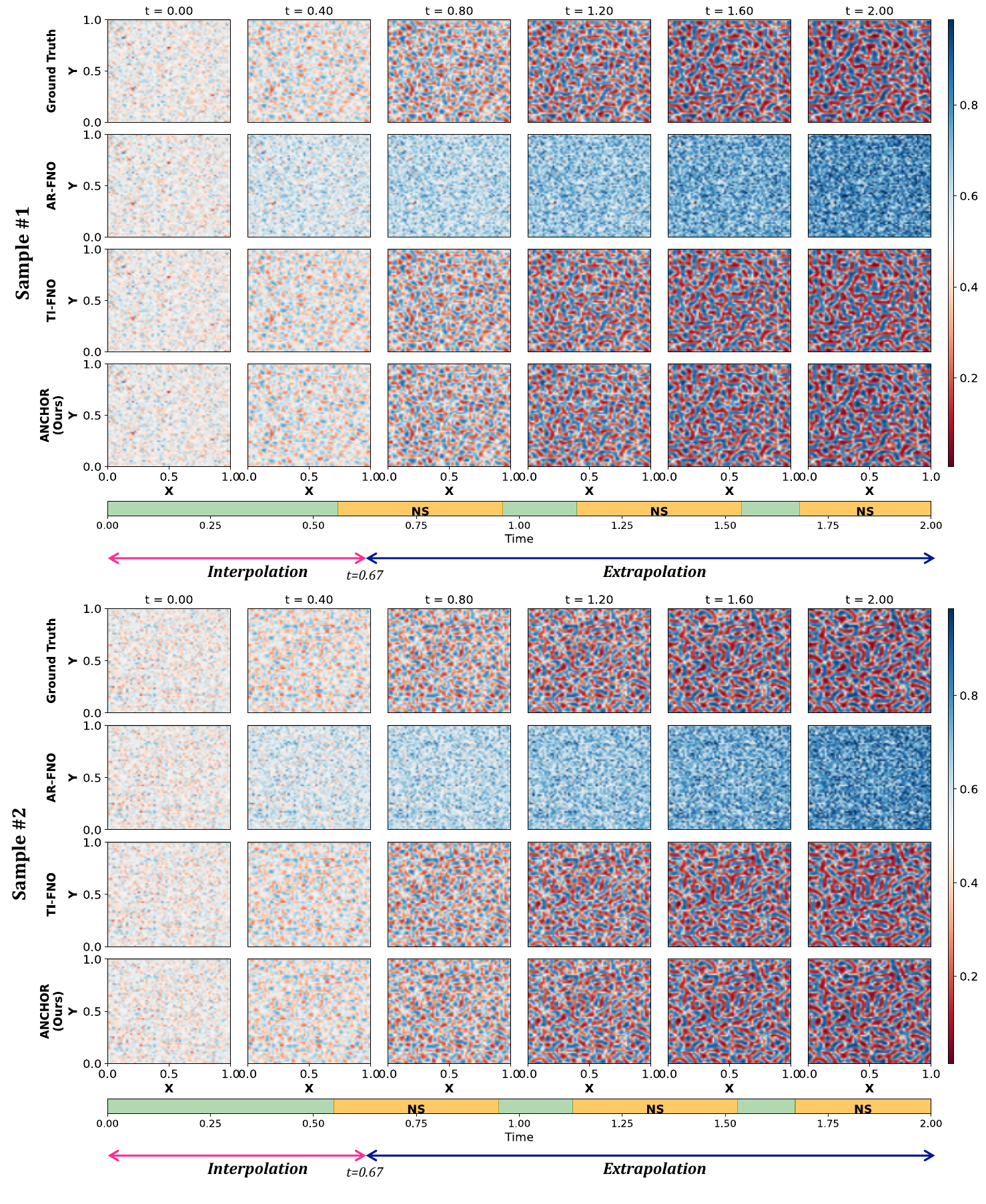}
    \caption{\rev{2D Cahn-Hillard equation: Solution contours over $t\in[0,1]$, where $t\in[0,0.67]$ corresponds to the interpolation regime and $t\in[0.67,2.0]$ corresponds to extrapolation for all frameworks, illustrated using two representative samples. The color bar below each set of contours indicates the time steps solved by TI-FNO (green) and by the high-fidelity numerical solver (orange).}}
    \label{fig:2d_allen_cahn_sol_contours}
\end{figure}

\clearpage
\subsection{\revv{Two-dimensional Navier-Stokes Equation}}
\begin{figure}[htb!]
    \centering
    \includegraphics[width=\linewidth]{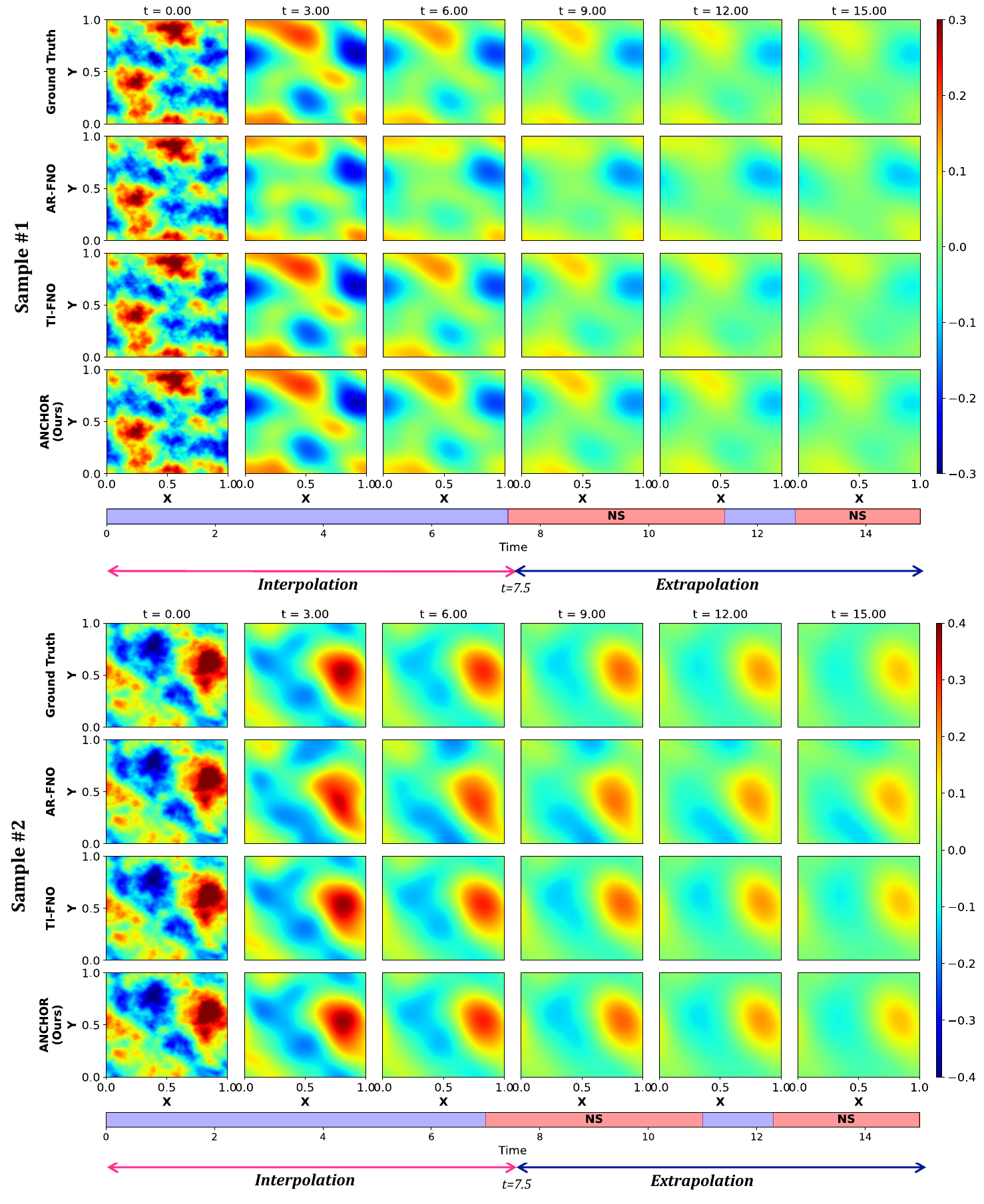}
    \caption{\revv{2D Navier-Stokes equation: Solution contours over $t\in[0,15]$, where $t\in[0,7.5]$ corresponds to the interpolation regime and $t\in[7.5,15.0]$ corresponds to extrapolation for all frameworks, illustrated using two representative samples. The color bar below each set of contours indicates the time steps solved by TI-FNO (blue) and by the high-fidelity numerical solver (red).}}
    \label{fig:2d_allen_cahn_sol_contours}
\end{figure}

\newpage
\subsection{Three-dimensional Heat Conduction}
\begin{figure}[htb!]
    \centering
    \includegraphics[width=0.93\linewidth]{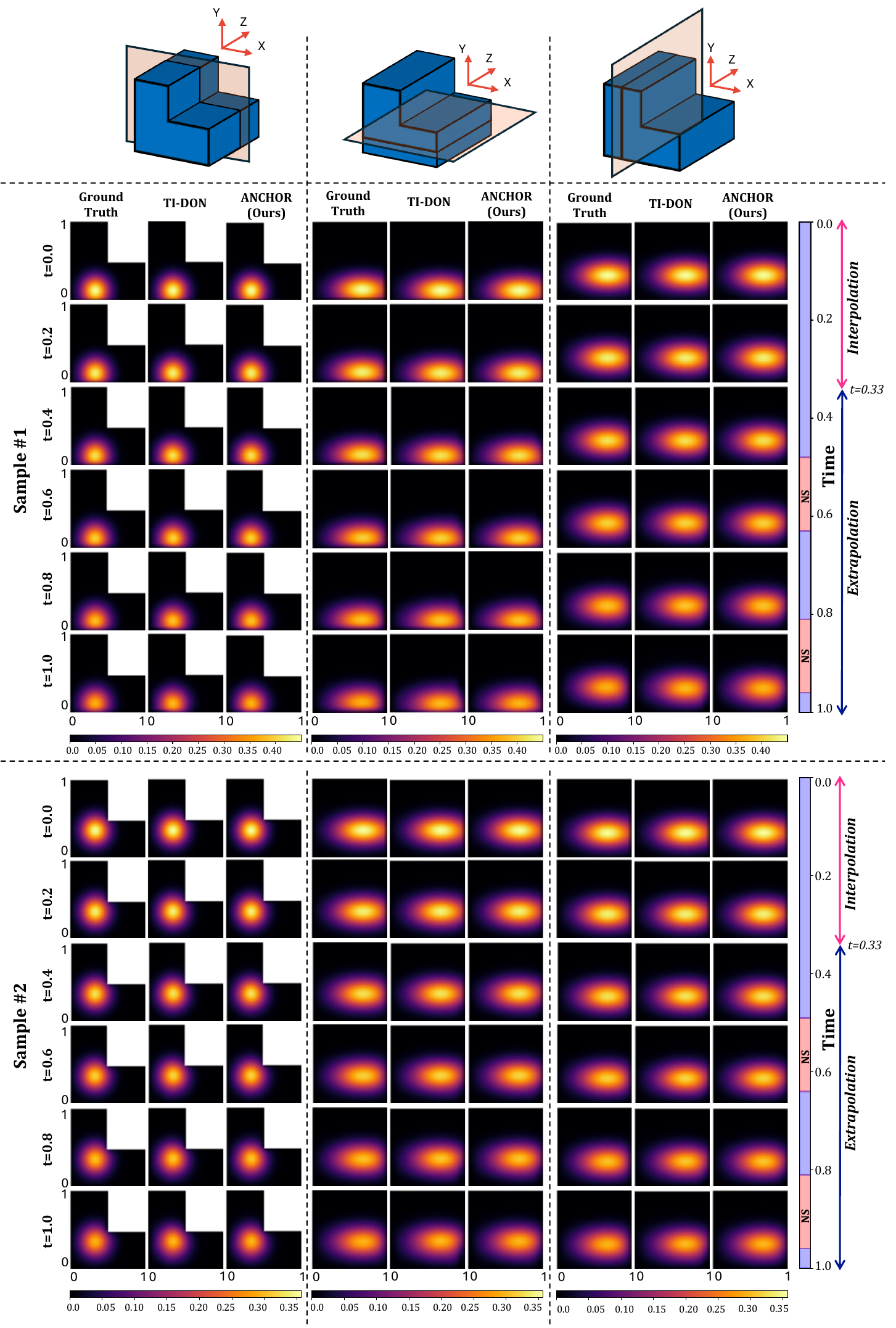}
    \caption{3D Heat conduction equation: Ground truth and predicted solution fields for TI-DON and the ANCHOR framework over $t \in [0,1]$ at selected time steps. For visualization, two-dimensional solution contours on three slicing planes - XY, ZX, and YZ (shown in this order) are presented for two representative test samples (sample~\#2). The color bar at the extreme right indicates the time steps solved by TI-DON (blue) and by the high-fidelity numerical solver (pink). Here, $t \in [0,0.33]$ corresponds to the interpolation regime, while $t \in [0.33,1.0]$ denotes the extrapolation regime.}
    \label{fig:3d_heat_sol_contours}
\end{figure}

\end{document}